\documentclass[draft]{agujournal2019}
\usepackage{url} %this package should fix any errors with URLs in refs.
\usepackage{subcaption}
\usepackage{lineno}
\usepackage[inline]{trackchanges} %for better track changes. finalnew option will compile document with changes incorporated.
\usepackage{soul}
\usepackage{amssymb}
\usepackage{amsmath}
\usepackage{caption}

\usepackage{subcaption}
\graphicspath{{arxivfig/}}
\draftfalse

\journalname{ }

\begin{document}

\title{Generative Model-Based Parameterization and Geometry-Aware Surrogate Model for Fracture Network Inversion}

\authors{Weiquan Ouyang\affil{1}, Su Jiang\affil{1}}

\affiliation{1}{Department of Civil and Environmental Engineering, Carnegie Mellon University, Pittsburgh, PA, 15213}

\correspondingauthor{Su Jiang}{sujiang@andrew.cmu.edu}

\begin{keypoints}
\item Vertex-based flow matching enables generative parameterization of complex fracture network geometries.
\item A geometry-aware surrogate accurately predicts flow and transport responses in fractured systems.
\item Latent-space Bayesian inversion reduces uncertainty in fracture geometries and reservoir predictions.
\end{keypoints}

%%%%%%%%%%%%%%%%%%%%%%%%%%%%%%%%%%%%%%%%%%%%%%%
%
%  ABSTRACT and PLAIN LANGUAGE SUMMARY
%
% A good Abstract will begin with a short description of the problem
% being addressed, briefly describe the new data or analyses, then
% briefly states the main conclusion(s) and how they are supported and
% uncertainties.

% The Plain Language Summary should be written for a broad audience,
% including journalists and the science-interested public, that will not have 
% a background in your field.
%
% A Plain Language Summary is required in GRL, JGR: Planets, JGR: Biogeosciences,
% JGR: Oceans, G-Cubed, Reviews of Geophysics, and JAMES.
% see http://sharingscience.agu.org/creating-plain-language-summary/)
%
%%%%%%%%%%%%%%%%%%%%%%%%%%%%%%%%%%%%%%%%%%%%%%%

%% \begin{abstract} starts the second page

\begin{abstract}
Characterizing fracture networks from limited observations is challenging due to uncertain fracture geometries and nonlinear relationships between fracture connectivity and flow and transport responses. We develop a Bayesian inversion framework that integrates generative model-based parameterization, geometry-aware surrogate modeling, and latent-space sampling. A vertex-based flow matching model maps Gaussian latent variables to discrete fracture networks, directly representing individual fractures through their vertex coordinates without image-based discretization. A geometry-aware surrogate combines graph neural networks and attention mechanisms to predict spatially and temporally varying temperature and tracer concentration fields using an embedded discrete fracture model representation. The generative and surrogate models are integrated with parallel-tempering Markov chain Monte Carlo for latent-space history matching. The framework is evaluated using synthetic fractured geothermal reservoir cases with different observation types and monitoring configurations. The generative model reproduces the fracture statistics and associated flow and transport responses of the training realizations, and the surrogate provides accurate approximations of high-fidelity simulations. History matching reduces uncertainty in fracture geometries and reservoir predictions, with the extent of uncertainty reduction depending on the observation types and monitoring configurations. Combining temperature and tracer observations improves fracture characterization, and additional monitoring locations further constrain the posterior fracture distributions and corresponding predictions. The proposed framework can be extended to other fractured subsurface systems for characterization and uncertainty quantification.
\end{abstract}

\section*{Plain Language Summary}
Fractures in underground rock formations provide pathways for fluid flow and heat transport. Characterizing these fracture networks is important for geothermal energy production, oil and gas recovery, and groundwater management. However, their geometries are difficult to determine from limited observations. In this study, we develop a computational framework that combines machine learning and statistical methods to characterize fracture networks and quantify their uncertainty. The framework generates realistic fracture networks, efficiently predicts their flow and transport behavior, and uses temperature and tracer observations to constrain the possible fracture configurations. We evaluate the approach using simulated geothermal reservoir cases. The results show that combining different types of observations and increasing the number of monitoring locations can improve fracture characterization and reduce uncertainty in reservoir predictions. 

%%%%%%%%%%%%%%%%%%%%%%%%%%%%%%%%%%%%%%%%%%%%%%%
%
%  BODY TEXT
%
%%%%%%%%%%%%%%%%%%%%%%%%%%%%%%%%%%%%%%%%%%%%%%%

%%% Suggested section heads:
% \section{Introduction}
%
% The main text should start with an introduction. Except for short
% manuscripts (such as comments and replies), the text should be divided
% into sections, each with its own heading.

% Headings should be sentence fragments and do not begin with a
% lowercase letter or number. Examples of good headings are:

% \section{Materials and Methods}
% Here is text on Materials and Methods.
%
% \subsection{A descriptive heading about methods}
% More about Methods.
%
% \section{Data} (Or section title might be a descriptive heading about data)
%
% \section{Results} (Or section title might be a descriptive heading about the
% results)
%
% \section{Conclusions}

\section{Introduction} \label{sec:1_intro}

Fracture networks strongly influence fluid flow and transport in subsurface systems, including geothermal reservoirs, unconventional hydrocarbon reservoirs, and groundwater aquifers. Fracture geometry and connectivity control flow pathways and transport behavior, particularly in low-permeability formations where fractures provide the primary pathways for fluid movement. However, fracture networks are typically characterized by limited geological and geophysical observations. Uncertainty in fracture locations, lengths, orientations, and connectivity makes it difficult to construct and calibrate representative fracture network models.

Data assimilation enables fracture network characterization by conditioning model parameters on observations, including pressure, temperature, and tracer measurements. History matching of discrete fracture network (DFN) models presents two major challenges. First, direct adjustment of fracture geometries can produce unrealistic configurations, and changes in fracture connectivity can lead to highly nonlinear model responses. Second, repeated forward simulations are computationally expensive, particularly for Bayesian inversion. Addressing these challenges requires a parameterization that preserves realistic fracture geometries and an efficient approach for predicting flow and transport responses.

Deep generative models have been developed for fracture network parameterization and characterization. These models learn mappings from continuous latent variables to fracture geometries, allowing inversion to be performed in a lower-dimensional or more regular parameter space. \citeA{chen2023fracture} developed an approach based on variational autoencoders (VAEs) and generative adversarial networks (GANs) for stochastic fracture network inversion. \citeA{teng2024generative} introduced a latent diffusion model for generating DFNs, and \citeA{teng2025generating} developed a GAN-based approach that preserves prior fracture statistics and connectivity. \citeA{fan2026} applied latent diffusion parameterization to history matching of fractured geothermal reservoirs. Most of these approaches use image-based representations of fracture networks. Although these representations can reproduce fracture statistics and connectivity patterns, rasterization may introduce errors in fracture endpoints, orientations, and intersections. Direct representation of fracture coordinates avoids these discretization errors and provides an alternative for generative parameterization.

Numerical simulation of flow and transport in fractured systems is computationally demanding, particularly when many fracture network realizations are considered. Discrete fracture models (DFMs) typically require fracture-conforming grids that must be regenerated as fracture geometries change \cite{berre2019fractured}. Embedded discrete fracture models (EDFMs) represent fractures within a nonconforming background grid, avoiding repeated matrix-grid generation \cite{li2008efficient,moinfar2014development}. However, repeated EDFM simulations remain expensive for history matching. Deep-learning-based surrogate models have been developed to accelerate forward simulations. \citeA{kim2023fractured} developed a convolutional-recurrent neural network surrogate for predicting well responses in fractured reservoirs. \citeA{jiang2024fractured} and \citeA{al2025learning} applied graph neural networks (GNNs) to model flow and transport in fractured media. Despite these developments, predicting spatially and temporally varying responses for different fracture geometries remains challenging. Accurate predictions require representation of both local fracture-matrix interactions and long-range connectivity effects.

Various inversion methods have been applied to characterize fracture networks using pressure, tracer, and temperature observations. \citeA{somogyvari2017synthetic} applied reversible-jump Markov chain Monte Carlo (MCMC) to infer fracture geometries from tracer tomography data. \citeA{ringel2021stochastic} extended transdimensional inversion to three-dimensional fracture networks using hydraulic tomography. Ensemble-based methods have also been considered. \citeA{yao2018tuning} developed a Hough-transform-based parameterization and applied an iterative ensemble smoother to update fracture geometries from dynamic observations. \citeA{wu2024selecting} used ensemble-based inversion to estimate fracture properties in geothermal reservoirs, and \citeA{fan2026} combined generative parameterization with ensemble-based history matching. These methods rely on ensemble covariances to approximate relationships between parameters and observations, which can limit their ability to represent strongly non-Gaussian and multimodal posterior distributions. MCMC methods provide a more general approach to posterior sampling, but their computational cost can be substantial for high-dimensional fracture network inversion.

In this work, we develop a Bayesian inversion framework that combines generative model-based parameterization, geometry-aware surrogate modeling, and latent-space sampling. A vertex-based flow matching model maps Gaussian latent variables directly to fracture vertex coordinates, avoiding image-based representations. A geometry-aware surrogate model is developed using the EDFM discretization, combining local GNN message passing and global attention to predict spatially and temporally varying temperature and tracer concentration fields. The generative and surrogate models are integrated with parallel-tempering MCMC for posterior sampling in the latent space. We evaluate the framework using synthetic fractured geothermal reservoir cases. The generative and surrogate models are first evaluated against prior fracture statistics and high-fidelity simulation results. History matching is then performed to investigate the effects of observation types and monitoring configurations on fracture characterization and uncertainty reduction in reservoir predictions.

This paper proceeds as follows. Section~\ref{sec:2_flow_matching} describes the vertex-based flow matching model. Section~\ref{sec:3_surrogate} presents the EDFM discretization and geometry-aware surrogate model. Section~\ref{sec:4_inversion} describes the latent-space Bayesian inversion framework. Section~\ref{sec:5_results} presents the numerical setup, model evaluation, and history matching results. Conclusions and future work are provided in Section~\ref{sec:6_conclusions}.

\section{Vertex-based Flow Matching for Fracture Parameterization}
\label{sec:2_flow_matching}

This section presents the fracture network representation based on vertex coordinates and the flow matching model for generative parameterization of discrete fracture networks (DFNs).

\subsection{Fracture Network Representation}
\label{sec:2_1_DFN}

Discrete fracture networks (DFNs) are characterized by the locations, lengths, and orientations of individual fractures. These geometric parameters are typically sampled from prescribed statistical distributions, which may be non-Gaussian and exhibit dependencies due to geometric constraints. The resulting joint distribution can be difficult to represent and update directly during history matching. We therefore develop a generative model to represent DFN geometries through Gaussian latent variables.

A two-dimensional DFN consists of a fixed number of fractures, $N_{\mathrm{f}}$, with each fracture represented by a line segment defined by its two vertices, $(x_1^i,y_1^i,x_2^i,y_2^i)$, $i=1,\ldots,N_{\mathrm{f}}$. A DFN realization is represented by a vector $\mathbf{m}\in\mathbb{R}^{4N_{\mathrm{f}}}$ containing the fracture vertex coordinates. Different permutations of the fracture coordinates can represent the same DFN. To obtain a consistent representation for training, fractures are sorted in ascending order of their midpoint $y$-coordinates, with midpoint $x$-coordinates used to break ties. The two vertices of each fracture are ordered by their $y$-coordinates, with the smaller value first. Each component of the sorted coordinate vector is normalized using its mean and standard deviation over the training realizations. The resulting normalized vector, denoted by $\mbox{\boldmath$\xi$}_1\in\mathbb{R}^{4N_{\mathrm{f}}}$, defines the target distribution for flow matching.

\subsection{Vertex-Based Flow Matching Model}
\label{sec:2_2_flow_matching}

\begin{figure}[!ht]
    \centering
    \begin{minipage}{0.9\linewidth}
        \centering
        \includegraphics[width=\linewidth]{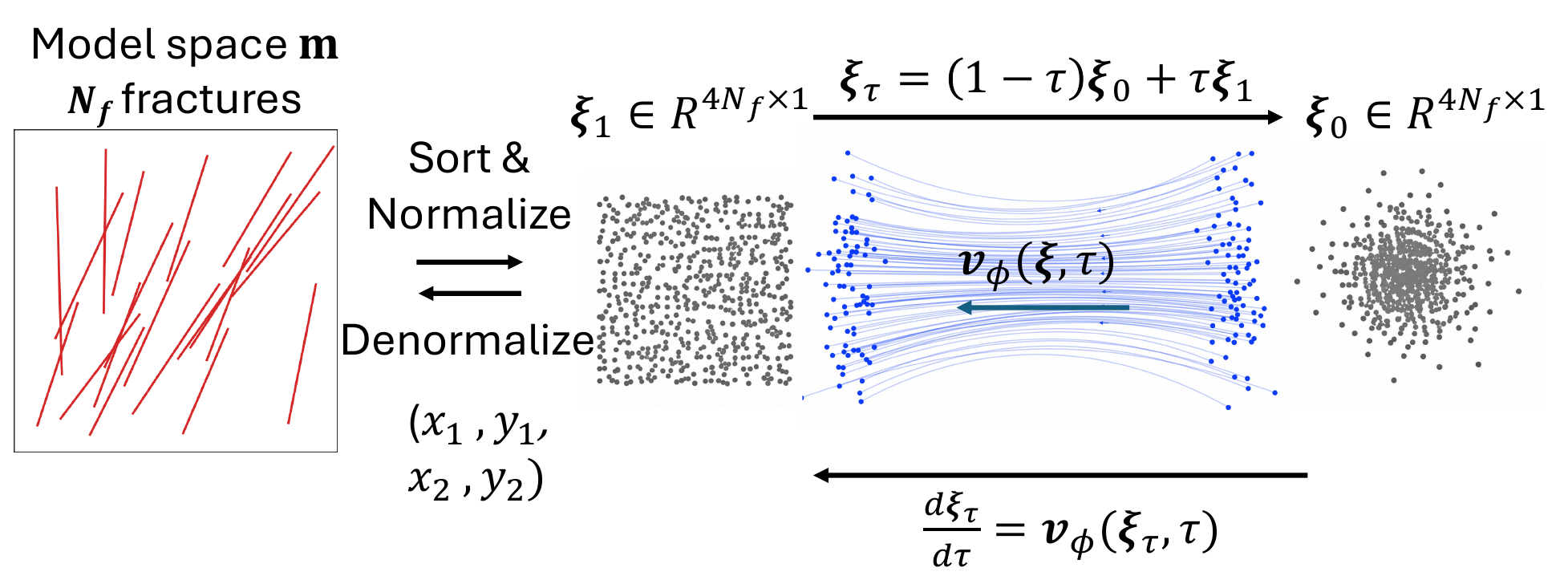}
        \subcaption{Vertex-based flow matching framework}
    \end{minipage}

    \begin{minipage}{\linewidth}
        \centering
        \includegraphics[width=0.9\linewidth]{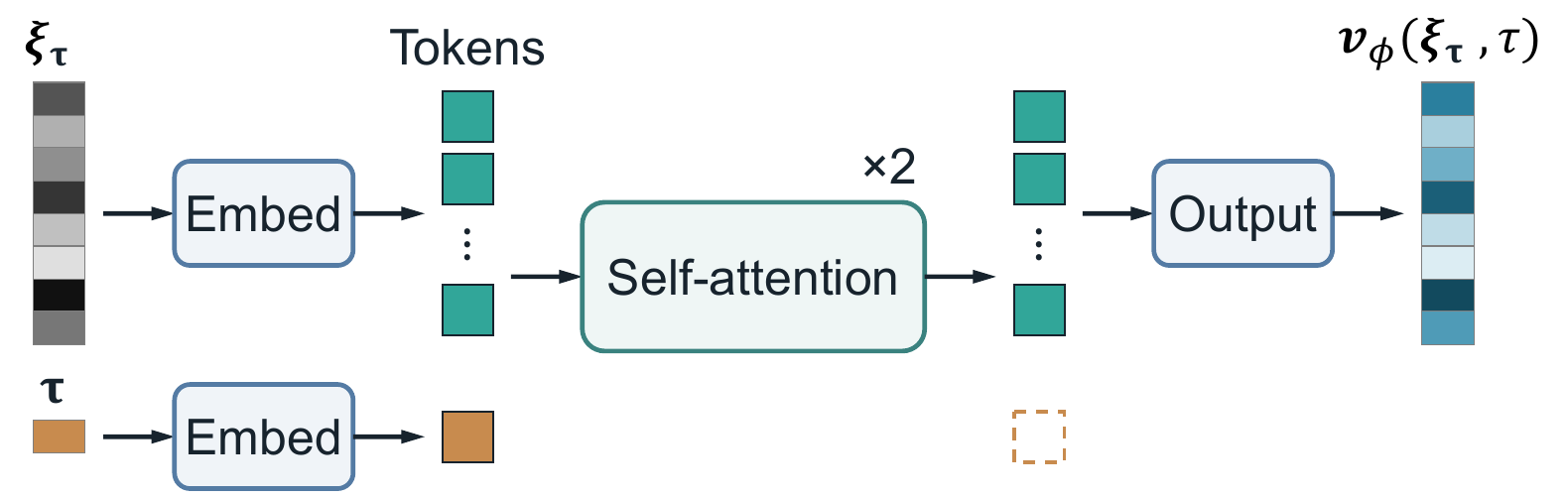}
        \subcaption{Velocity network architecture}
    \end{minipage}
    \caption{Vertex-based flow matching model for DFN parameterization. (a) Mapping from Gaussian latent variables $\mbox{\boldmath$\xi$}$ to DFN realizations $\mathbf{m}$ through the learned velocity field. (b) Transformer-based velocity network, where green and orange squares denote fracture tokens and the flow-time token, respectively. Self-attention captures interactions among fractures and incorporates flow-time information into the velocity prediction.}
    \label{fig:1_flow_matching}
\end{figure}

Flow matching is a generative modeling approach that learns a velocity field to transform samples from a prescribed distribution to a target distribution \cite{lipman2023flow,liu2023flow}. Here, flow matching is used to construct a mapping from Gaussian latent variables to fracture vertex coordinates, as illustrated in Fig.~\ref{fig:1_flow_matching}(a).

The flow matching model transforms samples from a standard Gaussian distribution, $\mbox{\boldmath$\xi$}_0=\mbox{\boldmath$\xi$}\sim\mathcal{N}(\mathbf{0},\mathbf{I})$, to the distribution of normalized fracture vertex coordinates $\mbox{\boldmath$\xi$}_1$. During training, independent samples of $\mbox{\boldmath$\xi$}_0$ and $\mbox{\boldmath$\xi$}_1$ are connected through a linear interpolation,
\begin{linenomath*}
\begin{equation}
\mbox{\boldmath$\xi$}_\tau=(1-\tau)\mbox{\boldmath$\xi$}_0+\tau\mbox{\boldmath$\xi$}_1,
\label{eq:fm_linear}
\end{equation}
\end{linenomath*}
where $\tau\in[0,1]$ is the flow-time variable and $\mbox{\boldmath$\xi$}_\tau$ denotes the intermediate state.

The velocity along each interpolation path is $\mbox{\boldmath$\xi$}_1-\mbox{\boldmath$\xi$}_0$. A transformer-based neural network \cite{vaswani2017attention}, denoted by $\mathbf{v}_\phi(\mbox{\boldmath$\xi$}_\tau,\tau)$, is used to approximate the velocity field, where $\phi$ denotes the trainable network parameters. As shown in Fig.~\ref{fig:1_flow_matching}(b), individual fractures are represented as tokens, and self-attention captures interactions among fractures. The network is trained by minimizing the flow matching loss,
\begin{linenomath*}
\begin{equation}
L(\phi)=\frac{1}{N_{\mathrm b}}\sum_{j=1}^{N_{\mathrm b}}
\left\|
\mathbf{v}_\phi(\mbox{\boldmath$\xi$}_\tau^{(j)},\tau^{(j)})
-
\left(\mbox{\boldmath$\xi$}_1^{(j)}-\mbox{\boldmath$\xi$}_0^{(j)}\right)
\right\|_2^2,
\label{eq:loss}
\end{equation}
\end{linenomath*}
where $N_{\mathrm b}$ is the minibatch size, $\tau^{(j)}$ is sampled uniformly from $[0,1]$, $\mbox{\boldmath$\xi$}_0^{(j)}$ is sampled from the standard Gaussian distribution, and $\mbox{\boldmath$\xi$}_1^{(j)}$ is sampled from the normalized training DFN realizations. The intermediate state $\mbox{\boldmath$\xi$}_\tau^{(j)}$ is calculated from Eq.~\ref{eq:fm_linear}. Details of the network architecture and training procedure are provided in Text~S1 and Table~S1 of the Supporting Information \cite{kingma2015adam,su2024roformer}.

After training, new DFN realizations are generated by integrating the learned velocity field. Starting from a Gaussian latent vector $\mbox{\boldmath$\xi$}\sim\mathcal{N}(\mathbf{0},\mathbf{I})$, the following ordinary differential equation (ODE) is integrated from $\tau=0$ to $\tau=1$:
\begin{linenomath*}
\begin{equation}
\frac{d\mbox{\boldmath$\xi$}_\tau}{d\tau}
=
\mathbf{v}_{\phi^*}(\mbox{\boldmath$\xi$}_\tau,\tau),
\qquad
\mbox{\boldmath$\xi$}_{\tau=0}=\mbox{\boldmath$\xi$},
\label{eq:ode}
\end{equation}
\end{linenomath*}
where $\phi^*$ denotes the trained network parameters. The terminal state $\mbox{\boldmath$\xi$}_{\tau=1}$ is denormalized using the training statistics to obtain the physical fracture vertex coordinates. This defines a deterministic nonlinear mapping $\mathbf{m}=G(\mbox{\boldmath$\xi$})$ from Gaussian latent variables to DFN realizations. The mapping provides a generative parameterization of the fracture geometry prior, allowing Bayesian history matching to be performed in the latent space with a standard Gaussian prior.

\section{Geometry-Aware Surrogate Model}
\label{sec:3_surrogate}

This section describes the embedded discrete fracture model (EDFM) used for flow and transport simulations and the geometry-aware surrogate model developed to predict reservoir responses for different fracture network realizations.

\subsection{Embedded Discrete Fracture Model}
\label{sec:3_1_edfm}

The embedded discrete fracture model (EDFM) \cite{li2008efficient,moinfar2014development} represents fractures within a nonconforming background grid, avoiding the need for fracture-conforming remeshing as fracture geometries change. This makes EDFM particularly suitable for fracture network history matching, which requires repeated forward simulations for different DFN realizations.

The discretization consists of $N_{\mathrm{c}}$ matrix cells and $N_{\mathrm{fe}}$ fracture elements, with fractures partitioned at matrix-cell boundaries. The matrix cells and fracture elements are treated as control volumes, indexed by $i,j=1,\ldots,N_{\mathrm{c}}+N_{\mathrm{fe}}$. Flow between connected control volumes $i$ and $j$ is expressed as
\begin{linenomath*}
\begin{equation}
q_{ij}=\lambda T_{ij}(P_i-P_j),
\label{eq:darcy}
\end{equation}
\end{linenomath*}
where $\lambda$ is the fluid mobility, $T_{ij}$ is the transmissibility determined by the connection geometry and permeability, and $P_i$ and $P_j$ are the pressures of the two control volumes.

Four types of connections are considered: matrix-matrix, matrix-fracture, along-fracture, and fracture-intersection connections. These describe flow between neighboring matrix cells, between matrix cells and fracture elements, along individual fractures, and between intersecting fractures, respectively. The EDFM discretization also defines the graph used by the surrogate model, with matrix cells and fracture elements represented as nodes and their connections represented as edges. The resulting graph topology and geometric features are determined by each DFN realization.

\subsection{Geometry-Aware Surrogate Model}
\label{sec:3_2_architecture}

A geometry-aware surrogate model is developed to approximate the flow and transport responses for different DFN realizations. For a fracture network $\mathbf{m}\in\mathbb{R}^{4N_{\mathrm{f}}}$, the surrogate is expressed as
\begin{linenomath*}
\begin{equation}
\hat{\mathbf{d}}=\hat{f}(\mathbf{m};\mathbf{w}),
\label{eq:surrogate}
\end{equation}
\end{linenomath*}
where $\hat{\mathbf{d}}\in\mathbb{R}^{N_{\mathrm{d}}}$ is the predicted response vector and $\mathbf{w}$ denotes the trained network parameters. Separate networks with the same architecture are trained for temperature and tracer concentration. Their outputs are denoted by $\hat{\mathbf{T}}\in\mathbb{R}^{N_{\mathrm{c}}\times N_{\mathrm{t},T}}$ and $\hat{\mathbf{C}}\in\mathbb{R}^{N_{\mathrm{c}}\times N_{\mathrm{t},C}}$, where $N_{\mathrm{t},T}$ and $N_{\mathrm{t},C}$ denote the corresponding numbers of prediction times. The predicted fields are assembled into the response vector used for data assimilation in Section~\ref{sec:4_inversion}.

For each DFN realization, the fracture geometry $\mathbf{m}$ is used to construct the EDFM graph described in Section~\ref{sec:3_1_edfm}. The graph topology and geometric features are determined by the fracture geometry, and physical properties are incorporated into the node and edge features. The input feature vectors for matrix-cell nodes, fracture-element nodes, and directed edges are defined as
\begin{linenomath*}
\begin{equation}
\begin{aligned}
\mathbf{f}_i^{\mathrm{c}} &=
[x_i,\ y_i,\ k_i,\ V_i]^{\mathsf{T}},\\
\mathbf{f}_i^{\mathrm{fe}} &=
[x_{i,1},\ y_{i,1},\ x_{i,2},\ y_{i,2},\ k_i,\ b_i,\ V_i]^{\mathsf{T}},\\
\mathbf{f}_{ij}^{\mathrm{edge}} &=
[\Delta x_{ij},\ \Delta y_{ij},\ d_i,\ d_j,\ A_{ij},\ k_i,\ k_j,\ \mathbf{c}_{ij}^{\mathsf{T}}]^{\mathsf{T}},
\end{aligned}
\label{eq:graph_features}
\end{equation}
\end{linenomath*}
where $(x_i,y_i)$ are the center coordinates of matrix cell $i$, and $(x_{i,1},y_{i,1})$ and $(x_{i,2},y_{i,2})$ are the endpoints of fracture element $i$. The quantities $k_i$, $b_i$, and $V_i$ denote permeability, fracture aperture, and control-volume volume, respectively. For an edge connecting nodes $i$ and $j$, $(\Delta x_{ij},\Delta y_{ij})$ denotes their relative position, $d_i$ and $d_j$ are the distances from the control volumes to the connection interface, and $A_{ij}$ is the connection area. The vector $\mathbf{c}_{ij}\in\mathbb{R}^{4}$ is a one-hot encoding of the connection type.

\begin{figure}[!htp]
    \centering
    \begin{minipage}{\linewidth}
        \centering
        \includegraphics[width=\linewidth]{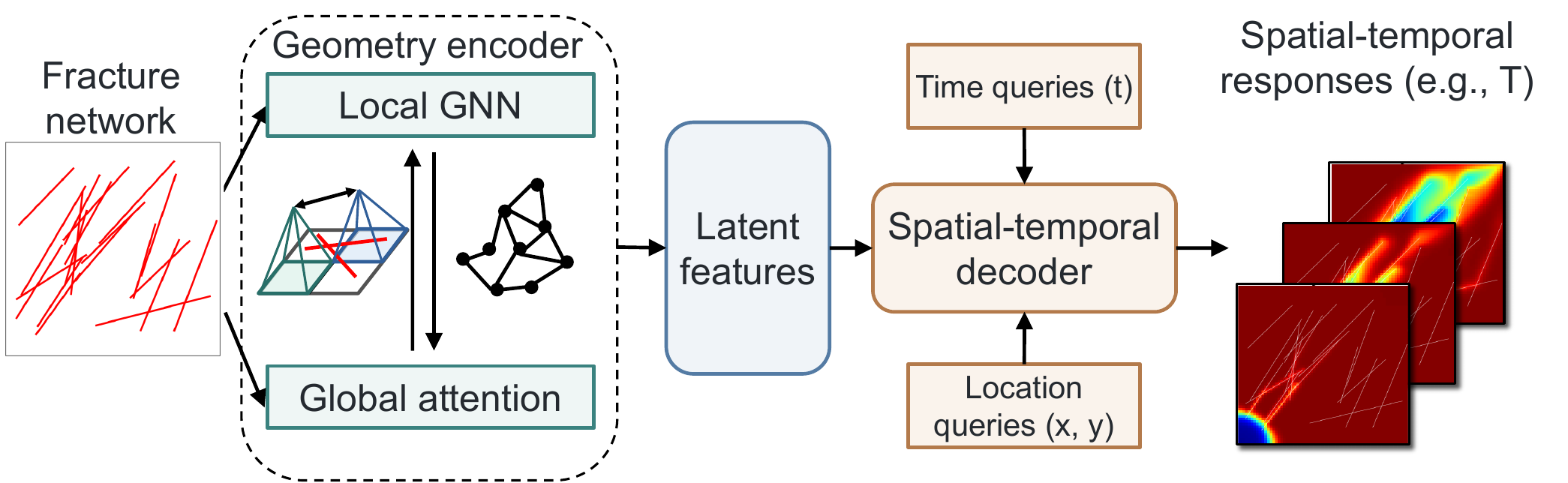}
        \subcaption{Overall architecture}
    \end{minipage}
    \begin{minipage}{\linewidth}
        \centering
        \includegraphics[width=\linewidth]{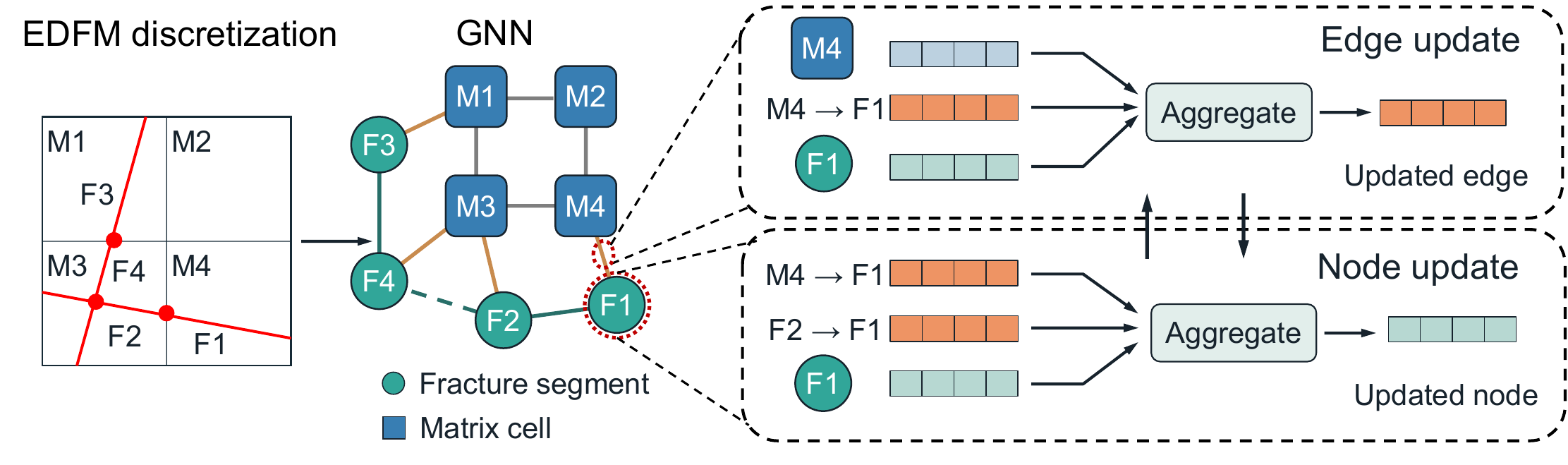}
        \subcaption{Local GNN}
    \end{minipage}
    \begin{minipage}{\linewidth}
        \centering
        \includegraphics[width=\linewidth]{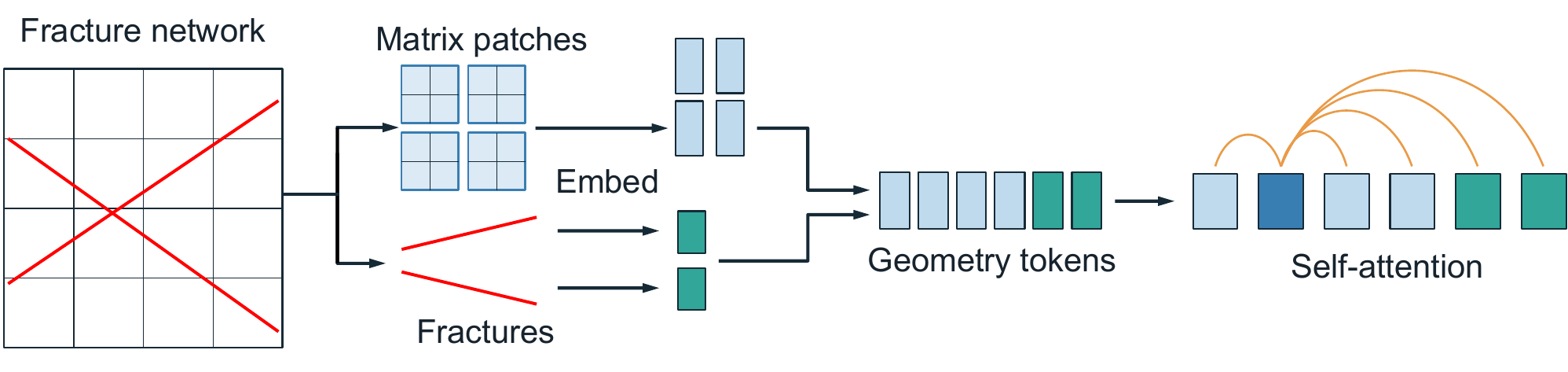}
        \subcaption{Global attention}
    \end{minipage}
    \begin{minipage}{\linewidth}
        \centering
        \includegraphics[width=\linewidth]{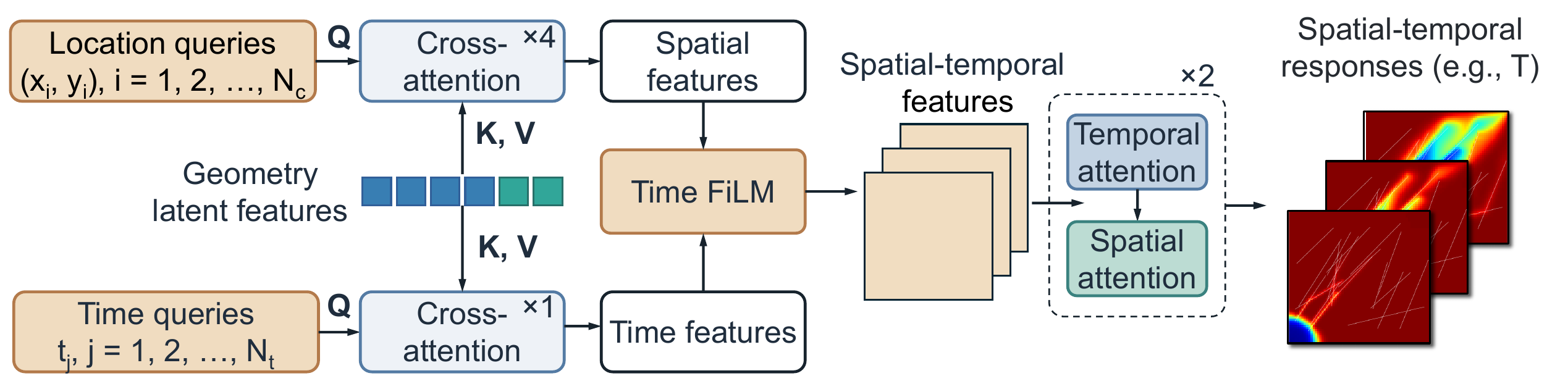}
        \subcaption{Spatial-temporal decoder}
    \end{minipage}
    \caption{Geometry-aware surrogate model. (a) Overall architecture consisting of a geometry encoder and a spatial-temporal decoder. (b) Local GNN for message passing along EDFM connections. Gray, orange, solid green, and dashed green edges represent matrix-matrix, matrix-fracture, along-fracture, and fracture-intersection connections, respectively. (c) Global attention for information exchange between matrix regions and fractures. (d) Spatial-temporal decoder for predicting reservoir responses at specified locations and times.}
    \label{fig:2_surrogate}
\end{figure}

The local GNN updates node and edge features through message passing along EDFM connections (Fig.~\ref{fig:2_surrogate}b). The input features are first mapped to learned representations. Let $\mathbf{h}_i^{(l)}$ and $\mathbf{e}_{ij}^{(l)}$ denote the latent features of node $i$ and the directed edge from node $i$ to node $j$ at layer $l$, respectively. Following the graph network formulation of \citeA{battaglia2018relational}, the edge features are updated as
\begin{linenomath*}
\begin{equation}
\mathbf{e}_{ij}^{(l+1)}
=
\mathbf{e}_{ij}^{(l)}
+
\mathrm{MLP}_{\mathrm{e}}
\left(
[\mathbf{h}_i^{(l)},\mathbf{h}_j^{(l)},\mathbf{e}_{ij}^{(l)}]
\right),
\label{eq:edge}
\end{equation}
\end{linenomath*}
where $\mathrm{MLP}_{\mathrm{e}}$ is a multilayer perceptron (MLP) and the brackets denote concatenation. The node features are then updated using the mean of the incoming edge features,
\begin{linenomath*}
\begin{equation}
\mathbf{h}_j^{(l+1)}
=
\mathbf{h}_j^{(l)}
+
\mathrm{MLP}_{\mathrm{h}}
\left(
[\mathbf{h}_j^{(l)},\bar{\mathbf{e}}_j^{(l+1)}]
\right),
\label{eq:node}
\end{equation}
\end{linenomath*}
where $\bar{\mathbf{e}}_j^{(l+1)}$ is the mean of the updated features of edges directed toward node $j$, and $\mathrm{MLP}_{\mathrm{h}}$ is the node-update network. Repeated message passing propagates information through connected matrix cells and fracture elements.

The global attention module represents interactions between matrix regions and fractures beyond local EDFM connections (Fig.~\ref{fig:2_surrogate}c). Matrix cells are grouped into nonoverlapping $4\times4$ patches, and their features are aggregated through a convolutional layer to form matrix-patch latent features. Fracture-element features are aggregated through learned weighted pooling to obtain a latent feature vector for each fracture. The resulting representation contains $N_{\mathrm{p}}$ matrix-patch features and $N_{\mathrm{f}}$ fracture features, where $N_{\mathrm{p}}$ is the number of matrix patches. Type embeddings distinguish the two feature types, and positional embeddings are added to the matrix-patch features. Denoting the combined latent feature matrix by $Z$, the self-attention operation is
\begin{linenomath*}
\begin{equation}
\mathrm{Attention}(Z)
=
\mathrm{softmax}
\left(
\frac{ZW_Q(ZW_K)^{\mathsf{T}}}{\sqrt{d_{\mathrm{k}}}}
\right)ZW_V,
\label{eq:attention}
\end{equation}
\end{linenomath*}
where $W_Q$, $W_K$, and $W_V$ are learned projection matrices, and $d_{\mathrm{k}}$ is the query and key dimension. Self-attention enables information exchange between matrix regions and fractures, including those not directly connected in the EDFM graph. The fracture latent features are additionally processed through self-attention with pairwise geometric bias to account for relative fracture positions, orientations, and intersections. The geometric bias is described in Text~S2 of the Supporting Information.

The geometry encoder contains four rounds of local GNN and global attention updates. Each round includes four GNN layers followed by global attention and information exchange between graph nodes and latent features. The node features are aggregated into the corresponding matrix-patch and fracture representations, and the updated latent features are mapped back to the graph nodes. A final node-to-feature update is applied after the fourth round. The resulting geometry features contain information from both local EDFM connections and global interactions.

The spatial-temporal decoder predicts reservoir responses at specified matrix-cell locations and times (Fig.~\ref{fig:2_surrogate}d). Matrix-cell coordinates and prediction times are embedded separately to form spatial and temporal queries. Cross-attention with the encoded geometry features produces a spatial feature vector $\mathbf{s}_i$ for matrix cell $i$ and a temporal context vector $\mathbf{r}_j$ for time $t_j$. Let $\mathbf{u}_j$ denote the temporal query embedding. The spatial and temporal features are combined using feature-wise linear modulation (FiLM) \cite{perez2018film},
\begin{linenomath*}
\begin{equation}
\mathbf{a}_{ij}
=
(1+\mbox{\boldmath$\gamma$}_j)\odot\mathbf{s}_i
+
\mbox{\boldmath$\delta$}_j
+
\mathbf{u}_j,
\label{eq:film}
\end{equation}
\end{linenomath*}
where $\mbox{\boldmath$\gamma$}_j$ and $\mbox{\boldmath$\delta$}_j$ are scale and shift vectors computed from $\mathbf{u}_j$ and $\mathbf{r}_j$, $\odot$ denotes element-wise multiplication, and $\mathbf{a}_{ij}$ is the resulting feature vector for cell $i$ at time $t_j$. Separate spatial and temporal queries reduce the number of cross-attention queries from $N_{\mathrm{c}}N_{\mathrm{t}}$ to $N_{\mathrm{c}}+N_{\mathrm{t}}$, where $N_{\mathrm{t}}$ is the number of prediction times.

The combined features are processed through two spatial-temporal attention blocks, each consisting of causal temporal attention followed by spatial attention. Temporal attention operates across prediction times at each matrix cell, with each time step attending only to itself and preceding times. Spatial attention exchanges information between matrix patches at each time step. The resulting features are mapped to the response fields through a pointwise MLP and a convolutional correction. A softplus activation is applied to the tracer output to obtain nonnegative concentration predictions.

The surrogate networks are trained by minimizing
\begin{linenomath*}
\begin{equation}
L=L_{\mathrm{d}}+w_1L_{\mathrm{s}}+w_2L_{\mathrm{t}},
\label{eq:surrogate_loss}
\end{equation}
\end{linenomath*}
where $L_{\mathrm{d}}$ measures the mismatch between simulated and predicted responses, and $L_{\mathrm{s}}$ and $L_{\mathrm{t}}$ measure differences in their spatial gradients and temporal derivatives, respectively. Both weights are set to $w_1=w_2=0.25$. The data loss uses relative absolute error for temperature and mean squared error for tracer concentration, with an additional term for monitoring locations in the tracer loss. The network parameters corresponding to the lowest validation loss are retained. Additional architecture details, loss formulations, and training settings are provided in Text~S2 and Tables~S2-S5 of the Supporting Information.
\section{Latent-Space Bayesian Inversion}
\label{sec:4_inversion}

This section presents the Bayesian inversion framework that integrates the flow matching parameterization and geometry-aware surrogate model described in Sections~\ref{sec:2_flow_matching} and \ref{sec:3_surrogate}.

\subsection{Bayesian Formulation}
\label{sec:4_1_bayesian}

Fracture network characterization is formulated as a Bayesian inverse problem using observations of reservoir responses. The full response vector is denoted by $\mathbf{d}\in\mathbb{R}^{N_{\mathrm{d}}}$, where $N_{\mathrm{d}}$ is the total number of response values over the spatial locations and prediction times considered. The observed data $\mathbf{d}_{\mathrm{obs}}\in\mathbb{R}^{N_{\mathrm{obs}}}$ are expressed as
\begin{linenomath*}
\begin{equation}
\mathbf{d}_{\mathrm{obs}}
=
H\mathbf{d}_{\mathrm{true}}
+
\mbox{\boldmath$\epsilon$},
\label{eq:observations}
\end{equation}
\end{linenomath*}
where $\mathbf{d}_{\mathrm{true}}\in\mathbb{R}^{N_{\mathrm{d}}}$ is the full response vector of the true system, $H\in\mathbb{R}^{N_{\mathrm{obs}}\times N_{\mathrm{d}}}$ is an extraction matrix that selects responses at the observation locations and times, and $\mbox{\boldmath$\epsilon$}\in\mathbb{R}^{N_{\mathrm{obs}}}$ represents measurement error.

The fracture geometry is parameterized using the Gaussian latent variables $\mbox{\boldmath$\xi$}$ introduced in Section~\ref{sec:2_flow_matching}, with $\mathbf{m}=G(\mbox{\boldmath$\xi$})$. The geometry-aware surrogate predicts the full response vector as $\hat{\mathbf{d}}(\mbox{\boldmath$\xi$})=\hat{f}(G(\mbox{\boldmath$\xi$});\mathbf{w})$, where $\mathbf{w}$ denotes the trained network parameters. The predicted observations are obtained as
\begin{linenomath*}
\begin{equation}
\hat{\mathbf{d}}_{\mathrm{obs}}(\mbox{\boldmath$\xi$})
=
H\hat{\mathbf{d}}(\mbox{\boldmath$\xi$}),
\label{eq:pred_data}
\end{equation}
\end{linenomath*}
where $\hat{\mathbf{d}}_{\mathrm{obs}}\in\mathbb{R}^{N_{\mathrm{obs}}}$ contains the surrogate predictions at the observation locations and times.

The posterior distribution of the latent variables conditioned on the observations follows Bayes' theorem,
\begin{linenomath*}
\begin{equation}
p(\mbox{\boldmath$\xi$}\mid\mathbf{d}_{\mathrm{obs}})
=
\frac{
p(\mbox{\boldmath$\xi$})
p(\mathbf{d}_{\mathrm{obs}}\mid\mbox{\boldmath$\xi$})
}{
p(\mathbf{d}_{\mathrm{obs}})
},
\label{eq:bayes}
\end{equation}
\end{linenomath*}
where $p(\mbox{\boldmath$\xi$})$ is the prior, $p(\mathbf{d}_{\mathrm{obs}}\mid\mbox{\boldmath$\xi$})$ is the likelihood, and $p(\mathbf{d}_{\mathrm{obs}})$ is the normalization constant. The latent variables follow a standard Gaussian prior, $\mbox{\boldmath$\xi$}\sim\mathcal{N}(\mathbf{0},\mathbf{I})$.

Measurement and surrogate model errors are assumed to be independent, zero-mean Gaussian variables, with covariance matrices $C_{\mathrm{D}}$ and $C_{\mathrm{sur}}$, respectively. The total error covariance is $C_{\mathrm{tot}}=C_{\mathrm{D}}+C_{\mathrm{sur}}$. The likelihood is given by
\begin{linenomath*}
\begin{equation}
p(\mathbf{d}_{\mathrm{obs}}\mid\mbox{\boldmath$\xi$})
\propto
\exp\left[
-\frac{1}{2}
\left(
\mathbf{d}_{\mathrm{obs}}
-
\hat{\mathbf{d}}_{\mathrm{obs}}(\mbox{\boldmath$\xi$})
\right)^{\mathsf{T}}
C_{\mathrm{tot}}^{-1}
\left(
\mathbf{d}_{\mathrm{obs}}
-
\hat{\mathbf{d}}_{\mathrm{obs}}(\mbox{\boldmath$\xi$})
\right)
\right],
\label{eq:likelihood}
\end{equation}
\end{linenomath*}
where $C_{\mathrm{tot}}\in\mathbb{R}^{N_{\mathrm{obs}}\times N_{\mathrm{obs}}}$. Correlations between observations are neglected, resulting in a diagonal covariance matrix with entries $\sigma_i^2=\sigma_{\mathrm{meas},i}^2+\sigma_{\mathrm{sur},i}^2$. Other sources of model error are not considered.

\subsection{Data Assimilation Workflow}
\label{sec:4_2_workflow}

\begin{figure}[!ht]
    \centering
    \includegraphics[width=\textwidth]{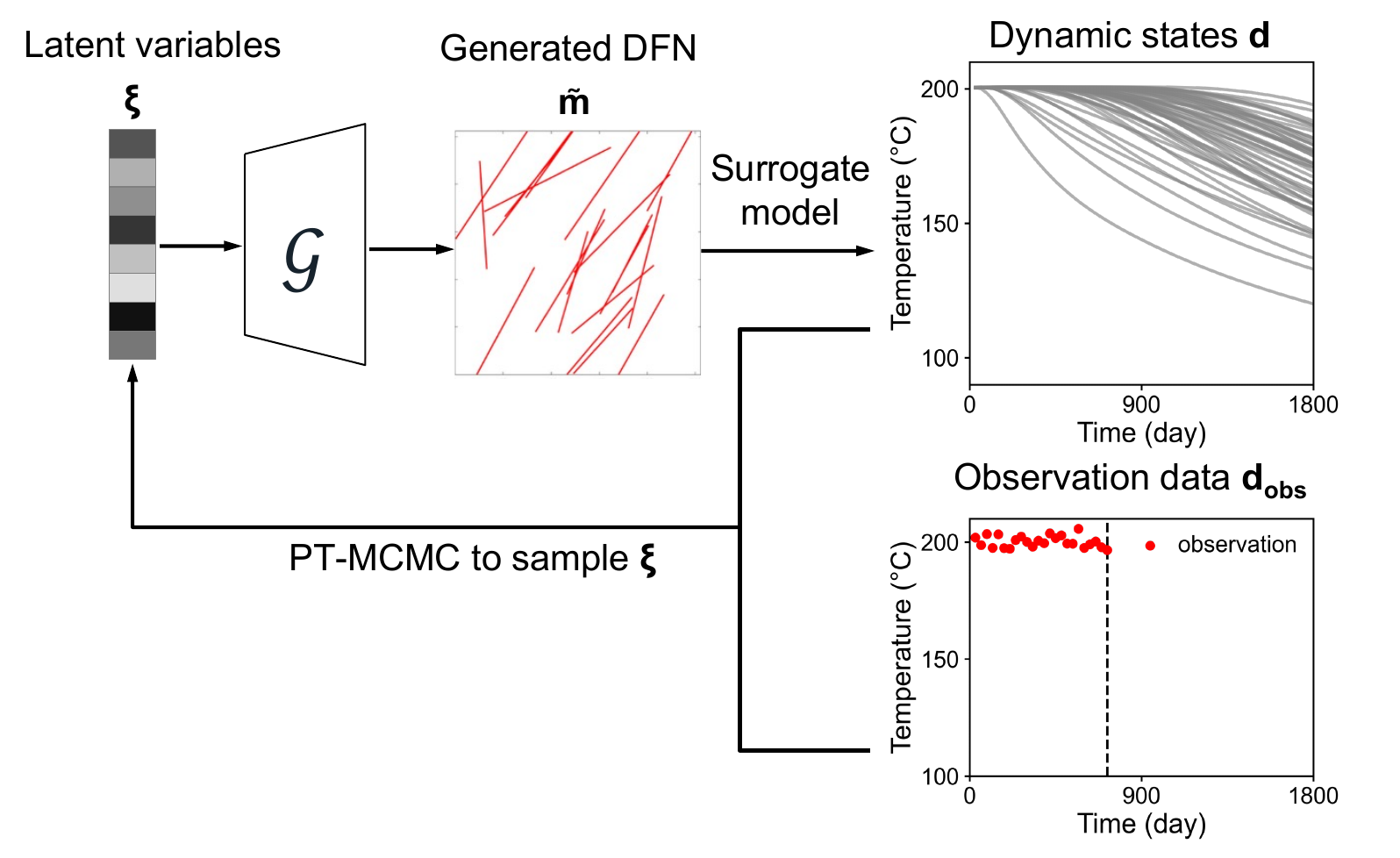}
    \caption{Workflow for latent-space Bayesian inversion. Gaussian latent variables are mapped to DFN realizations using the flow matching model, and the geometry-aware surrogate predicts the corresponding reservoir responses. The predicted observations are used to evaluate the likelihood for posterior sampling. Posterior realizations are used to quantify uncertainty in fracture geometries and reservoir responses.}
    \label{fig:3_history_matching_workflow}
\end{figure}

The data assimilation workflow is illustrated in Fig.~\ref{fig:3_history_matching_workflow}. The flow matching and geometry-aware surrogate models are trained separately and integrated for latent-space inversion. For each proposed latent vector, the flow matching model generates a DFN realization, and the surrogate predicts the corresponding reservoir responses. The predicted observations and likelihood are evaluated using Eqs.~\ref{eq:pred_data} and \ref{eq:likelihood}, respectively. This procedure enables repeated forward evaluations without additional high-fidelity simulations.

Markov chain Monte Carlo (MCMC) is used to sample the posterior distribution of the latent variables. At each iteration, a new latent vector is proposed and accepted or rejected according to the Metropolis-Hastings criterion \cite{hastings1970monte}. Given the standard Gaussian prior, we use the preconditioned Crank-Nicolson (pCN) proposal \cite{cotter2013mcmc},
\begin{linenomath*}
\begin{equation}
\mbox{\boldmath$\xi$}'
=
\sqrt{1-\beta^2}\,
\mbox{\boldmath$\xi$}^{(n)}
+
\beta\mbox{\boldmath$\eta$},
\label{eq:pcn}
\end{equation}
\end{linenomath*}
where $\mbox{\boldmath$\xi$}^{(n)}$ and $\mbox{\boldmath$\xi$}'$ are the current and proposed latent vectors, respectively, $\mbox{\boldmath$\eta$}\sim\mathcal{N}(\mathbf{0},\mathbf{I})$ is an independent Gaussian vector, and $\beta\in(0,1]$ controls the proposal step size. The pCN proposal preserves the Gaussian prior, such that the Metropolis-Hastings acceptance probability depends only on the likelihood ratio.

The posterior distribution may contain multiple modes due to the nonuniqueness of fracture network inversion, making sampling with a single MCMC chain inefficient. Parallel tempering (PT) \cite{geyer1991markov,earl2005parallel} is therefore employed to improve sampling across different posterior modes. PT considers $N_\omega$ inverse-temperature levels, each targeting a tempered distribution,
\begin{linenomath*}
\begin{equation}
p_\ell(\mbox{\boldmath$\xi$}\mid\mathbf{d}_{\mathrm{obs}})
\propto
p(\mbox{\boldmath$\xi$})
\left[
p(\mathbf{d}_{\mathrm{obs}}\mid\mbox{\boldmath$\xi$})
\right]^{\omega_\ell},
\qquad
\ell=1,\ldots,N_\omega,
\label{eq:tempered}
\end{equation}
\end{linenomath*}
where $1=\omega_1>\omega_2>\cdots>\omega_{N_\omega}>0$ are the inverse temperatures. The first level targets the original posterior, and the remaining levels target progressively flatter distributions that allow the chains to move more readily between regions of high posterior probability. Each level performs pCN updates with a level-specific step size $\beta_\ell$. Periodic exchanges of states between neighboring levels allow the first level to sample different posterior modes.

Samples from the level with $\omega_1=1$ are retained for posterior characterization. These samples are mapped to DFN realizations, and the corresponding reservoir responses are predicted using the geometry-aware surrogate. The ensemble smoother with multiple data assimilation (ESMDA) \cite{emerick2013ensemble} is also considered for comparison. Implementation details and the comparison between PT-MCMC and ESMDA are provided in Texts~S3 and S4 of the Supporting Information, respectively.

\section{Model Evaluation and History Matching Results}
\label{sec:5_results}

This section presents the numerical setup, evaluates the flow matching and geometry-aware surrogate models, and investigates the effects of observation types and monitoring configurations on fracture network history matching. The performance is also assessed for multiple true DFN realizations.

\subsection{Problem Setup}
\label{sec:5_1_setup}

The proposed framework is evaluated using a synthetic enhanced geothermal system (EGS) with uncertain fracture geometries. The two-dimensional model domain is $256~\mathrm{m}\times256~\mathrm{m}$. Each discrete fracture network (DFN) contains $N_{\mathrm{f}}=20$ fractures, characterized by their lengths $l$, orientations $\theta$, and midpoint locations $(x_m,y_m)$. Fracture lengths follow a truncated power-law distribution \cite{dedreuzy2002},
\begin{linenomath*}
\begin{equation}
p(l)\propto l^{-a},\qquad l_{\min}\leq l\leq l_{\max},
\label{eq:length}
\end{equation}
\end{linenomath*}
where $a$ is the scaling exponent, and $l_{\min}$ and $l_{\max}$ are the minimum and maximum fracture lengths. We set $a=2.34$, $l_{\min}=89.6~\mathrm{m}$, and $l_{\max}=153.6~\mathrm{m}$. Fracture orientations follow a Gaussian distribution with a mean of $60^\circ$ and a standard deviation of $15^\circ$. Given the sampled length and orientation, the fracture midpoint is sampled uniformly within
$l|\cos\theta|/2\leq x_m\leq256-l|\cos\theta|/2$ and
$l|\sin\theta|/2\leq y_m\leq256-l|\sin\theta|/2$,
ensuring that the entire fracture lies within the domain. Fractures are generated independently. A total of 5,000 DFN realizations are generated for flow matching training, with the corresponding simulation results used for surrogate training and validation.

\begin{figure}[!ht]
    \centering
    \begin{minipage}[t]{0.32\textwidth}
        \centering
        \makebox[\linewidth][c]{\includegraphics[width=1.05\linewidth]{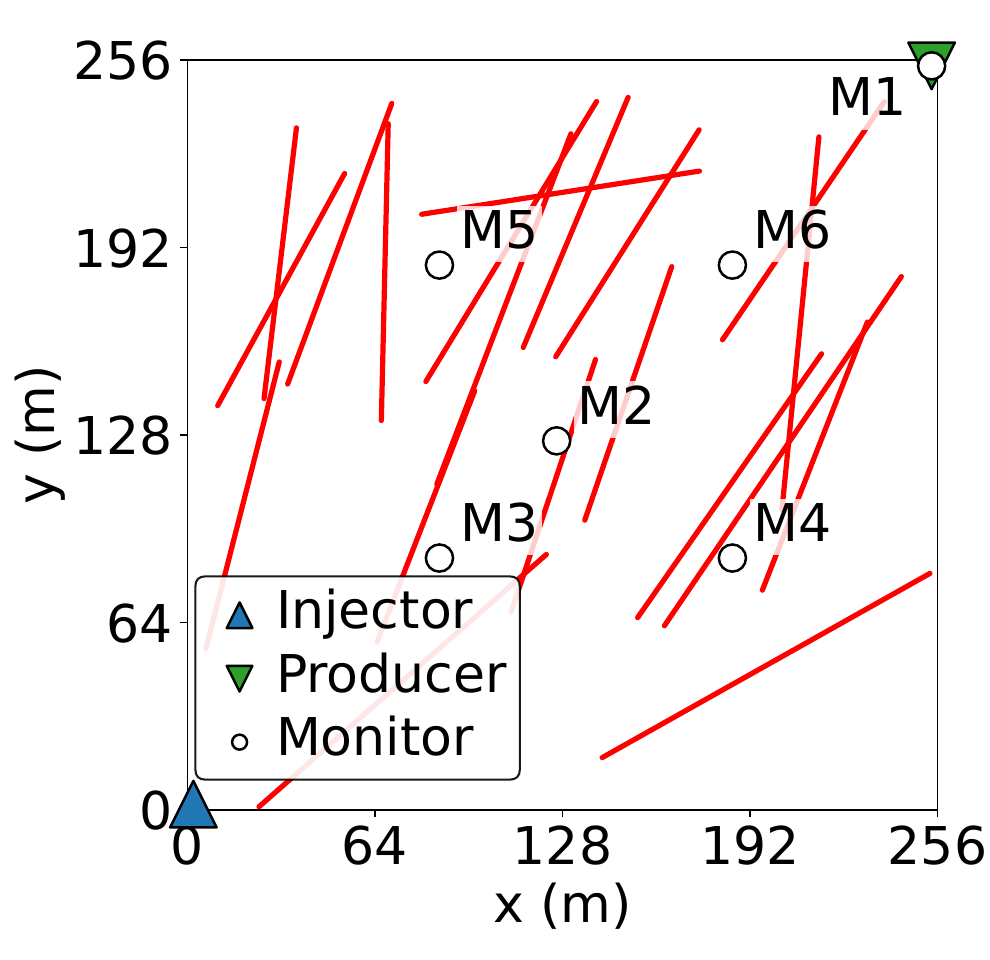}}
        \subcaption{True DFN}
    \end{minipage}\hfill
    \begin{minipage}[t]{0.32\textwidth}
        \centering
        \makebox[\linewidth][c]{\includegraphics[width=1.05\linewidth]{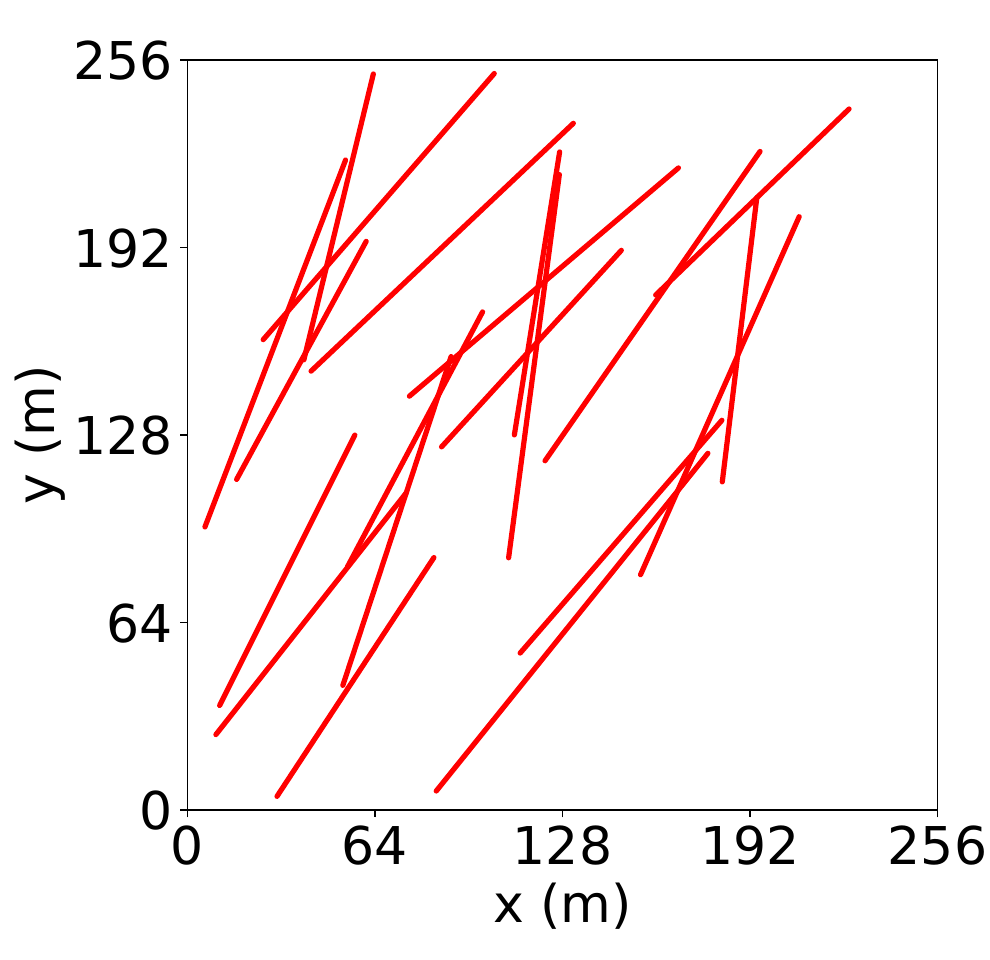}}
        \subcaption{Training realization 1}
    \end{minipage}\hfill
    \begin{minipage}[t]{0.32\textwidth}
        \centering
        \makebox[\linewidth][c]{\includegraphics[width=1.05\linewidth]{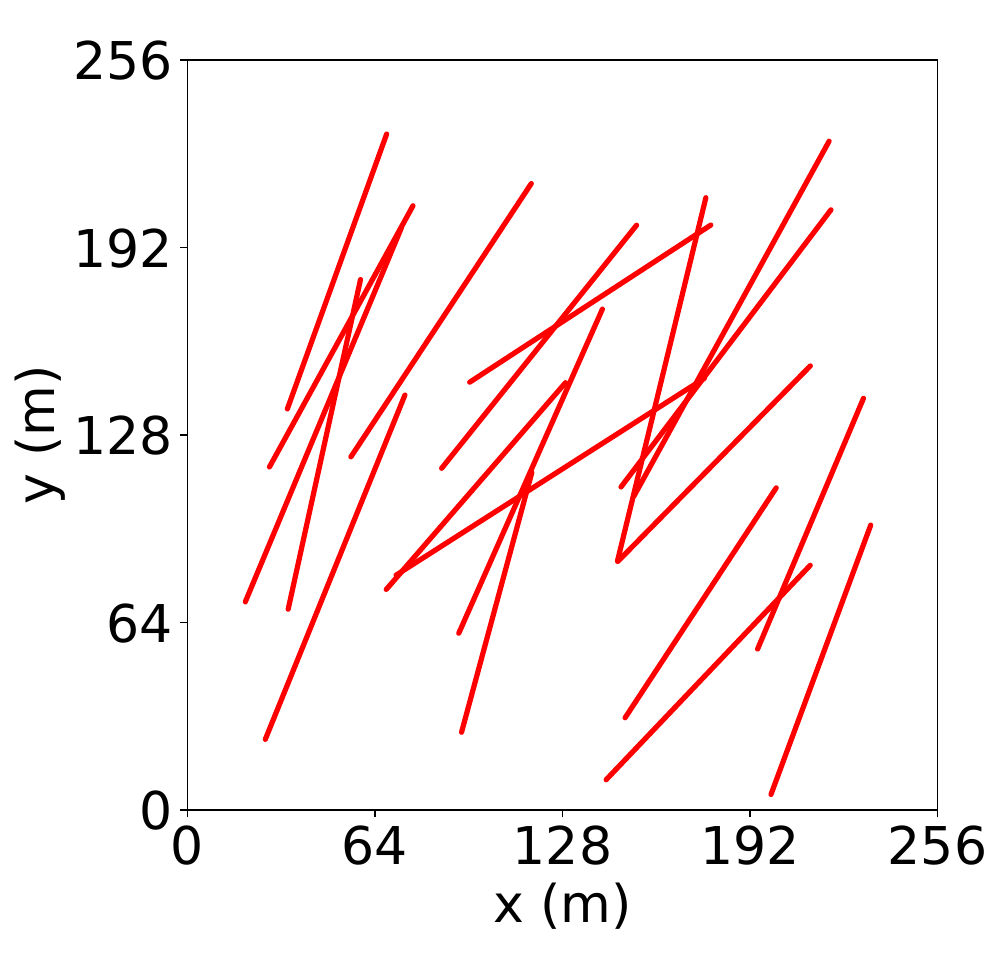}}
        \subcaption{Training realization 2}
    \end{minipage}

    \begin{minipage}[t]{0.32\textwidth}
        \centering
        \makebox[\linewidth][c]{\includegraphics[width=1.05\linewidth]{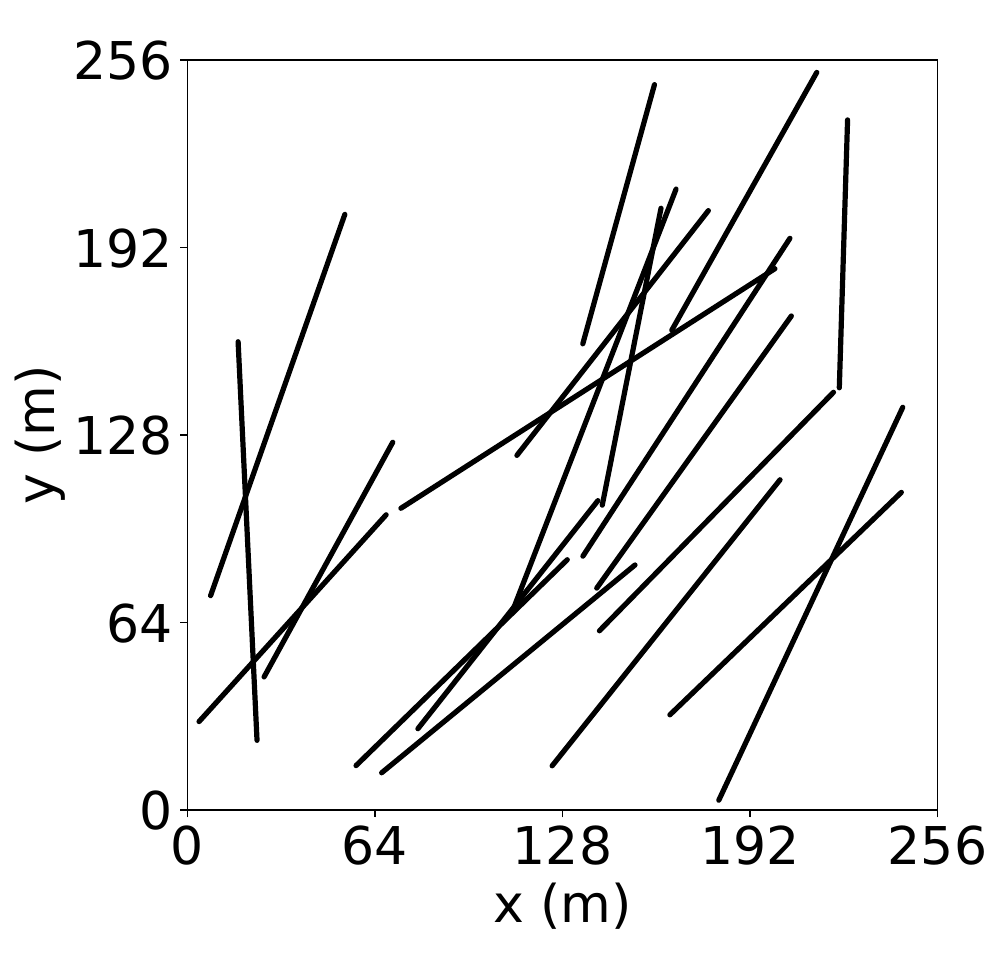}}
        \subcaption{Generated realization 1}
    \end{minipage}\hfill
    \begin{minipage}[t]{0.32\textwidth}
        \centering
        \makebox[\linewidth][c]{\includegraphics[width=1.05\linewidth]{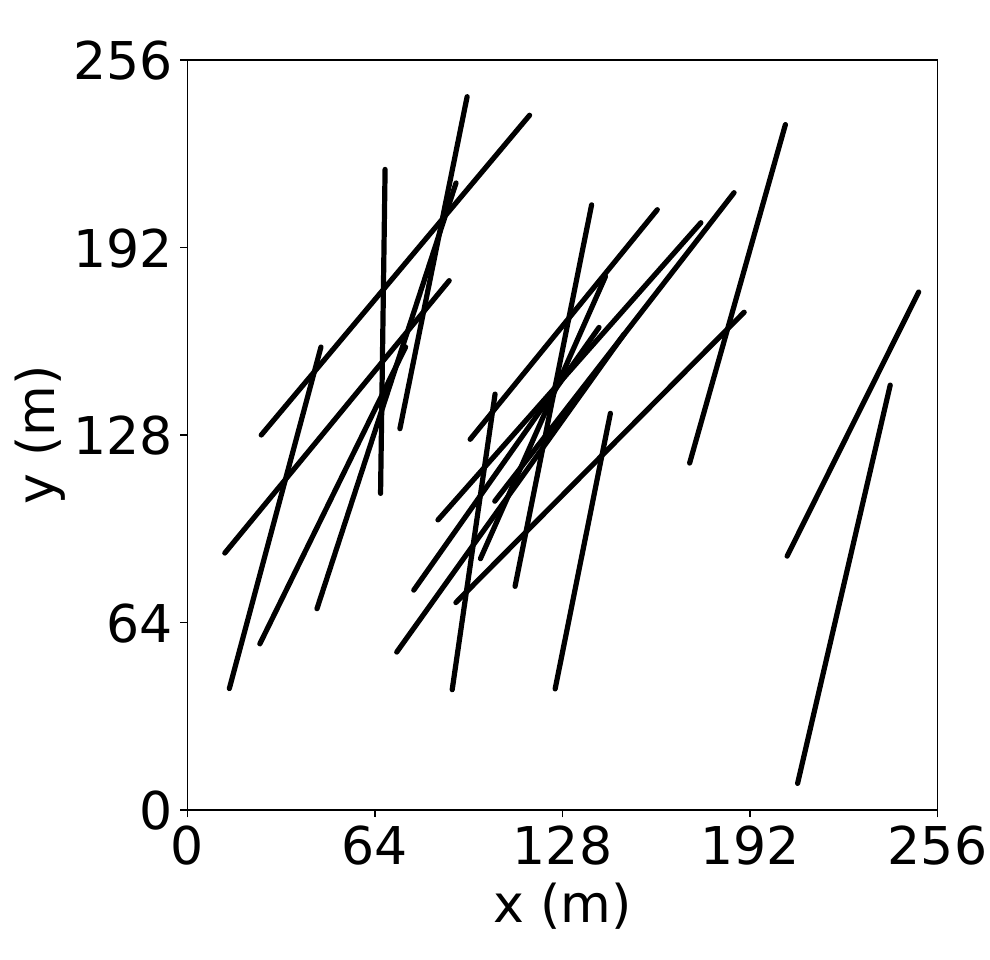}}
        \subcaption{Generated realization 2}
    \end{minipage}\hfill
    \begin{minipage}[t]{0.32\textwidth}
        \centering
        \makebox[\linewidth][c]{\includegraphics[width=1.05\linewidth]{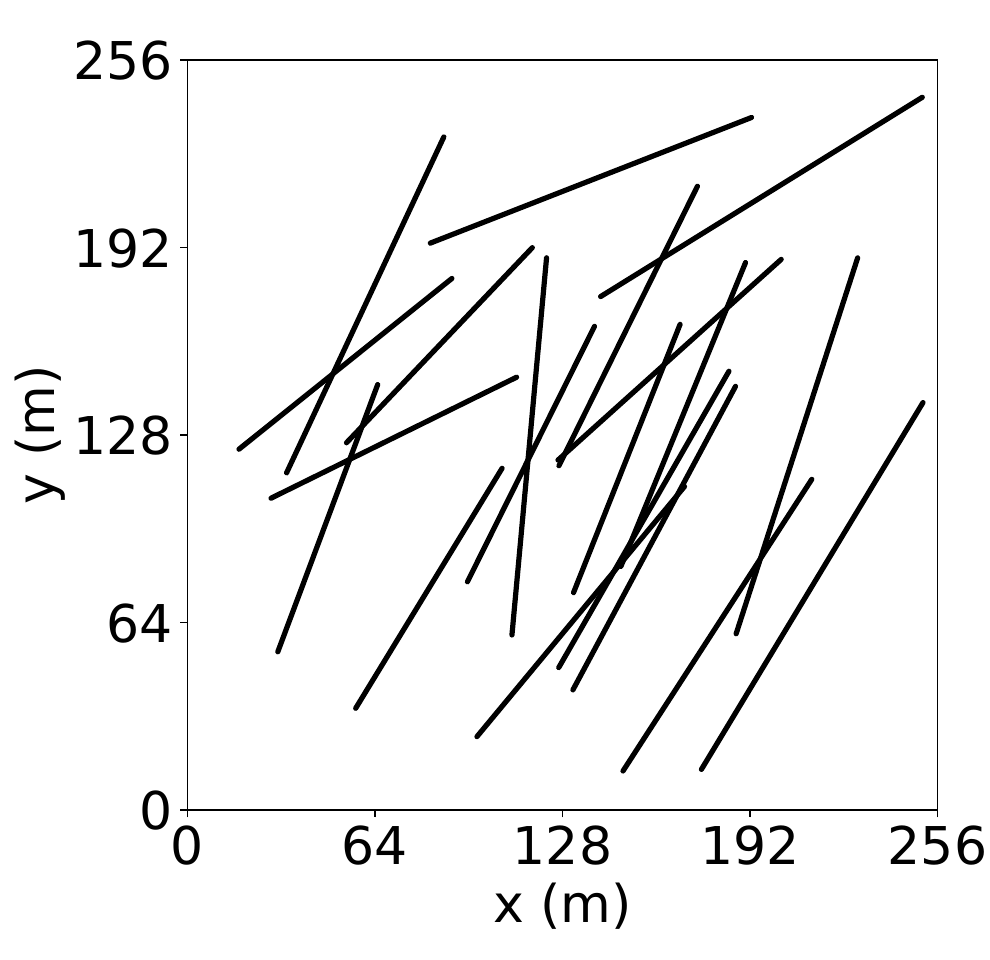}}
        \subcaption{Generated realization 3}
    \end{minipage}

    \caption{Examples of DFN realizations. (a) True DFN used for surrogate evaluation and history matching, excluded from the generative and surrogate training datasets; (b,c) training DFNs; and (d-f) DFNs generated by the flow matching model. Injection, production, and monitoring well locations are indicated in (a).}
    \label{fig:4_dfn_realizations}
\end{figure}

Flow and transport simulations are performed using TOUGH+RealGasBrine (TRGB) \cite{moridis2014core,moridis2014rgb} with the EDFM discretization described in Section~\ref{sec:3_1_edfm}. The simulations account for nonisothermal single-phase aqueous flow, heat transfer, and tracer transport. The model has a thickness of 1~m and is discretized into $64\times64$ matrix cells, each measuring $4~\mathrm{m}\times4~\mathrm{m}$. Fractures are partitioned at matrix-cell boundaries, starting from the lower vertex. Fracture elements shorter than 1~m are merged with the next element, except for the last element, which is merged with the preceding one. Merged elements spanning multiple matrix cells are connected to each intersected cell, with connection areas and distances determined from the fracture portion within each cell. Gravity is neglected. The injection and production wells are located in the lower-left and upper-right corner cells, respectively (Fig.~\ref{fig:4_dfn_realizations}a).

The matrix permeability and porosity are $5\times10^{-15}~\mathrm{m}^2$ and 0.1, respectively. The fracture permeability is $10^{-8}~\mathrm{m}^2$, with an aperture of 2~mm and a porosity of 0.99. The rock density, thermal conductivity, and specific heat are $2700~\mathrm{kg/m^3}$, $2.5~\mathrm{W/(m\cdot K)}$, and $1000~\mathrm{J/(kg\cdot K)}$, respectively. The initial reservoir pressure and temperature are 29~MPa and $200~^\circ\mathrm{C}$. Water at approximately $50~^\circ\mathrm{C}$ is injected at a constant rate of 0.1~kg/s, with the production well maintained at 29~MPa. A tracer at a concentration of 100~mg/kg is co-injected during the first 12~hours.

Each simulation covers 1800~days. Temperature fields are recorded every 30~days over the full simulation period, giving $N_{\mathrm{t},T}=60$ snapshots. Tracer concentration fields are recorded every 15~days over the first 720~days, giving $N_{\mathrm{t},C}=48$ snapshots. TRGB simulations are performed for all 5,000 training DFNs, with each simulation requiring approximately 5~minutes on a single Intel Xeon Platinum 8481C CPU core.

Six potential monitoring locations are considered (M1-M6), as shown in Fig.~\ref{fig:4_dfn_realizations}(a). M1 coincides with the production well, M2 is at the domain center, and M3-M6 are located at $(86,86)$, $(186,86)$, $(86,186)$, and $(186,186)$~m. Three monitoring configurations are considered, using one (M1), two (M1 and M2), and six (M1-M6) locations. For each configuration, three observation setups are evaluated: temperature only (T), tracer concentration only (C), and both temperature and tracer concentration (T+C).

The history matching period is 720~days, with temperature and tracer observations collected every 30 and 15~days, respectively. This gives 24 temperature and 48 tracer observations per monitoring location. Synthetic observations are generated from TRGB simulations of the true DFN in Fig.~\ref{fig:4_dfn_realizations}(a), which is excluded from both model training datasets. Independent Gaussian measurement errors are added, with standard deviations of $2~^\circ\mathrm{C}$ for temperature and 0.01~mg/kg for tracer concentration. Surrogate error variances are estimated separately for each monitoring location and response type using the validation realizations over the history matching period. The total observation error variance is $\sigma_i^2=\sigma_{\mathrm{meas},i}^2+\sigma_{\mathrm{sur},i}^2$, as described in Section~\ref{sec:4_1_bayesian}.

\subsection{Evaluation of the Vertex-based Flow Matching Model}

The vertex-based flow matching model is trained using the 5,000 DFN realizations described in Section~\ref{sec:5_1_setup}. Model performance is evaluated by comparing 1,000 generated DFNs with 1,000 randomly selected training DFNs in terms of fracture geometry and simulated flow and transport responses.

\begin{figure}[!ht]
    \centering
    \begin{minipage}[t]{0.48\textwidth}
        \centering
        \includegraphics[width=\linewidth]{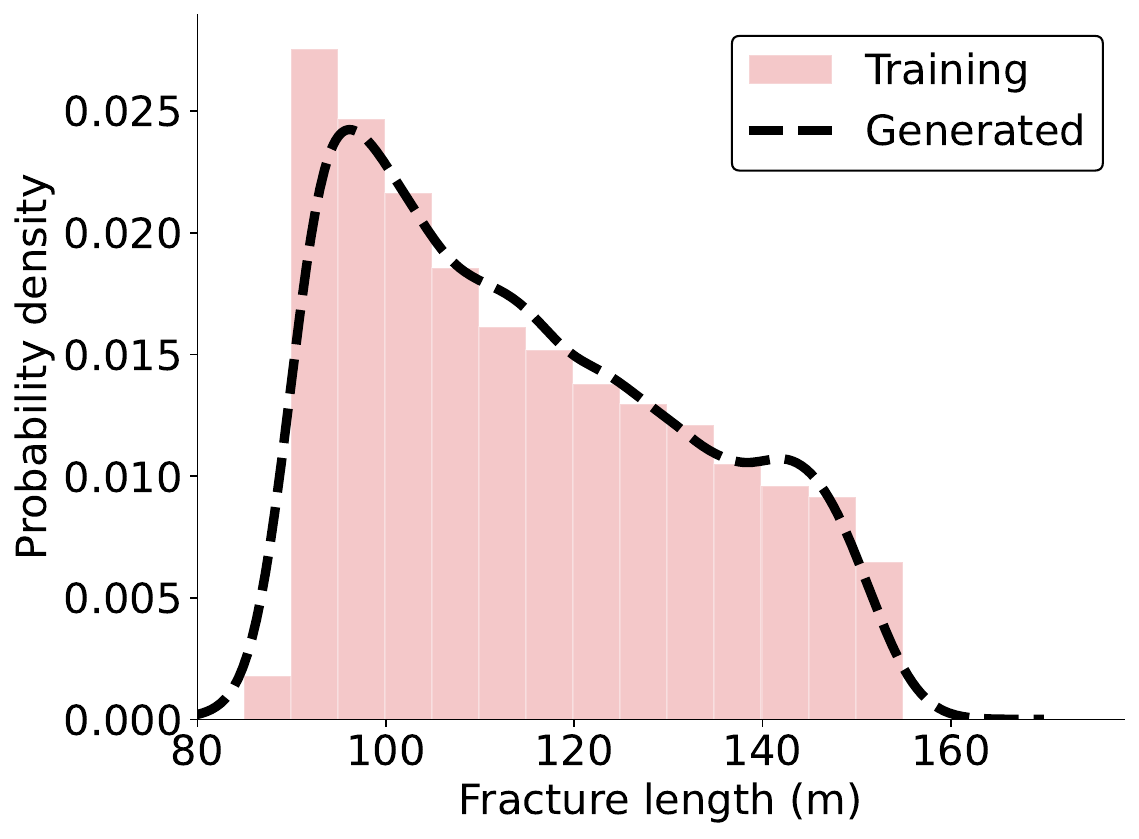}
        \subcaption{Individual fracture length}
    \end{minipage}\hfill
    \begin{minipage}[t]{0.48\textwidth}
        \centering
        \includegraphics[width=\linewidth]{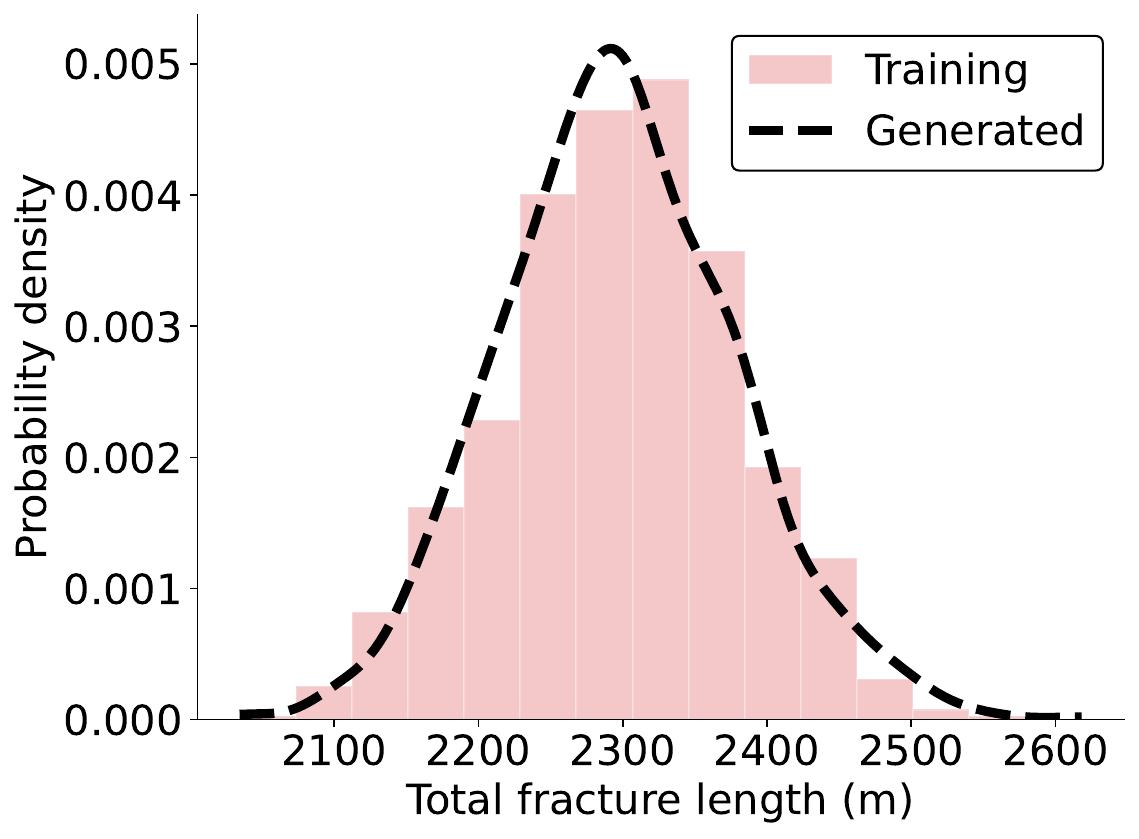}
        \subcaption{Total fracture length per network}
    \end{minipage}

    \begin{minipage}[t]{0.48\textwidth}
        \centering
        \includegraphics[width=\linewidth]{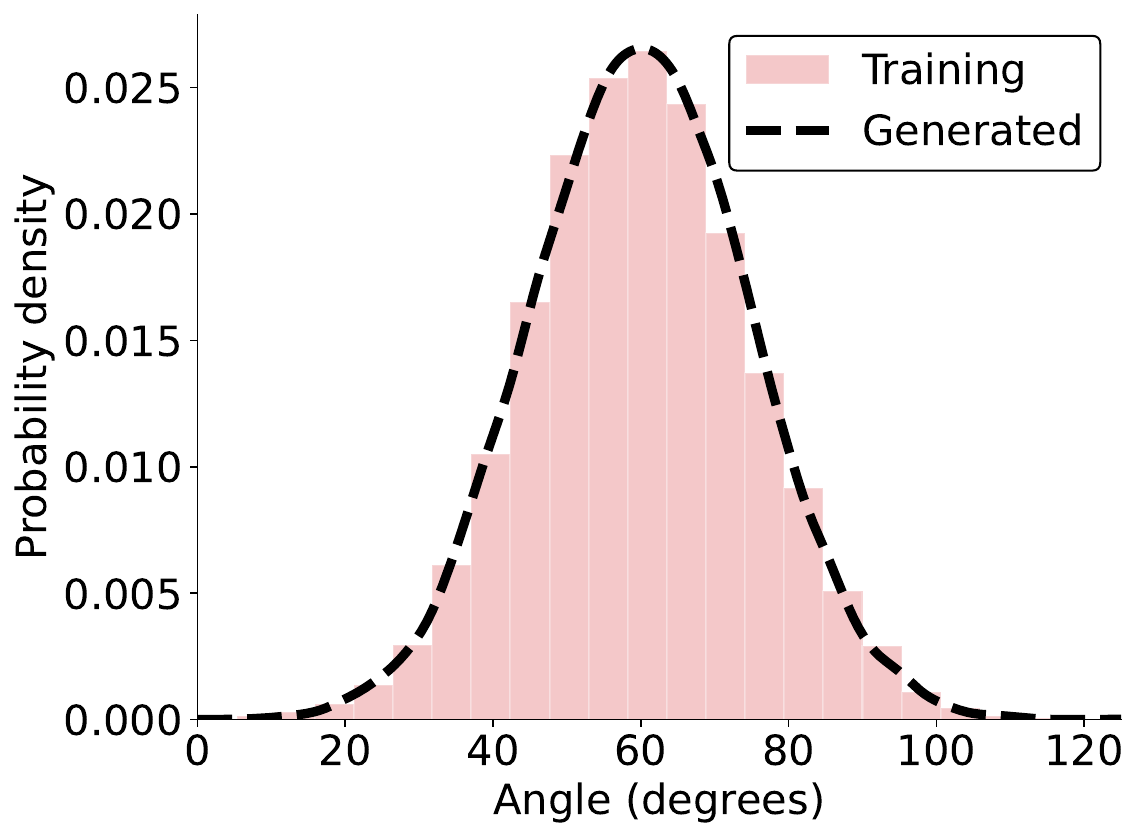}
        \subcaption{Fracture orientation}
    \end{minipage}\hfill
    \begin{minipage}[t]{0.48\textwidth}
        \centering
        \includegraphics[width=0.8\linewidth]{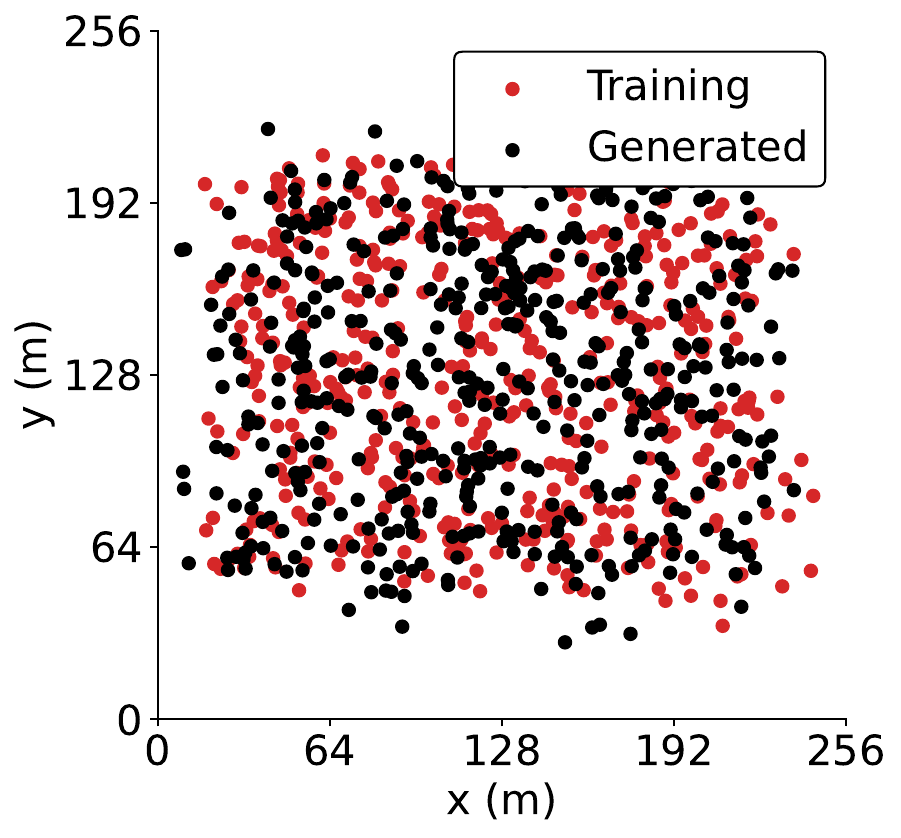}
        \subcaption{Fracture midpoint location}
    \end{minipage}

    \caption{Comparison of geometric statistics for 1,000 training and 1,000 generated DFN realizations: (a) individual fracture length, (b) total fracture length per network, (c) fracture orientation, and (d) 500 randomly selected fracture midpoints from each set.}
    \label{fig:5_dfn_statistics}
\end{figure}

Examples of training and generated DFNs are shown in Fig.~\ref{fig:4_dfn_realizations}. The generated realizations exhibit similar fracture orientations, lengths, spatial distributions, and connectivity patterns to those of the training DFNs. Figure~\ref{fig:5_dfn_statistics} compares the distributions of individual fracture length, total fracture length per network, fracture orientation, and fracture midpoint location. Reasonable agreement is observed for all four quantities. The generated fracture lengths follow the prescribed truncated power-law distribution, and the distributions of total fracture length and orientation are reproduced. The spatial distributions of fracture midpoints also show similar characteristics.

\begin{figure}[!ht]
    \centering
    \begin{minipage}[t]{0.48\textwidth}
        \centering
        \includegraphics[width=\linewidth]{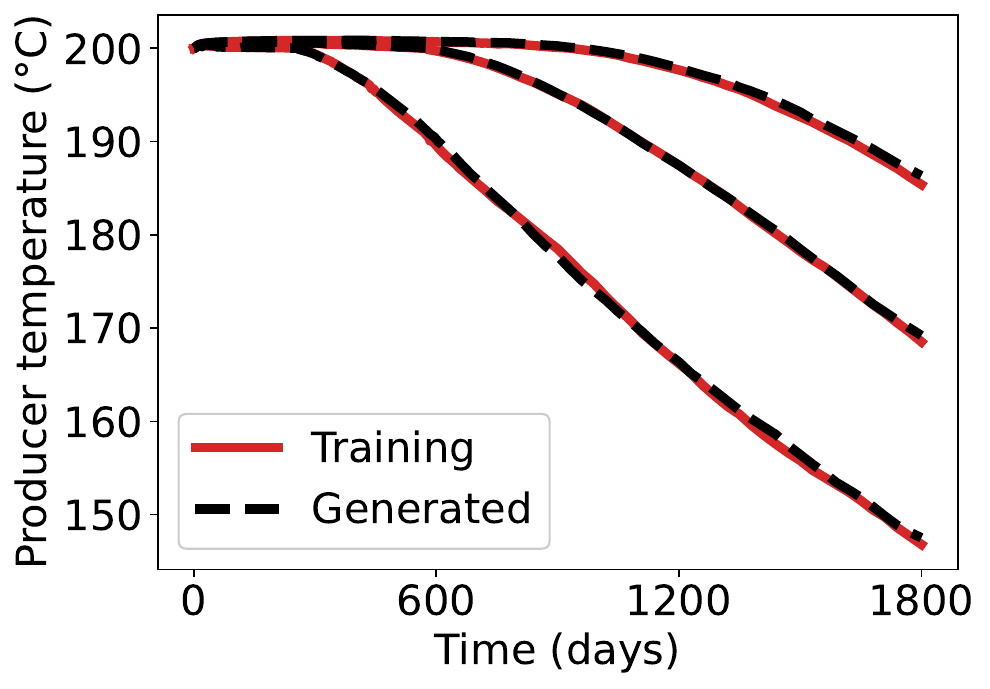}
        \subcaption{Temperature}
    \end{minipage}\hfill
    \begin{minipage}[t]{0.48\textwidth}
        \centering
        \includegraphics[width=\linewidth]{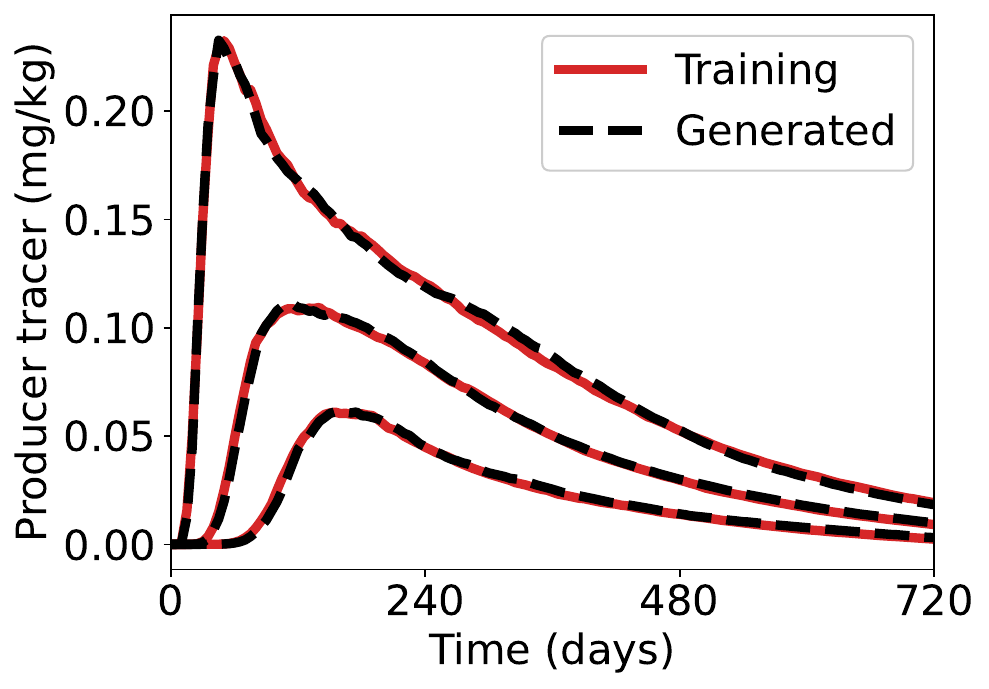}
        \subcaption{Tracer concentration}
    \end{minipage}

    \caption{Comparison of (a) producer temperature and (b) tracer concentration from TRGB simulations of 1,000 training and 1,000 generated DFN realizations. Red solid and black dashed curves denote the training and generated results, respectively. The lower, middle, and upper curves represent the $P_{\mathrm{10}}$, $P_{\mathrm{50}}$, and $P_{\mathrm{90}}$ responses at each time.}
    \label{fig:6_generative_flow_response}
\end{figure}

The ability of the generative model to reproduce reservoir responses is evaluated using TRGB simulations of the 1,000 generated DFNs. Figure~\ref{fig:6_generative_flow_response} compares the $P_{\mathrm{10}}$, $P_{\mathrm{50}}$, and $P_{\mathrm{90}}$ producer temperature and tracer concentration responses from the training and generated DFNs. The percentiles are computed across the 1,000 realizations at each time. The two sets show close agreement throughout the simulation periods for both responses. These results indicate that the generated DFNs reproduce not only the geometric statistics of the training networks but also the distributions of their flow and transport responses.

\subsection{Evaluation of the Geometry-Aware Surrogate Model}
\label{sec:5_3_surrogate}

The geometry-aware surrogate is trained using the DFN realizations and corresponding TRGB simulation results described in Section~\ref{sec:5_1_setup}. Of the 5,000 realizations, 4,500 are used for training and 500 for validation. Separate networks are trained for temperature and tracer concentration. The mean absolute errors (MAEs), averaged over all matrix cells, prediction times, and validation realizations, are $0.94~^\circ\mathrm{C}$ for temperature and 0.0051~mg/kg for tracer concentration. For history matching, surrogate error standard deviations are estimated at each monitoring location from the validation results over the 720-day assimilation period. These range from 0.2 to $2.9~^\circ\mathrm{C}$ for temperature and from 0.007 to 0.018~mg/kg for tracer concentration.

\begin{figure}[!htbp]
  \centering
  
  \begin{minipage}[t]{0.31\textwidth}
    \centering
    \includegraphics[height=3.7cm]{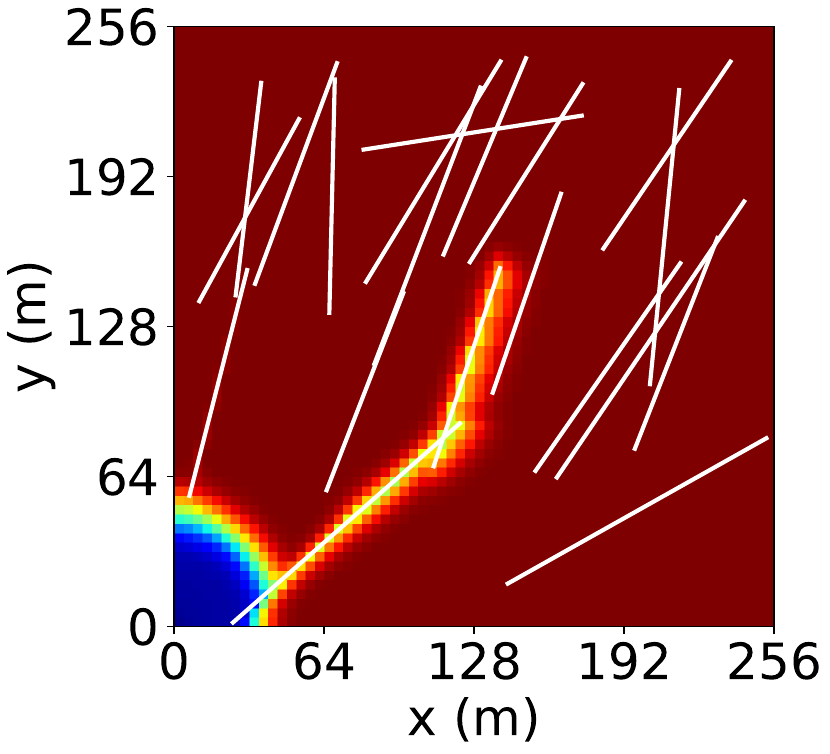}
    \subcaption{Simulation at 180 days}
  \end{minipage}\hfill
  \begin{minipage}[t]{0.31\textwidth}
    \centering
    \includegraphics[height=3.7cm]{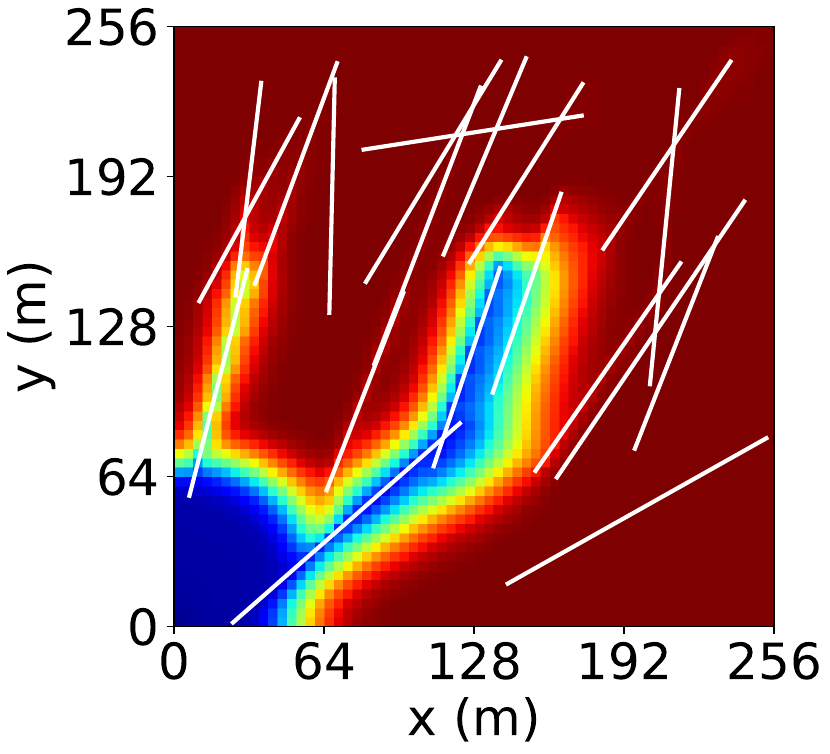}
    \subcaption{Simulation at 720 days}
\end{minipage}\hfill
  \begin{minipage}[t]{0.37\textwidth}
    \centering
    \includegraphics[height=3.7cm]{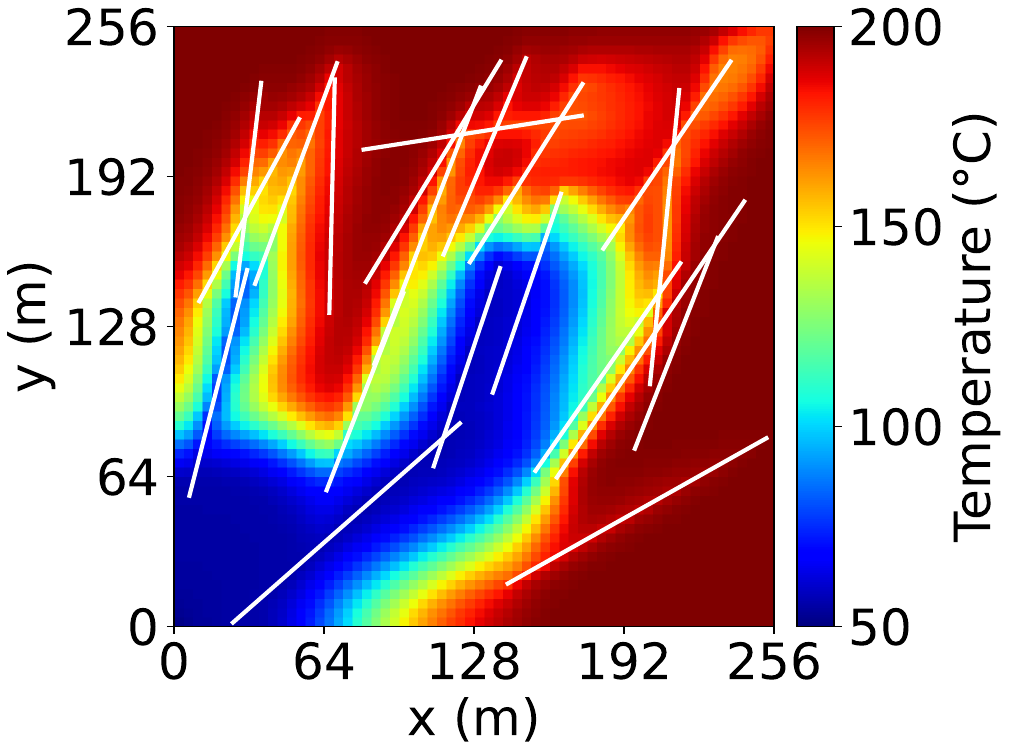}
    \subcaption{Simulation at 1,800 days}
  \end{minipage}

  \begin{minipage}[t]{0.31\textwidth}
    \centering
    \includegraphics[height=3.7cm]{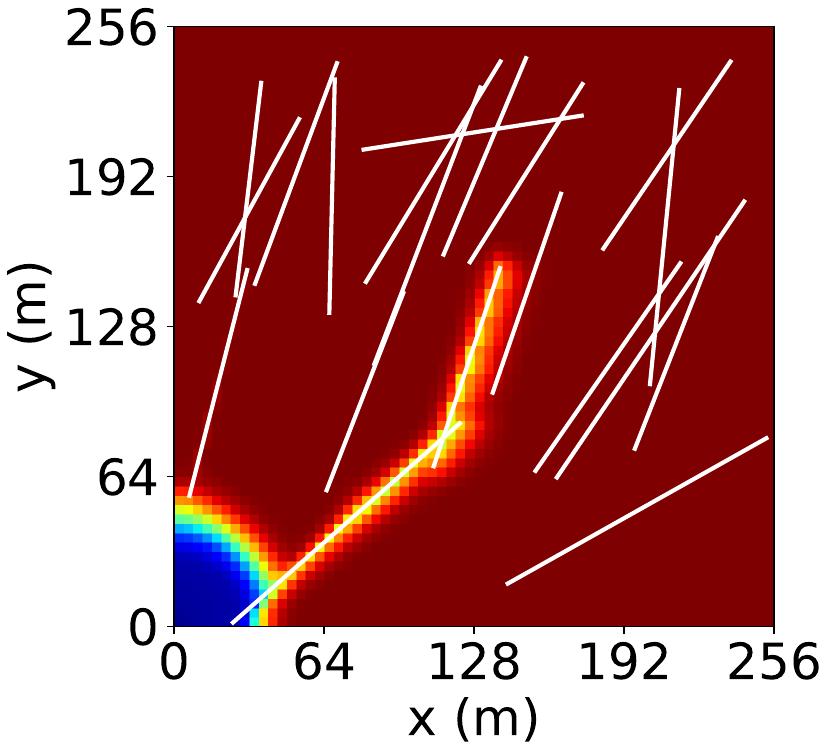}
    \subcaption{Prediction at 180 days}
  \end{minipage}\hfill
  \begin{minipage}[t]{0.31\textwidth}
    \centering
    \includegraphics[height=3.7cm]{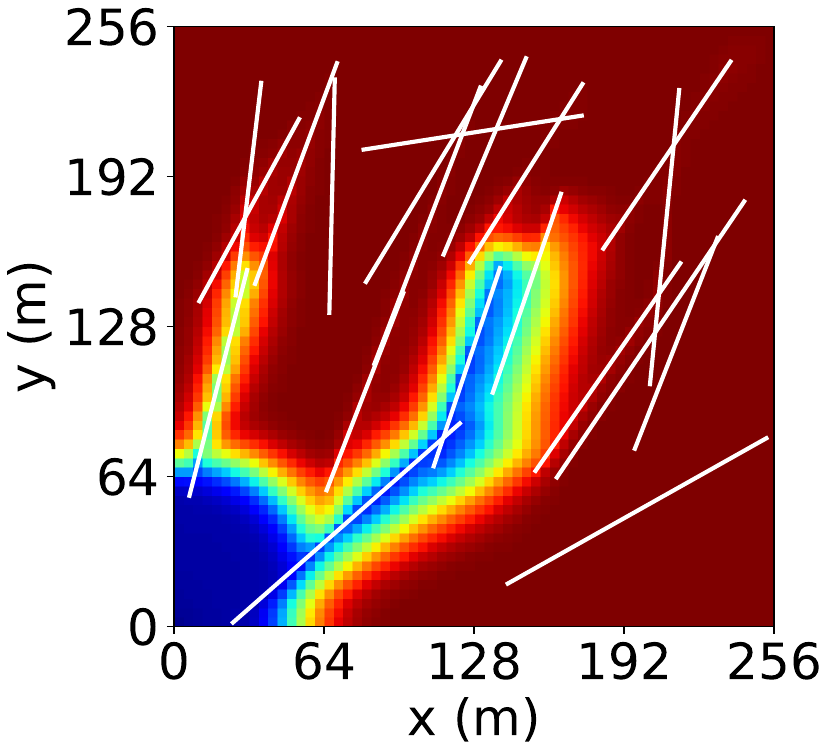}
    \subcaption{Prediction at 720 days}
  \end{minipage}\hfill
  \begin{minipage}[t]{0.37\textwidth}
    \centering
    \includegraphics[height=3.7cm]{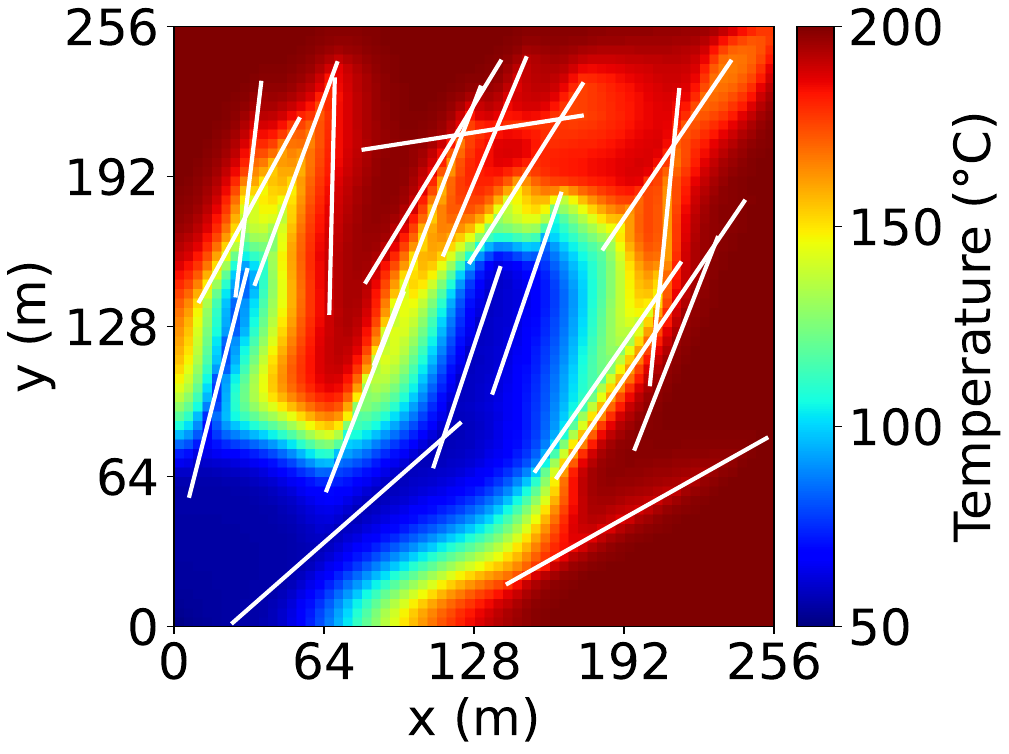}
    \subcaption{Prediction at 1,800 days}
  \end{minipage}

  \caption{Comparison of temperature fields from TRGB simulations (top row) and surrogate predictions (bottom row) at 180, 720, and 1800~days for the DFN realization shown in Fig.~\ref{fig:4_dfn_realizations}(a). White lines indicate fractures.}
  \label{fig:7_temperature_fields}
\end{figure}

\begin{figure}[!htbp]
  \centering

  \begin{minipage}[t]{0.31\textwidth}
    \centering
    \includegraphics[height=3.7cm]{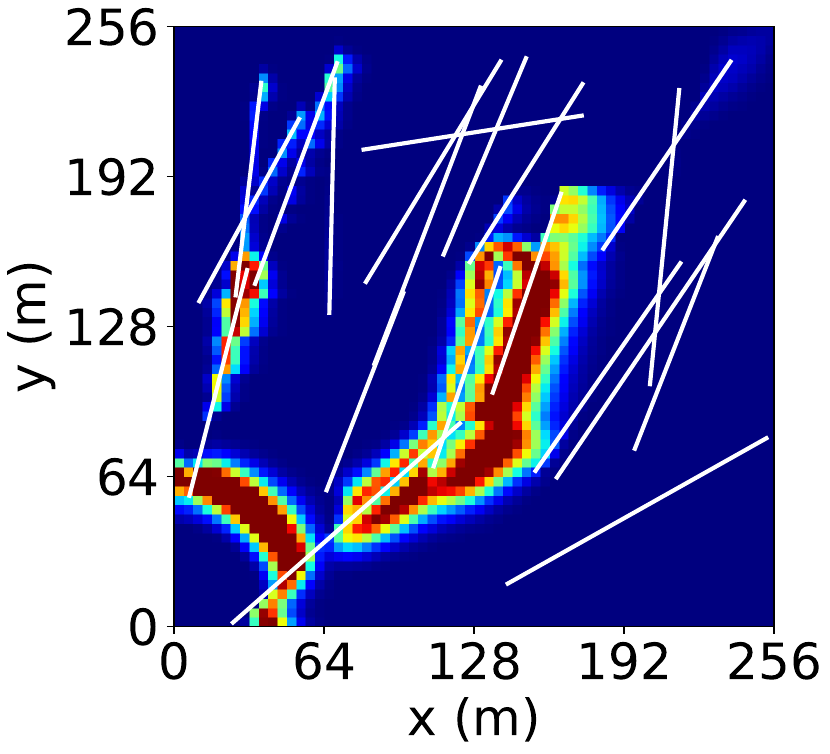}
    \subcaption{Simulation at 60 days}
  \end{minipage}\hfill
  \begin{minipage}[t]{0.31\textwidth}
    \centering
    \includegraphics[height=3.7cm]{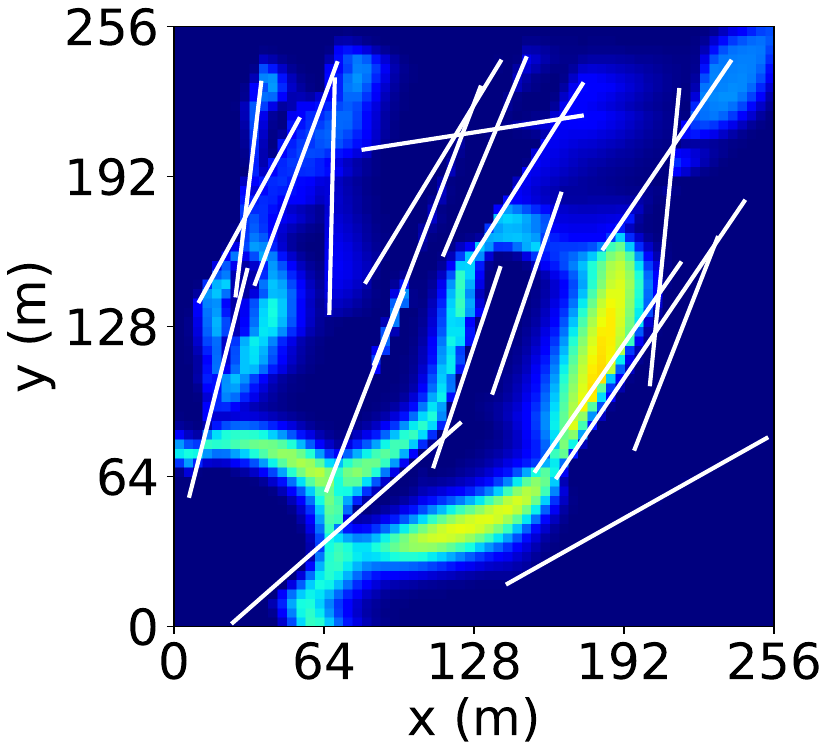}
    \subcaption{Simulation at 180 days}
  \end{minipage}\hfill
  \begin{minipage}[t]{0.37\textwidth}
    \centering
    \includegraphics[height=3.7cm]{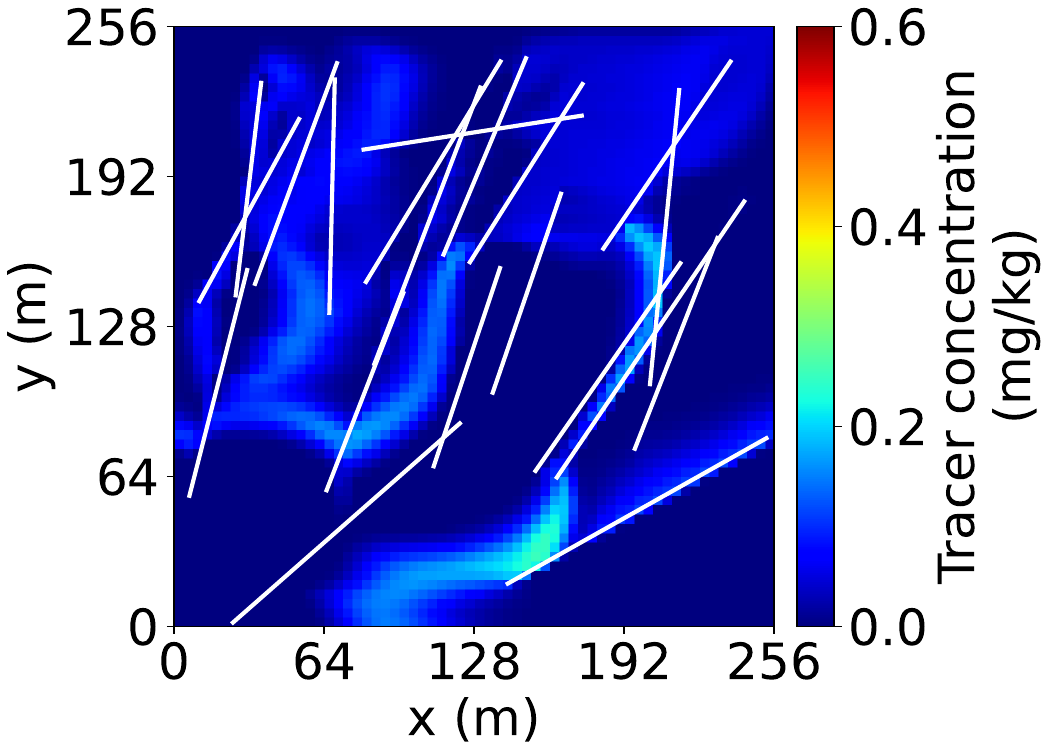}
    \subcaption{Simulation at 360 days}
  \end{minipage}

  \begin{minipage}[t]{0.31\textwidth}
    \centering
    \includegraphics[height=3.7cm]{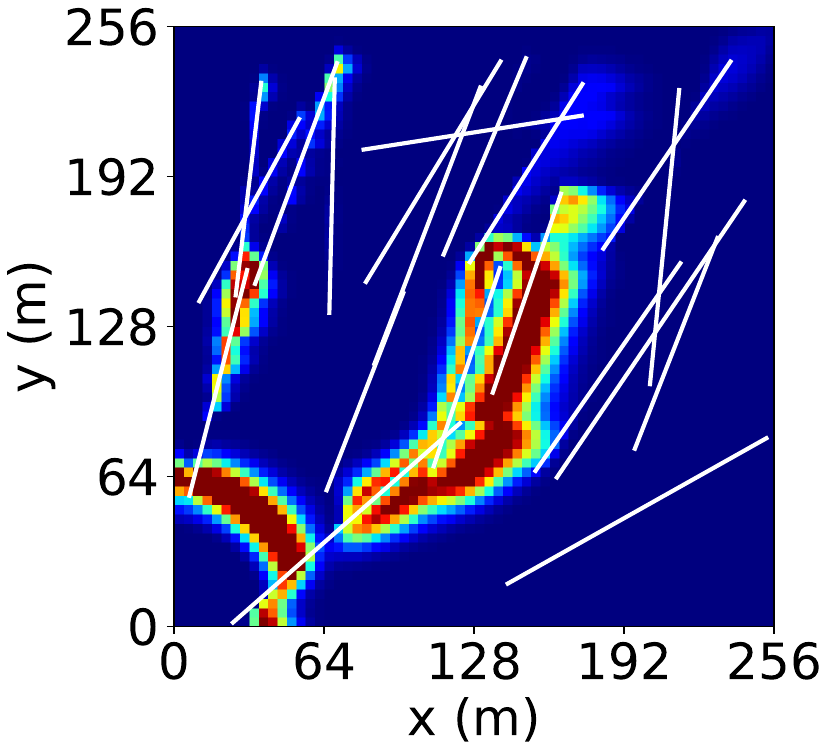}
    \subcaption{Prediction at 60 days}
  \end{minipage}\hfill
  \begin{minipage}[t]{0.31\textwidth}
    \centering
    \includegraphics[height=3.7cm]{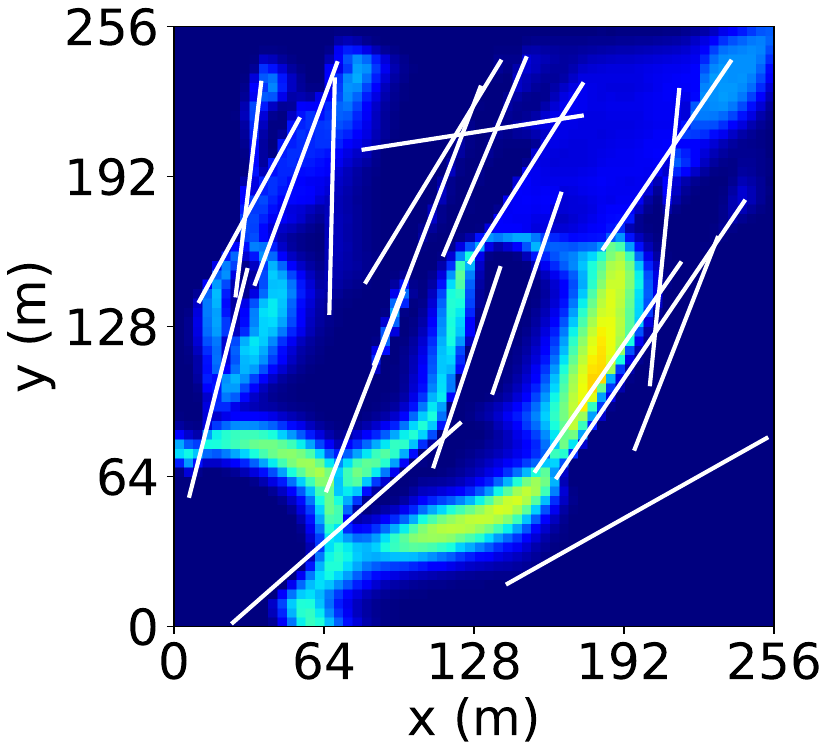}
    \subcaption{Prediction at 180 days}
  \end{minipage}\hfill
  \begin{minipage}[t]{0.37\textwidth}
    \centering
    \includegraphics[height=3.7cm]{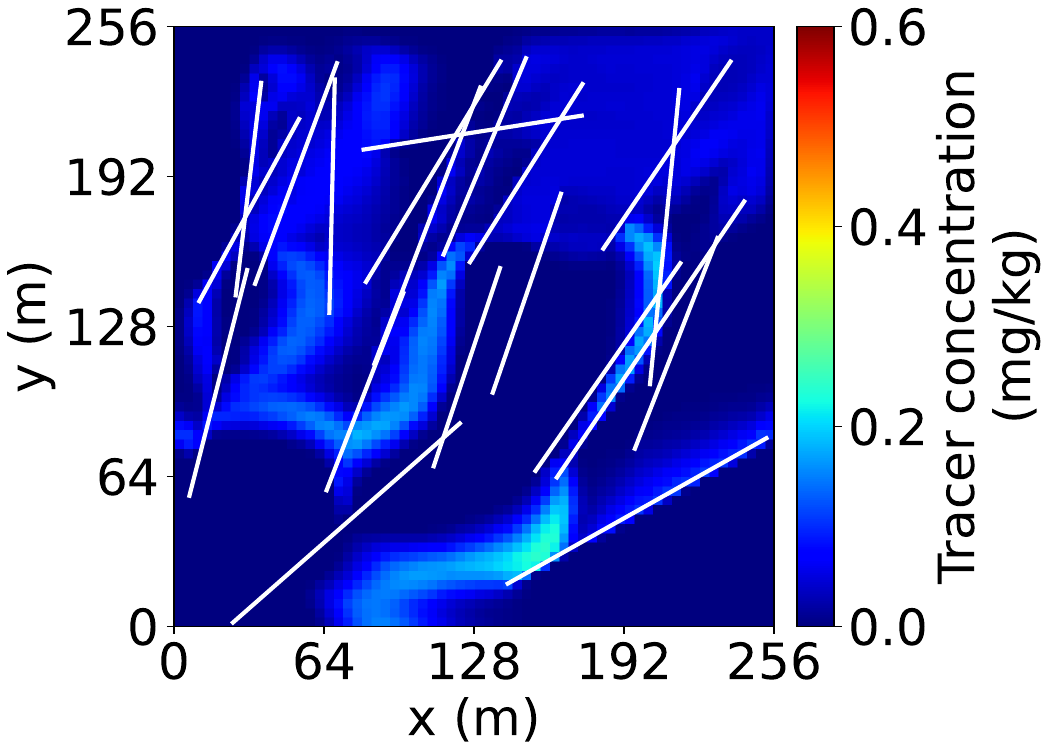}
    \subcaption{Prediction at 360 days}
  \end{minipage}

  \caption{Comparison of tracer concentration fields from TRGB simulations (top row) and surrogate predictions (bottom row) at 60, 180, and 360~days for the DFN realization shown in Fig.~\ref{fig:4_dfn_realizations}(a). White lines indicate fractures.}
  \label{fig:8_tracer_fields}
\end{figure}

Figures~\ref{fig:7_temperature_fields} and \ref{fig:8_tracer_fields} show the temperature and tracer concentration fields from TRGB simulations and surrogate predictions for the DFN realization in Fig.~\ref{fig:4_dfn_realizations}(a). The temperature and tracer fields exhibit preferential transport along the fractures, with the cooled region and tracer front extending from the injection well toward the production well. The surrogate predictions show reasonable agreement with the TRGB results at the times considered, including the spatial extent of the cooled region and the tracer front along individual fractures. Some discrepancies are observed at later times, particularly a slight underprediction of temperature in the upper-right portion of the domain at 1800~days.

\begin{figure}[!htbp]
  \centering
  \begin{minipage}[t]{0.48\textwidth}
    \centering
    \includegraphics[width=\linewidth]{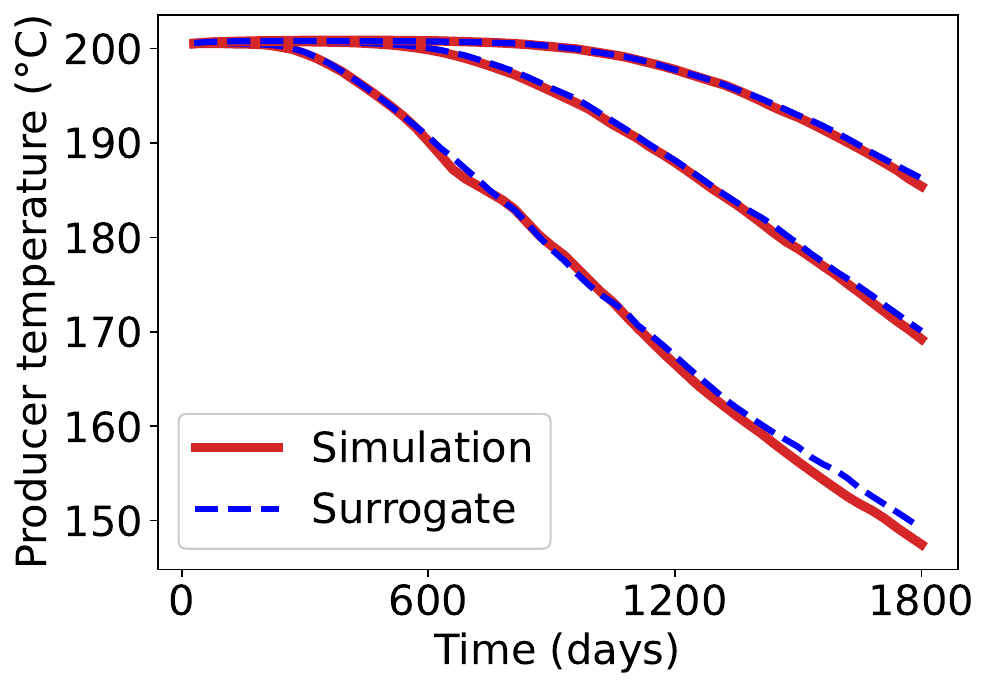}
    \subcaption{Temperature}
  \end{minipage}\hfill
  \begin{minipage}[t]{0.48\textwidth}
    \centering
    \includegraphics[width=\linewidth]{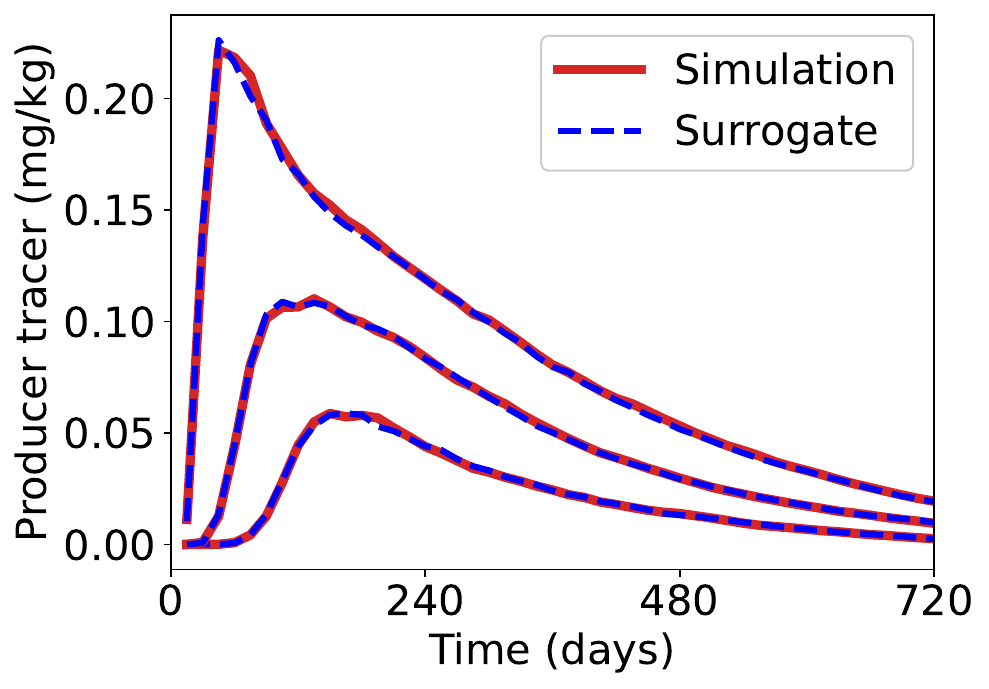}
    \subcaption{Tracer concentration}
  \end{minipage}
  \caption{Comparison of (a) producer temperature and (b) tracer concentration from TRGB simulations and surrogate predictions over 500 validation realizations. Red solid and blue dashed curves denote the TRGB and surrogate results. The lower, middle, and upper curves represent the $P_{\text{10}}$, $P_{\text{50}}$, and $P_{\text{90}}$ responses at each time.}
  \label{fig:9_surrogate_flow_response}
\end{figure}

The surrogate is further evaluated using the distributions of producer temperature and tracer concentration over the 500 validation realizations. Figure~\ref{fig:9_surrogate_flow_response} compares the $P_{\mathrm{10}}$, $P_{\mathrm{50}}$, and $P_{\mathrm{90}}$ responses from TRGB simulations and surrogate predictions. Reasonable agreement is observed for both responses, with a small discrepancy in the $P_{\mathrm{10}}$ temperature at later times. The results demonstrate that the surrogate captures both the spatial response fields and the variability in producer responses across different fracture networks.

\subsection{History Matching Results}
\label{sec:5_4_history_matching}

The effects of observation type and monitoring configuration on fracture network inversion are investigated using PT-MCMC. Results are first presented for the true DFN in Fig.~\ref{fig:4_dfn_realizations}(a), followed by additional true models. A comparison with ESMDA is provided in Text~S4 of the Supporting Information. In the following figures, gray regions represent the prior $P_{\mathrm{10}}$-$P_{\mathrm{90}}$ ranges, black curves denote the posterior $P_{\mathrm{10}}$, $P_{\mathrm{50}}$, and $P_{\mathrm{90}}$ responses, and red curves and symbols indicate the true responses and observations. Vertical dashed lines mark the end of the 720-day history matching period.

\begin{figure}[!htbp]
    \centering
    \captionsetup[subfigure]{singlelinecheck=off,justification=centering}
    \begin{minipage}[t]{0.33\textwidth}
        \centering
        \includegraphics[width=\linewidth]{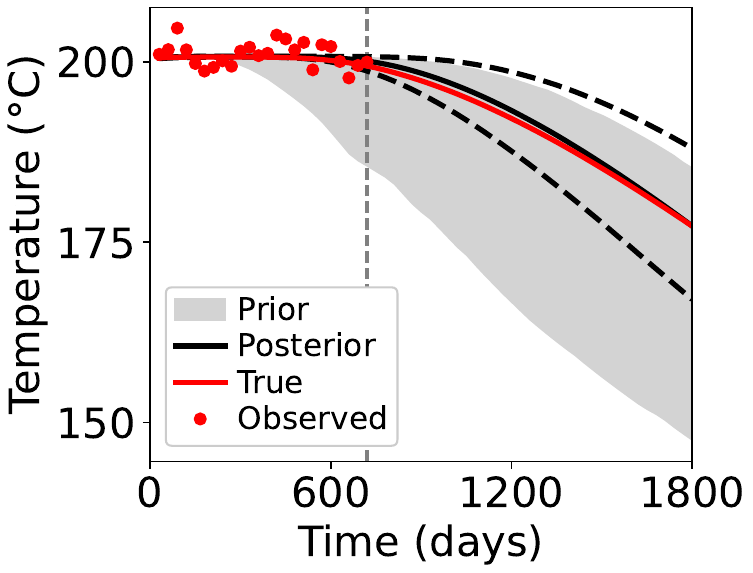}
        \subcaption{Temperature, T}
    \end{minipage}\hfill
    \begin{minipage}[t]{0.33\textwidth}
        \centering
        \includegraphics[width=\linewidth]{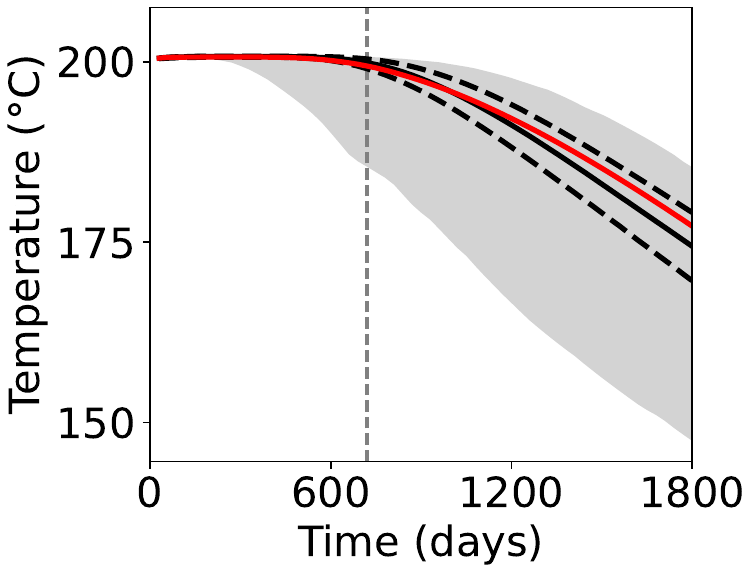}
        \subcaption{Temperature, C}
    \end{minipage}\hfill
    \begin{minipage}[t]{0.33\textwidth}
        \centering
        \includegraphics[width=\linewidth]{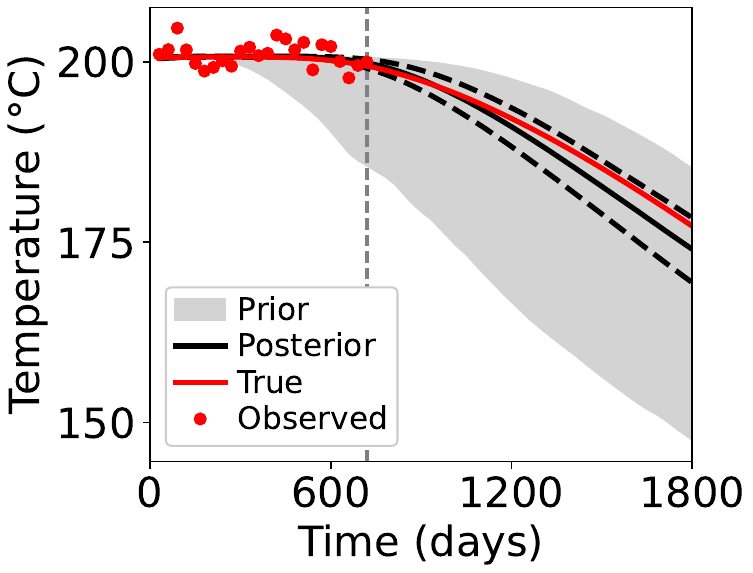}
        \subcaption{Temperature, T+C}
    \end{minipage}

    \medskip

    \begin{minipage}[t]{0.33\textwidth}
        \centering
        \includegraphics[width=\linewidth]{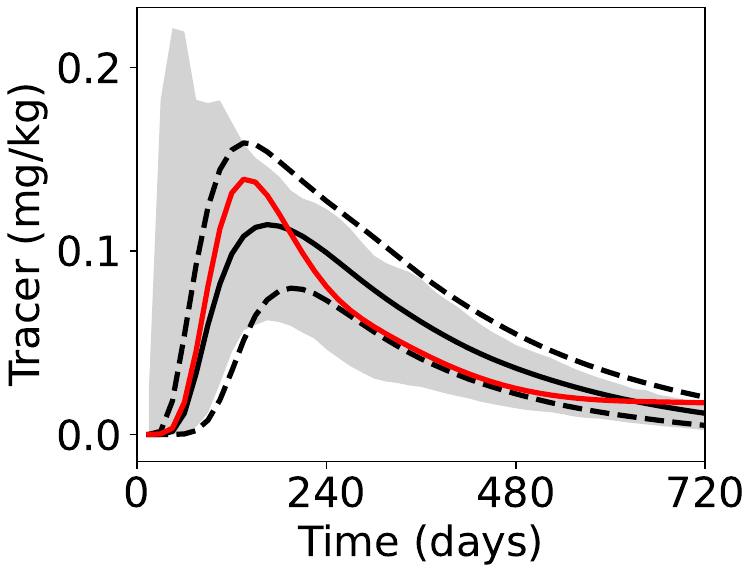}
        \subcaption{Tracer, T}
    \end{minipage}\hfill
    \begin{minipage}[t]{0.33\textwidth}
        \centering
        \includegraphics[width=\linewidth]{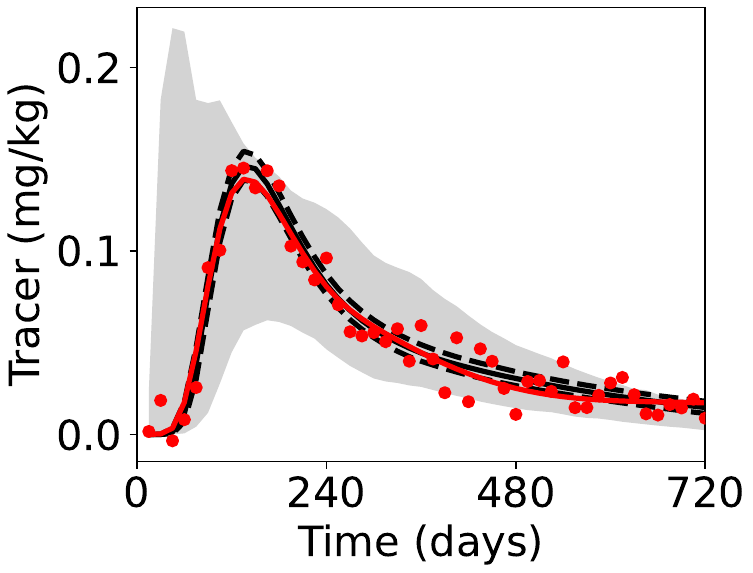}
        \subcaption{Tracer, C}
    \end{minipage}\hfill
    \begin{minipage}[t]{0.33\textwidth}
        \centering
        \includegraphics[width=\linewidth]{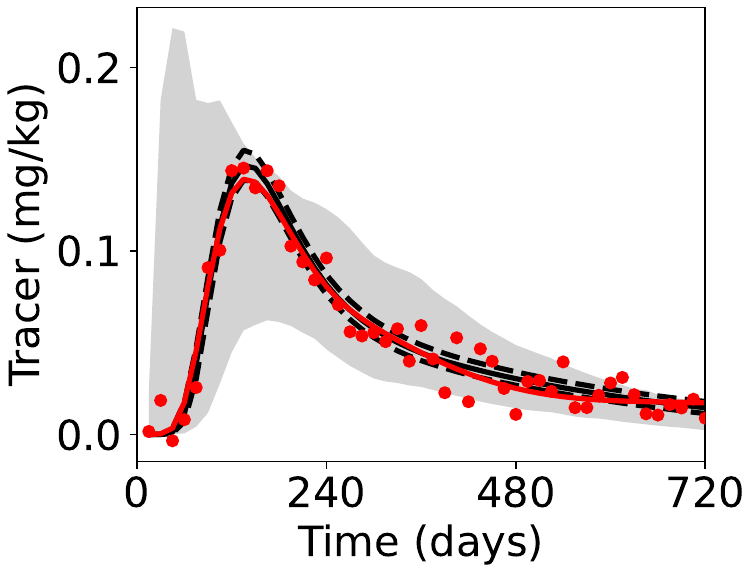}
        \subcaption{Tracer, T+C}
    \end{minipage}

    \caption{History matching results at the production well using temperature only (T), tracer concentration only (C), and both data types (T+C). Temperature and tracer concentration responses are shown in the top and bottom rows, respectively. Gray regions indicate the prior $P_{\mathrm{10}}$-$P_{\mathrm{90}}$ ranges, black curves show the posterior $P_{\mathrm{10}}$, $P_{\mathrm{50}}$, and $P_{\mathrm{90}}$ responses, and red curves and symbols denote the true responses and observations. Vertical dashed lines indicate the end of the history matching period. The legend in (a) applies to all panels.}
    \label{fig:10_history_matching_setup_compare}
\end{figure}

Figure~\ref{fig:10_history_matching_setup_compare} compares history matching results using temperature only (T), tracer concentration only (C), and both temperature and tracer concentration (T+C), with observations at the production well (M1). Temperature-only assimilation provides limited uncertainty reduction, particularly for tracer concentration predictions. During the history matching period, the producer temperature remains close to its initial value of $200~^\circ\mathrm{C}$ and is therefore relatively insensitive to differences in fracture geometry. In contrast, tracer breakthrough occurs within this period and provides information about the flow pathways connecting the injection and production wells. Assimilation of tracer data substantially reduces uncertainty in both tracer and temperature predictions. Combining temperature and tracer observations provides additional constraints and produces posterior responses in close agreement with the true responses. These results indicate that tracer observations are particularly informative for predicting the subsequent thermal response in this case.

\begin{figure}[!htbp]
    \centering
    \captionsetup[subfigure]{singlelinecheck=off,justification=centering}
    \begin{minipage}[t]{0.33\textwidth}
        \centering
        \includegraphics[width=\linewidth]{case4_1well_T+C_temperature_monitor1.pdf}
        \subcaption{M1, 1 well}
    \end{minipage}\hfill
    \begin{minipage}[t]{0.33\textwidth}
        \centering
        \includegraphics[width=\linewidth]{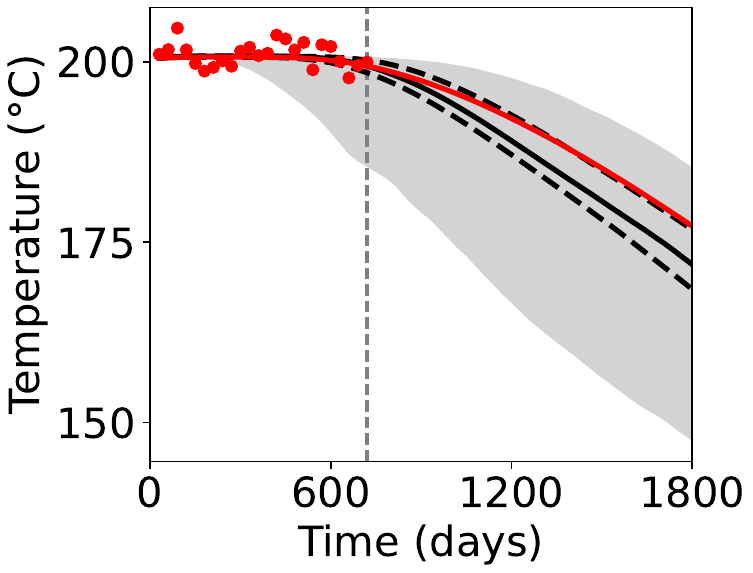}
        \subcaption{M1, 2 wells}
    \end{minipage}\hfill
    \begin{minipage}[t]{0.33\textwidth}
        \centering
        \includegraphics[width=\linewidth]{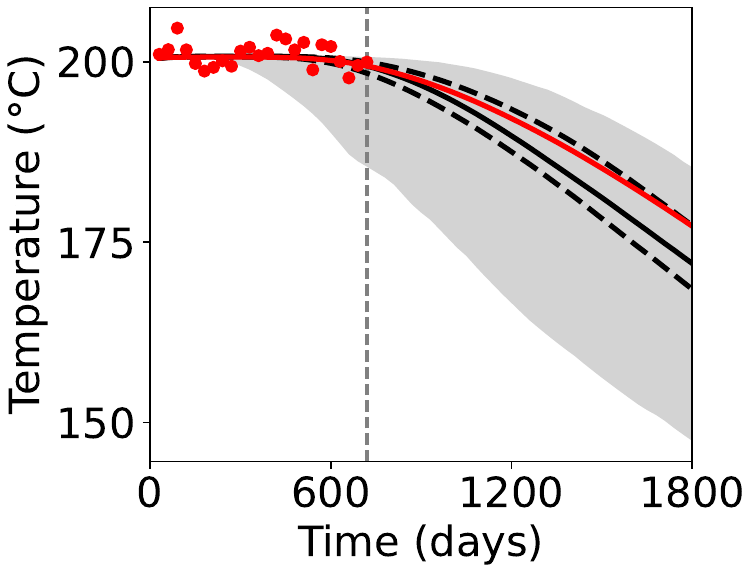}
        \subcaption{M1, 6 wells}
    \end{minipage}

    \medskip

    \begin{minipage}[t]{0.33\textwidth}
        \centering
        \includegraphics[width=\linewidth]{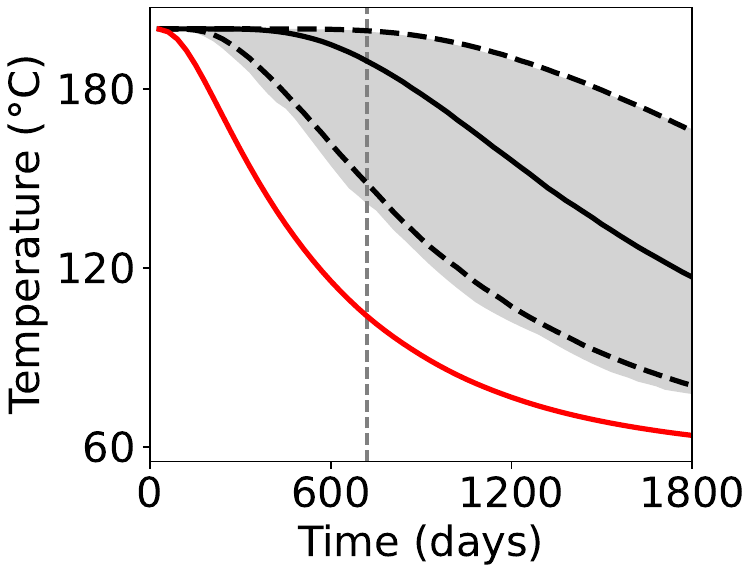}
        \subcaption{M2, 1 well}
    \end{minipage}\hfill
    \begin{minipage}[t]{0.33\textwidth}
        \centering
        \includegraphics[width=\linewidth]{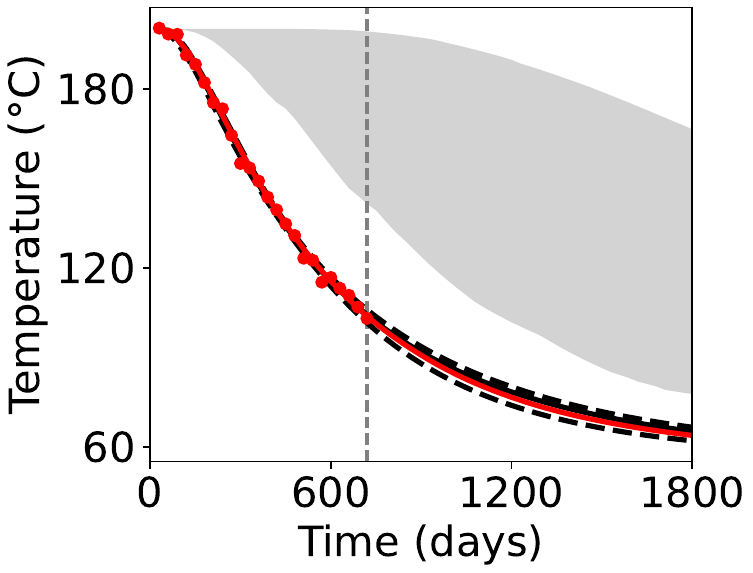}
        \subcaption{M2, 2 wells}
    \end{minipage}\hfill
    \begin{minipage}[t]{0.33\textwidth}
        \centering
        \includegraphics[width=\linewidth]{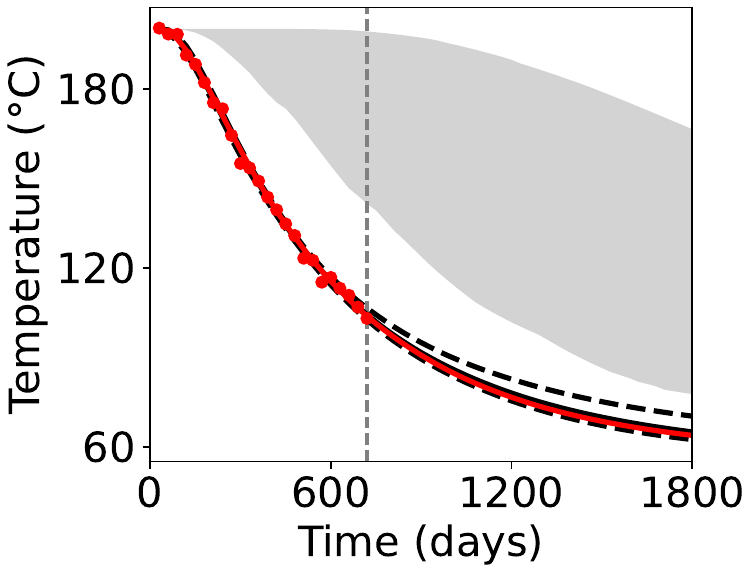}
        \subcaption{M2, 6 wells}
    \end{minipage}

    \medskip

    \begin{minipage}[t]{0.33\textwidth}
        \centering
        \includegraphics[width=\linewidth]{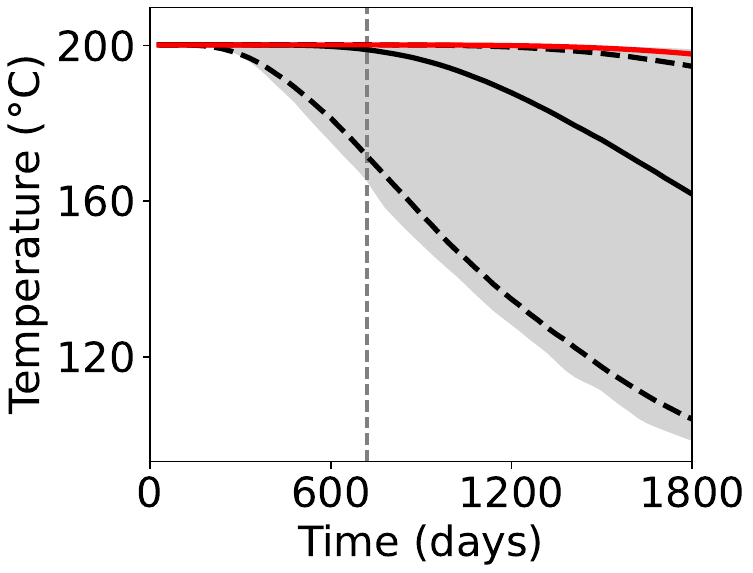}
        \subcaption{M5, 1 well}
    \end{minipage}\hfill
    \begin{minipage}[t]{0.33\textwidth}
        \centering
        \includegraphics[width=\linewidth]{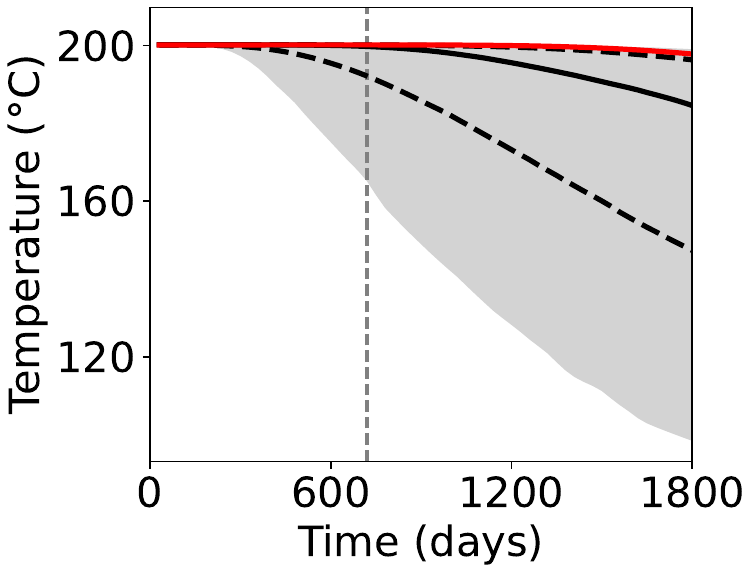}
        \subcaption{M5, 2 wells}
    \end{minipage}\hfill
    \begin{minipage}[t]{0.33\textwidth}
        \centering
        \includegraphics[width=\linewidth]{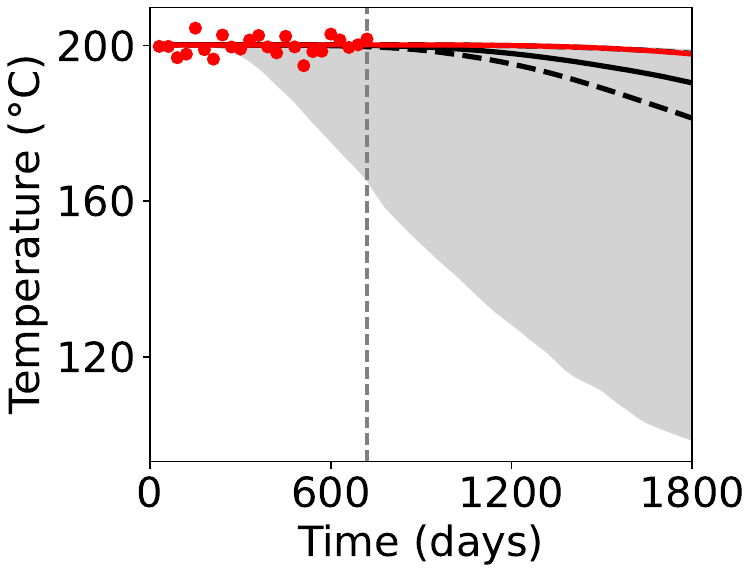}
        \subcaption{M5, 6 wells}
    \end{minipage}

    \caption{Posterior temperature predictions at M1, M2, and M5 (rows) using one, two, and six monitoring locations (columns) for the T+C setup. Gray regions indicate the prior $P_{\mathrm{10}}$-$P_{\mathrm{90}}$ ranges, black curves show the posterior $P_{\mathrm{10}}$, $P_{\mathrm{50}}$, and $P_{\mathrm{90}}$ responses, and red curves and symbols denote the true responses and observations. Vertical dashed lines indicate the end of the history matching period. The legend in (a) applies to all panels.}
    \label{fig:11_configuration_compare}
\end{figure}

The effect of monitoring configuration is evaluated using the T+C setup. Figure~\ref{fig:11_configuration_compare} compares posterior temperature predictions at M1, M2, and M5 using one, two, and six monitoring locations. At M1, all three configurations provide substantial uncertainty reduction and similar posterior predictions. However, observations at M1 alone provide limited constraints on temperature predictions at M2 and M5, with posterior predictions deviating from the true responses. Adding observations at M2 substantially reduces uncertainty and improves predictions at this location, but provides limited improvement at M5. With six monitoring locations, uncertainty at M5 is also reduced, and the posterior predictions show better agreement with the true response. These results indicate that monitoring observations provide predominantly local information in fractured systems. Reducing prediction uncertainty at a specific location therefore generally requires monitoring at or near that location, highlighting the importance of monitoring placement for spatial uncertainty reduction.

\begin{figure}[!htbp]
    \centering
    \captionsetup[subfigure]{singlelinecheck=off,justification=centering}
    \begin{minipage}[t]{0.2257\textwidth}
        \centering
        \includegraphics[width=\linewidth]{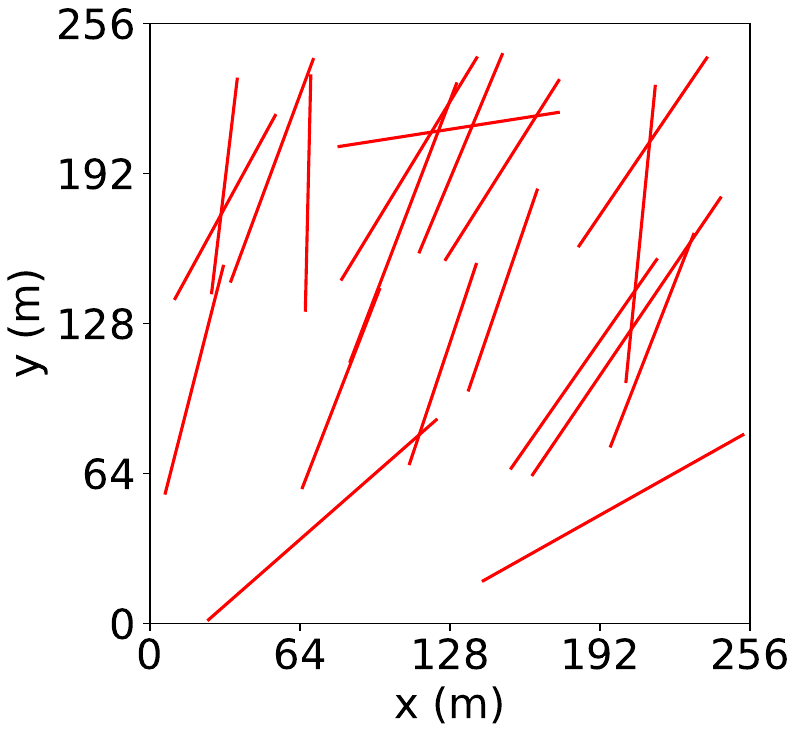}
        \subcaption{True DFN}
    \end{minipage}\hfill
    \begin{minipage}[t]{0.1741\textwidth}
        \centering
        \includegraphics[width=\linewidth]{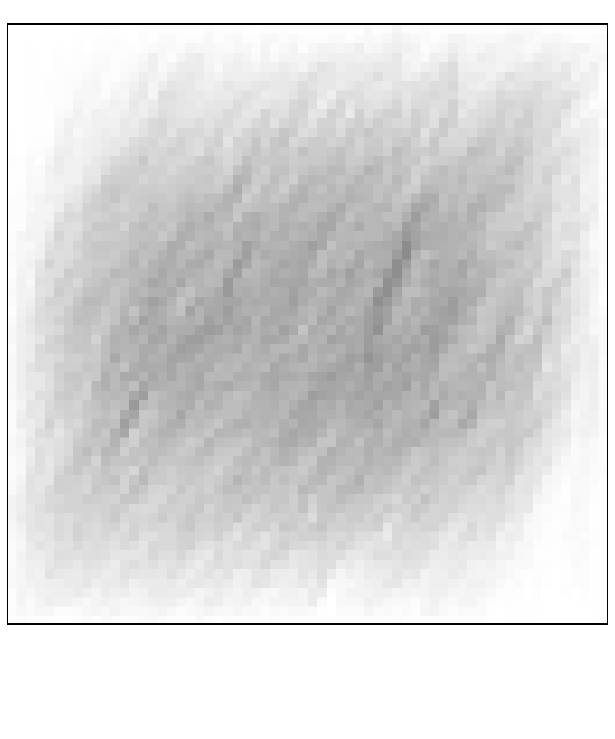}
        \subcaption{Prior}
    \end{minipage}\hfill
    \begin{minipage}[t]{0.1741\textwidth}
        \centering
        \includegraphics[width=\linewidth]{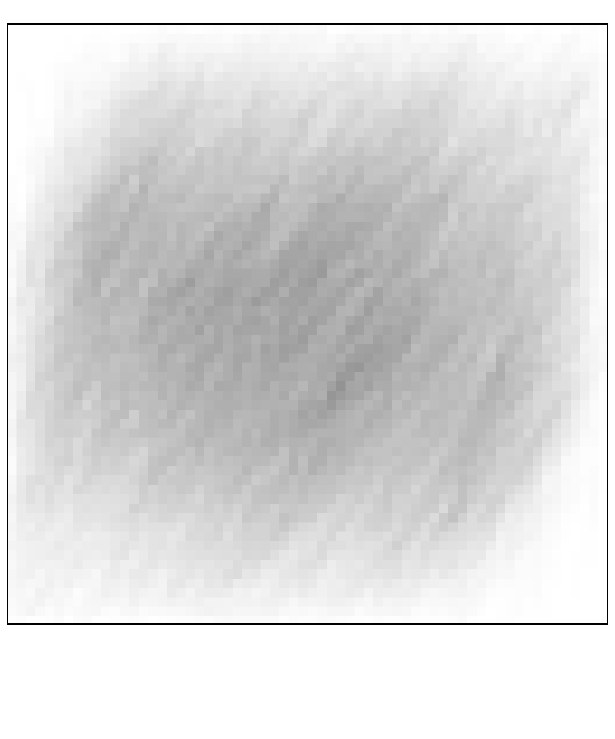}
        \subcaption{1 well}
    \end{minipage}\hfill
    \begin{minipage}[t]{0.1741\textwidth}
        \centering
        \includegraphics[width=\linewidth]{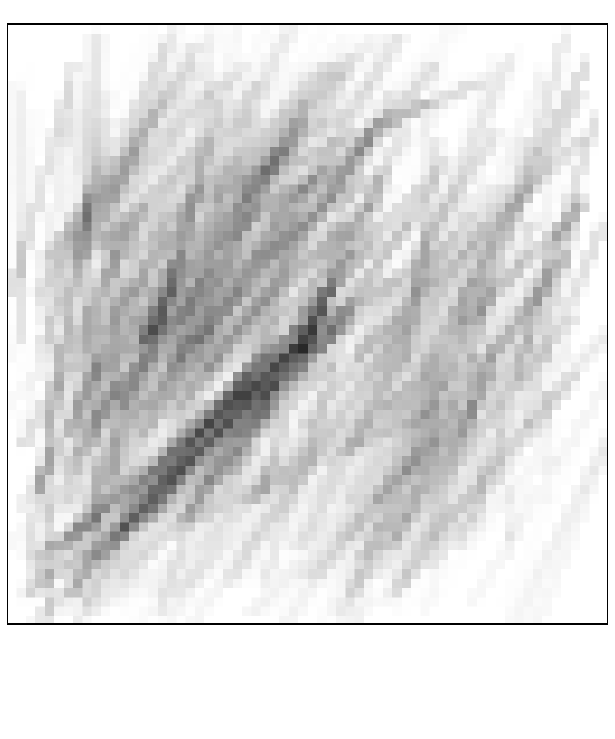}
        \subcaption{2 wells}
    \end{minipage}\hfill
    \begin{minipage}[t]{0.2282\textwidth}
        \centering
        \includegraphics[width=\linewidth]{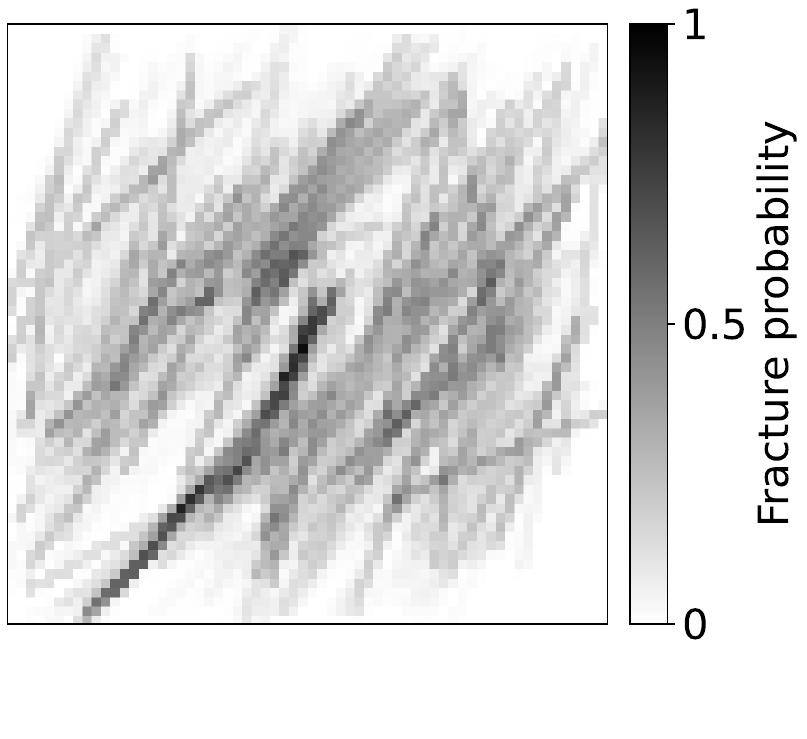}
        \captionsetup[subfigure]{margin={0pt,0.24\linewidth}}
        \subcaption{6 wells}
    \end{minipage}

    \medskip

    \begin{minipage}[t]{0.2257\textwidth}
        \centering
        \includegraphics[width=\linewidth]{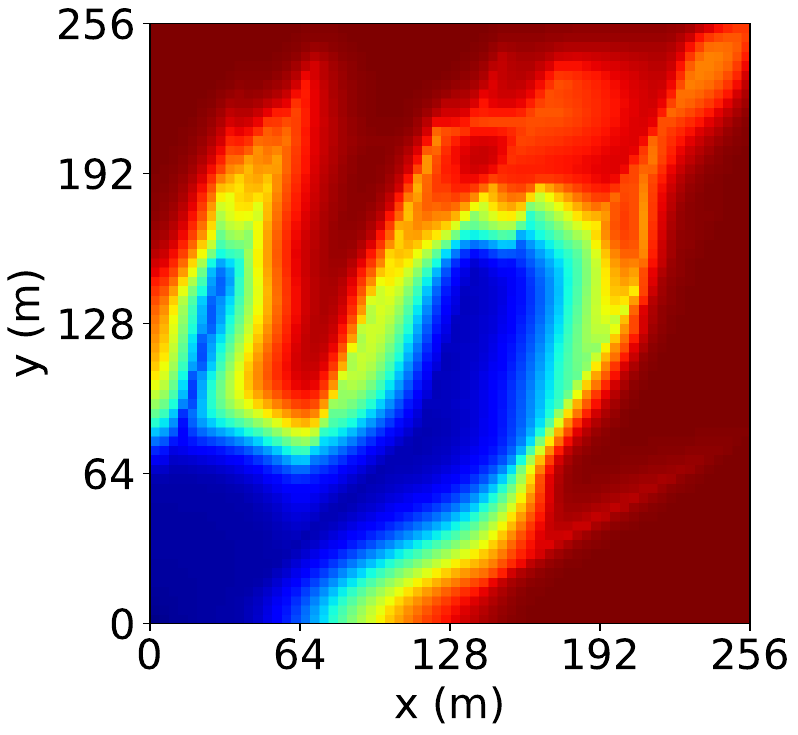}
        \subcaption{True temperature}
    \end{minipage}\hfill
    \begin{minipage}[t]{0.1741\textwidth}
        \centering
        \includegraphics[width=\linewidth]{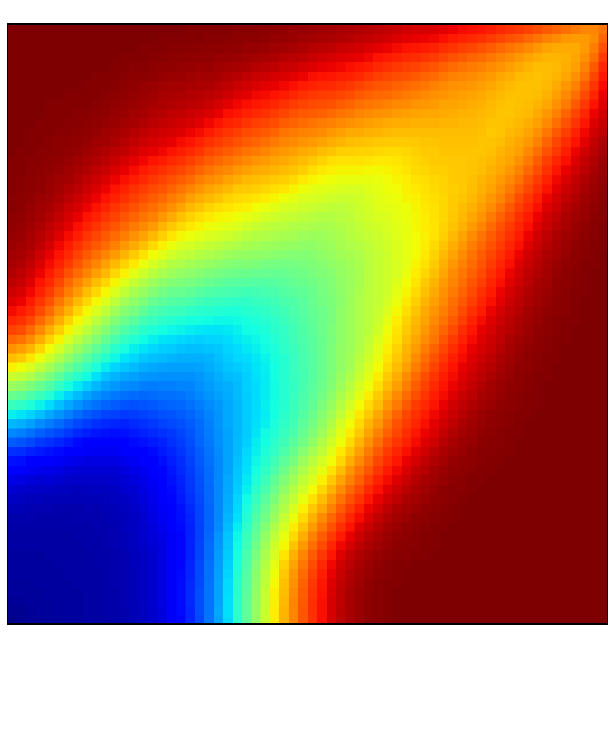}
        \subcaption{Prior}
    \end{minipage}\hfill
    \begin{minipage}[t]{0.1741\textwidth}
        \centering
        \includegraphics[width=\linewidth]{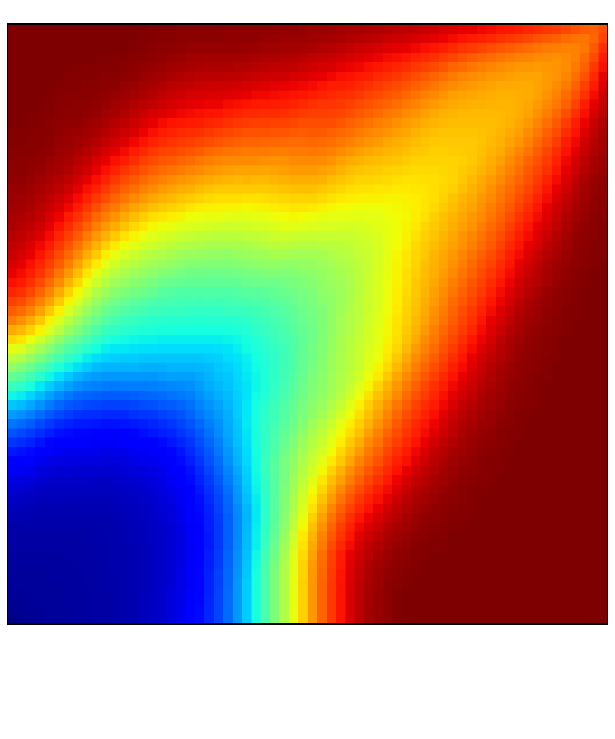}
        \subcaption{1 well}
    \end{minipage}\hfill
    \begin{minipage}[t]{0.1741\textwidth}
        \centering
        \includegraphics[width=\linewidth]{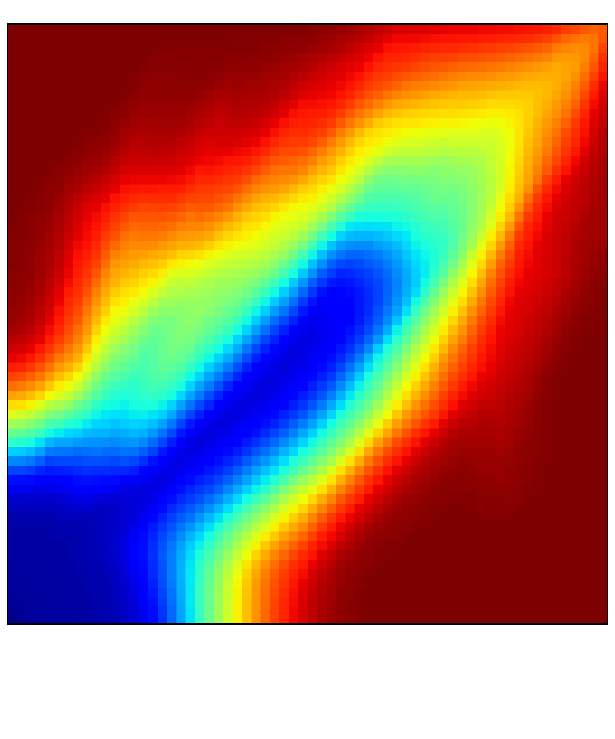}
        \subcaption{2 wells}
    \end{minipage}\hfill
    \begin{minipage}[t]{0.2320\textwidth}
        \centering
        \includegraphics[width=\linewidth]{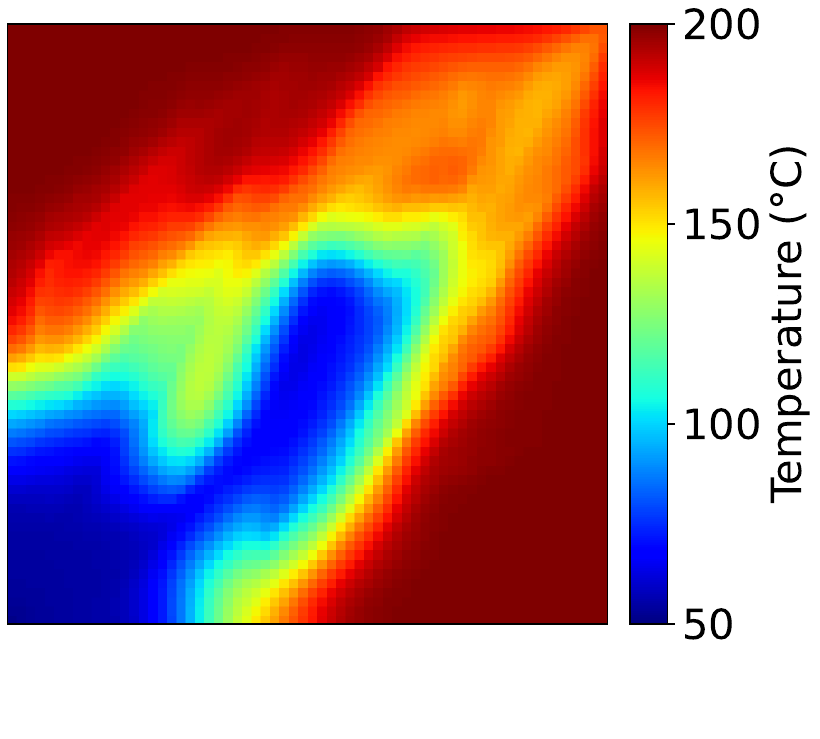}
        \captionsetup[subfigure]{margin={0pt,0.25\linewidth}}
        \subcaption{6 wells}
    \end{minipage}

    \caption{Prior and posterior fracture probability maps (top row) and mean temperature fields at 1800~days (bottom row) for the T+C setup with different monitoring configurations. The true DFN and temperature field are shown in (a) and (f). Prior and posterior mean temperature fields are computed using 200 and 500 realizations, respectively.}
    \label{fig:12_posterior_maps}
\end{figure}

Figure~\ref{fig:12_posterior_maps} shows the prior and posterior fracture probability maps for the three monitoring configurations. Fracture probability is calculated as the fraction of DFN realizations containing a fracture within each matrix cell. With observations at M1 only, the posterior fracture probability remains similar to the prior, indicating that the observations provide limited constraints on fracture locations despite the narrower posterior range of producer temperature predictions. Adding observations at M2 leads to more concentrated fracture probabilities along several linear features. With six monitoring locations, the posterior shows a more distinct high-probability feature extending from the lower-left portion of the domain toward the center, in reasonable agreement with the true DFN. However, fracture locations remain uncertain in other parts of the domain.

The corresponding mean temperature fields at 1800~days are shown in the second row of Fig.~\ref{fig:12_posterior_maps}. The prior and posterior means are computed from surrogate predictions for 200 prior and 500 posterior realizations, respectively, and the true temperature field is obtained from TRGB simulation. The one-well configuration provides limited improvement in the spatial temperature prediction. With two monitoring locations, the posterior mean captures the general direction of the cooled region, but discrepancies remain in its spatial extent. The six-well configuration provides better agreement with the true temperature field, particularly in the extent and spatial structure of the cooled region. These results indicate that additional monitoring locations provide information about fracture geometry and improve spatial temperature predictions. However, accurate predictions at the monitored locations do not necessarily imply accurate characterization of the entire fracture network.

\begin{figure}[!htbp]
    \centering
    \begin{minipage}[c]{0.8\textwidth}
        \centering
        \includegraphics[width=\linewidth]{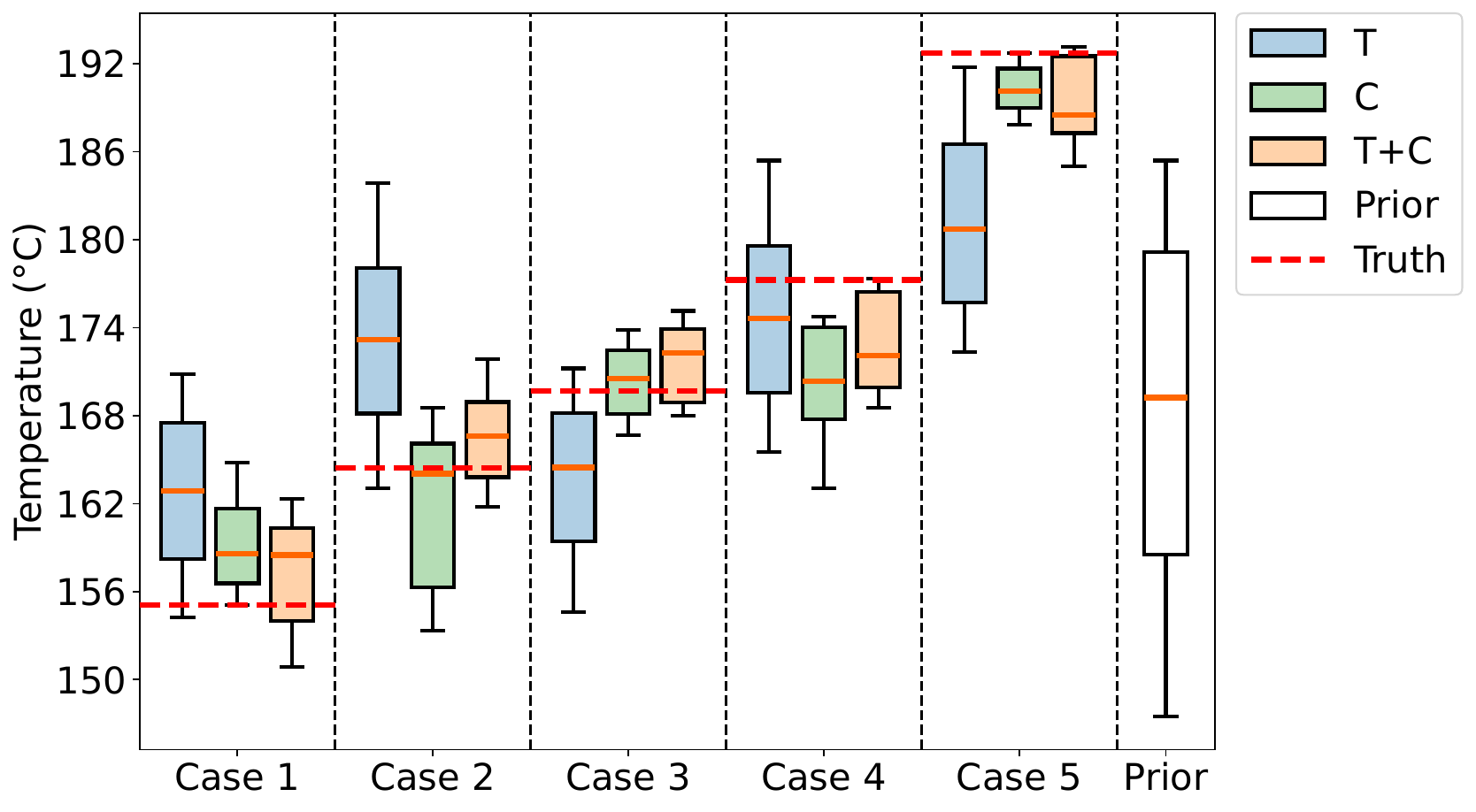}
    \end{minipage}

    \caption{Prior and posterior producer temperature distributions at 1800~days for five true DFN realizations using six monitoring locations. White, blue, green, and orange boxes denote the prior, T, C, and T+C results, respectively. Boxes indicate the $P_{\mathrm{25}}$-$P_{\mathrm{75}}$ ranges, whiskers extend from $P_{\mathrm{10}}$ to $P_{\mathrm{90}}$, and orange lines indicate the medians. Red dashed lines denote the true temperatures.}
    \label{fig:13_multiple_cases}
\end{figure}

To further evaluate the history matching performance, five true DFN realizations are considered using the six-well monitoring configuration. The true models are selected to span the prior distribution of producer temperature at 1800~days, with one realization from each of the five quintiles. Case 1 has the lowest producer temperature and Case 5 the highest. Case 4 corresponds to the true model considered above. Figure~\ref{fig:13_multiple_cases} compares the prior and posterior producer temperature distributions for the three observation setups. Assimilation of temperature data alone provides limited uncertainty reduction, with relatively broad posterior ranges. More substantial uncertainty reduction is observed with the C and T+C setups, and the posterior predictions show better agreement with the true temperatures for most cases. Some discrepancies remain, particularly for Cases 1 and 4. Overall, the framework provides reasonable history matching results across different fracture network realizations, demonstrating its applicability to cases with different fracture geometries and thermal responses. The results also highlight the importance of tracer observations for reducing uncertainty in long-term producer temperature predictions.

\section{Concluding Remarks}
\label{sec:6_conclusions}

In this work, we developed a deep-learning-based framework for fracture network inversion that combines vertex-based flow matching, geometry-aware surrogate modeling, and Bayesian data assimilation. The flow matching model provides a generative parameterization of fracture geometries through a mapping from Gaussian latent variables to fracture vertex coordinates. The geometry-aware surrogate model incorporates EDFM connectivity through graph neural networks and attention mechanisms to predict spatially and temporally varying flow and transport responses. These models are integrated with PT-MCMC to perform history matching in the latent space and quantify uncertainty in fracture geometries and corresponding reservoir responses.

The proposed framework was applied to a two-dimensional enhanced geothermal system with uncertain fracture geometries. A total of 5,000 DFN realizations were generated for model training and evaluation. The flow matching model reproduced the geometric statistics of the training fracture networks, as well as the distributions of temperature and tracer concentration responses. The geometry-aware surrogate model provided accurate predictions of temperature and tracer concentration fields for new DFN realizations, with mean absolute errors of $0.94~^\circ\mathrm{C}$ and 0.0051~mg/kg. The surrogate model also reproduced the statistical distributions of producer responses, enabling its use for Bayesian history matching.

History matching was performed using different observation types and monitoring configurations. Tracer observations provided useful information about fracture connectivity and substantially reduced uncertainty in long-term producer temperature predictions for the cases considered. Observations at the production well alone constrained producer temperature predictions but provided limited information about fracture locations and temperature distributions elsewhere in the reservoir. Additional monitoring locations improved fracture characterization and spatial temperature predictions, although uncertainty in fracture geometry remained. The framework also provided reasonable history matching results for multiple true DFN realizations with different fracture geometries and thermal responses. These results demonstrate the importance of observation type and monitoring configuration for fracture network inversion and highlight the distinction between reducing uncertainty in reservoir predictions and constraining the underlying fracture geometry.

Future work will consider extensions to three-dimensional fracture networks and more complex reservoir configurations. Incorporating monitoring optimization into the Bayesian inversion framework could help identify observation locations that provide the most information about fracture geometries and reservoir responses. Assimilation of additional data types, including pressure and geophysical measurements, may further constrain fracture connectivity and reduce the nonuniqueness of the inverse problem. These extensions could support application of the framework to more realistic fractured subsurface systems.
%%%%%%%%%%%%%%%%%%%%%%%%%%%%%%%%%%%%%%%%%%%%%%%
%
% DATA SECTION and ACKNOWLEDGMENTS
%
%%%%%%%%%%%%%%%%%%%%%%%%%%%%%%%%%%%%%%%%%%%%%%%

\section*{Open Research Section}
The DFN realizations, simulated data, and training results are available at Zenodo (https://doi.org/10.5281/zenodo.24655532; \citeA{ouyang2026data}). The code used for fracture network generation, geometry-aware surrogate modeling, and Bayesian inversion is available at https://github.com/QUIUQ/DFNinversion. 

\section*{Conflict of Interest disclosure}
The authors declare there are no conflicts of interest for this manuscript.

\acknowledgments
The authors acknowledge financial support from the Department of Civil and Environmental Engineering at Carnegie Mellon University and computational resources provided by the ORCHARD computing cluster.

%%%%%%%%%%%%%%%%%%%%%%%%%%%%%%%%%%%%%%%%%%%%%%%
% REFERENCES and BIBLIOGRAPHY
%
% \bibliography{<name of your .bib file>} don't specify the file extension
% don't specify bibliographystyle
%
%%%%%%%%%%%%%%%%%%%%%%%%%%%%%%%%%%%%%%%%%%%%%%%

%\bibliography{ enter your bibtex bibliography filename here }
\bibliography{references}

\clearpage

\title{Supporting Information for ``Generative Model-Based Parameterization and Geometry-Aware Surrogate Model for Fracture Network Inversion''}

\noindent\textbf{Contents of this file}
\begin{enumerate}
\item Text S1 to S4
\item Tables S1 to S5
\item Figures S1 to S6
\end{enumerate}

\noindent\textbf{Introduction}

This Supporting Information provides additional details on the methods and numerical results presented in the main text. Text S1 and Table S1 describe the architecture, training, and sampling procedure of the vertex-based flow matching model. Text S2 and Tables S2-S5 describe the architecture and training of the geometry-aware surrogate model. Text S3 describes the PT-MCMC and ESMDA implementations. Text S4 and Figures S1-S6 present the comparison between PT-MCMC and ESMDA and representative prior and posterior DFN realizations.

\noindent\textbf{Text S1. Flow Matching Architecture, Training, and Sampling}

The flow matching model described in Section~2.2 of the main text uses a neural network to learn the velocity field that transforms Gaussian latent variables into normalized fracture coordinates. Here, we provide details of the network architecture, training procedure, and numerical sampling. The velocity network uses a transformer architecture \cite{vaswani2017attention}, with the layer-by-layer operations and output dimensions provided in Table~S1. The input vector $\mbox{\boldmath$\xi$}_\tau\in\mathbb{R}^{4N_{\mathrm{f}}}$ is reshaped into $N_{\mathrm{f}}$ fracture tokens, each containing four coordinate components. The fracture tokens are mapped to 128-dimensional embeddings through a linear layer. The flow time $\tau$ is embedded separately using two linear layers with a sigmoid linear unit (SiLU) activation and appended as an additional token. The resulting $N_{\mathrm{f}}+1$ tokens are processed through two transformer layers, each containing four-head self-attention and a feed-forward network. Rotary position embedding (RoPE) \cite{su2024roformer} is used to encode the positions of the fracture tokens in the prescribed ordering. Residual connections, layer normalization, and dropout with a rate of 0.1 are used in the transformer layers. The time token is then removed, and a final linear layer maps each fracture token to four velocity components.

\begin{table}[!ht]
% \settablenum{S1}
\caption*{Table S1: Architecture of the transformer-based velocity network used in the vertex-based flow matching model. The network contains 414,468 trainable parameters.}
\centering
\begin{tabular}[t]{lll}
\hline
Component & Operation & Output dimensions \\
\hline
Input
 & Fracture coordinates $\mbox{\boldmath$\xi$}_\tau$ & $(N_{\mathrm{f}},4)$ \\
 & Flow time $\tau$ & $(1)$ \\
\hline
Embedding
 & Linear projection, $4 \rightarrow 128$ & $(N_{\mathrm{f}},128)$ \\
 & Time embedding, $1 \rightarrow 128 \rightarrow 128$ (SiLU) & $(1,128)$ \\
 & Token concatenation & $(N_{\mathrm{f}}+1,128)$ \\
\hline
Transformer ($\times 2$)
 & Self-attention, 4 heads, RoPE & $(N_{\mathrm{f}}+1,128)$ \\
 & Layer normalization & $(N_{\mathrm{f}}+1,128)$ \\
 & Feed-forward network, $128 \rightarrow 512 \rightarrow 128$ (ReLU) & $(N_{\mathrm{f}}+1,128)$ \\
 & Layer normalization & $(N_{\mathrm{f}}+1,128)$ \\
\hline
Output
 & Remove time token & $(N_{\mathrm{f}},128)$ \\
 & Linear projection, $128 \rightarrow 4$ & $(N_{\mathrm{f}},4)$ \\
\hline
\end{tabular}
\end{table}

The network is trained using 5,000 DFN realizations described in Section~2.1 of the main text. The sorted fracture vertex coordinates are normalized component-wise using their means and standard deviations over the training realizations. The normalized vectors provide the target samples for training. The network parameters are optimized by minimizing the flow matching loss in Eq.~(2) of the main text, using the Adam optimizer \cite{kingma2015adam} with a learning rate of $2\times10^{-4}$ and a minibatch size of 64. Training is performed for 1,000 epochs and requires approximately 13 minutes on a single NVIDIA H100 GPU.

New DFN realizations are generated by integrating the ordinary differential equation (ODE) in Eq.~(3) of the main text using the explicit Euler method with $N_\tau=100$ steps and $\Delta\tau=1/N_\tau$:
\begin{equation}
\mbox{\boldmath$\xi$}_{\tau+\Delta\tau}
=
\mbox{\boldmath$\xi$}_\tau
+
\Delta\tau\,\mathbf{v}_{\phi^*}(\mbox{\boldmath$\xi$}_\tau,\tau),
\label{eq:si_euler}
\end{equation}
where $\phi^*$ denotes the trained velocity-network parameters. The terminal state $\mbox{\boldmath$\xi$}_{\tau=1}$ is denormalized using the training statistics to obtain the physical fracture vertex coordinates.

\noindent\textbf{Text S2. Surrogate Model Architecture and Training}

The geometry-aware surrogate model is described in Section~3.2 of the main text, with network architectures and output dimensions summarized in Tables~S2-S5. The geometry encoder contains four rounds of local graph neural network (GNN) and global attention updates, with four GNN layers per round. The first GNN layer uses separate multilayer perceptrons (MLPs) for different node and edge types, while the remaining layers follow Eqs.~(7) and (8) of the main text. The node and edge latent features have 32 dimensions. After the fourth round, a final node-to-feature update produces the encoded geometry features.

\begin{table}[!ht]
% \settablenum{S2}
\caption*{Table S2: Overall architecture of the geometry-aware surrogate model. Here $N_{\mathrm{c}}=4096$, $N_{\mathrm{fe}}$, $N_{\mathrm{e}}$, $N_{\mathrm{f}}$, and $N_{\mathrm{p}}=256$ denote the numbers of matrix cells, fracture elements, directed edges, fractures, and matrix patches, respectively. The number of prediction times is $N_{\mathrm{t}}=60$ for temperature and 48 for tracer concentration. Each encoder round contains four GNN layers, including the initial GNN layer in the first round. Detailed architectures are provided in Tables~S3--S5. Each surrogate network contains approximately $4.38\times10^6$ trainable parameters.}
\centering
\begin{tabular}[t]{l l l}
\hline
Network & Layer & Output \\
\hline
Input
 & matrix-cell features & $(N_{\mathrm{c}},4)$ \\
 & fracture-element features & $(N_{\mathrm{fe}},7)$ \\
 & directed-edge features & $(N_{\mathrm{e}},11)$ \\
\hline
Geometry encoder
 & matrix-patch latent feature construction & $(N_{\mathrm{p}},128)$ \\
 & fracture latent feature construction & $(N_{\mathrm{f}},128)$ \\
 & fracture self-attention with pairwise bias ($\times4$)
 & $(N_{\mathrm{f}},128)$ \\
\hline
Encoder round
 & local GNN (4 layers) & $(N_{\mathrm{c}}+N_{\mathrm{fe}},32)$ \\
($\times4$)
 & global self-attention & $(N_{\mathrm{p}}+N_{\mathrm{f}},128)$ \\
 & node-to-feature exchange & $(N_{\mathrm{p}}+N_{\mathrm{f}},128)$ \\
 & feature-to-node exchange & $(N_{\mathrm{c}}+N_{\mathrm{fe}},32)$ \\
\hline
Encoder output
 & final node-to-feature update & $(N_{\mathrm{p}}+N_{\mathrm{f}},128)$ \\
\hline
Spatial-temporal
 & input: encoded geometry features & $(N_{\mathrm{p}}+N_{\mathrm{f}},128)$ \\
decoder
 & spatial query & $(N_{\mathrm{c}},32)$ \\
 & temporal query & $(N_{\mathrm{t}},128)$ \\
 & feature-wise linear modulation (FiLM)
 & $(N_{\mathrm{t}},N_{\mathrm{c}},32)$ \\
 & spatial-temporal attention ($\times2$)
 & $(N_{\mathrm{t}},N_{\mathrm{c}},32)$ \\
 & grid reconstruction & $(N_{\mathrm{t}},N_{\mathrm{c}})$ \\
\hline
\end{tabular}
\end{table}

\begin{table}[!ht]
% \settablenum{S3}
\caption*{Table S3: Architecture of the local graph neural network (GNN) in the geometry encoder. Here $N_{\mathrm{c}}$, $N_{\mathrm{fe}}$, $N_{\mathrm{e}}$, $N_{\mathrm{e}}^{\mathrm{mm}}$, $N_{\mathrm{e}}^{\mathrm{mf}}$, and $N_{\mathrm{e}}^{\mathrm{ff}}$ denote the numbers of matrix cells, fracture elements, directed edges, matrix-matrix edges, matrix-fracture edges, and fracture-fracture edges, respectively. Each encoder round contains four GNN layers. The first layer of the first round uses separate multilayer perceptrons (MLPs) for different node and edge types. The remaining 15 layers follow Eqs.~(7) and (8) of the main text.}
\centering
\begin{tabular}[t]{l l l}
\hline
Network & Layer & Output \\
\hline
Input
 & matrix-cell features & $(N_{\mathrm{c}},4)$ \\
 & fracture-element features & $(N_{\mathrm{fe}},7)$ \\
 & directed-edge features & $(N_{\mathrm{e}},11)$ \\
\hline
Initial GNN layer
 & edge MLP, matrix-matrix (19 inputs) & $(N_{\mathrm{e}}^{\mathrm{mm}},32)$ \\
 & edge MLP, matrix-fracture (22 inputs) & $(N_{\mathrm{e}}^{\mathrm{mf}},32)$ \\
 & edge MLP, fracture-fracture (25 inputs) & $(N_{\mathrm{e}}^{\mathrm{ff}},32)$ \\
 & concatenated edge features & $(N_{\mathrm{e}},32)$ \\
 & matrix-node MLP (36 inputs) & $(N_{\mathrm{c}},32)$ \\
 & fracture-node MLP (39 inputs) & $(N_{\mathrm{fe}},32)$ \\
 & concatenated node features & $(N_{\mathrm{c}}+N_{\mathrm{fe}},32)$ \\
\hline
GNN layer
 & edge update: MLP, two linear layers, GELU
 & $(N_{\mathrm{e}},32)$ \\
 & node update: MLP, two linear layers, GELU
 & $(N_{\mathrm{c}}+N_{\mathrm{fe}},32)$ \\
\hline
\end{tabular}
\end{table}

\begin{table}[!ht]
% \settablenum{S4}
\caption*{Table S4: Architecture of the global attention module in the geometry encoder. Matrix-patch and fracture latent feature vectors have 128 dimensions. The fracture self-attention uses pairwise geometric information to modify the attention scores, as described in Text S2. The module also exchanges information with the local GNN through node-to-feature and feature-to-node operations.}
\centering
\begin{tabular}[t]{l l l}
\hline
Network & Layer & Output \\
\hline
Matrix-patch latent features
 & input: matrix-node features & $(N_{\mathrm{c}},4)$ \\
 & MLP, two linear layers, GELU & $(N_{\mathrm{c}},128)$ \\
 & $4\times4$ convolution, stride 4 & $(N_{\mathrm{p}},128)$ \\
\hline
Fracture latent features
 & input: fracture-node features & $(N_{\mathrm{fe}},7)$ \\
 & MLP, two linear layers, GELU & $(N_{\mathrm{fe}},128)$ \\
 & weighted pooling over fracture elements & $(N_{\mathrm{f}},128)$ \\
 & linear layer & $(N_{\mathrm{f}},128)$ \\
\hline
Fracture self-attention
 & pairwise geometric features & $(N_{\mathrm{f}},N_{\mathrm{f}},5)$ \\
 & pairwise bias MLP, SiLU & $(N_{\mathrm{f}},N_{\mathrm{f}},8)$ \\
 & self-attention, 8 heads ($\times4$) & $(N_{\mathrm{f}},128)$ \\
 & feed-forward, 512 hidden units, GELU ($\times4$) & $(N_{\mathrm{f}},128)$ \\
 & layer normalization & $(N_{\mathrm{f}},128)$ \\
\hline
Global attention
 & type and positional embeddings & $(N_{\mathrm{p}}+N_{\mathrm{f}},128)$ \\
 & self-attention, 8 heads & $(N_{\mathrm{p}}+N_{\mathrm{f}},128)$ \\
 & feed-forward, 512 hidden units, GELU
 & $(N_{\mathrm{p}}+N_{\mathrm{f}},128)$ \\
 & layer normalization & $(N_{\mathrm{p}}+N_{\mathrm{f}},128)$ \\
\hline
Node-to-feature exchange
 & matrix nodes: convolution, stride 4 & $(N_{\mathrm{p}},128)$ \\
 & fracture nodes: weighted pooling & $(N_{\mathrm{f}},128)$ \\
\hline
Feature-to-node exchange
 & matrix latent features: transposed convolution
 & $(N_{\mathrm{c}},32)$ \\
 & fracture latent features: linear projection
 & $(N_{\mathrm{fe}},32)$ \\
\hline
Encoder output
 & final node-to-feature update
 & $(N_{\mathrm{p}}+N_{\mathrm{f}},128)$ \\
 & layer normalization
 & $(N_{\mathrm{p}}+N_{\mathrm{f}},128)$ \\
\hline
\end{tabular}
\end{table}

\begin{table}[!ht]
% \settablenum{S5}
\caption*{Table S5: Architecture of the spatial-temporal decoder. The decoder uses the encoded geometry features to predict temperature or tracer concentration at $N_{\mathrm{t}}$ times and $N_{\mathrm{c}}$ matrix cells. Convolutional layers operate on the $64\times64$ matrix grid. The output dimensions are given in the order of time, matrix cells, and features.}
\centering
\begin{tabular}[t]{l l l}
\hline
Network & Layer & Output \\
\hline
Input
 & encoded geometry features
 & $(N_{\mathrm{p}}+N_{\mathrm{f}},128)$ \\
\hline
Spatial query
 & location MLP, two linear layers, GELU
 & $(N_{\mathrm{c}},128)$ \\
 & cross-attention, 8 heads ($\times4$)
 & $(N_{\mathrm{c}},128)$ \\
 & feed-forward, 512 hidden units, GELU ($\times4$)
 & $(N_{\mathrm{c}},128)$ \\
 & linear projection & $(N_{\mathrm{c}},32)$ \\
\hline
Temporal query
 & time MLP, two linear layers, GELU & $(N_{\mathrm{t}},32)$ \\
 & linear projection & $(N_{\mathrm{t}},128)$ \\
 & cross-attention, 8 heads & $(N_{\mathrm{t}},128)$ \\
\hline
FiLM
 & layer normalization, linear projection
 & $(N_{\mathrm{t}},64)$ \\
 & scale and shift of spatial features
 & $(N_{\mathrm{t}},N_{\mathrm{c}},32)$ \\
\hline
Spatial-temporal
 & causal temporal self-attention, 4 heads
 & $(N_{\mathrm{t}},N_{\mathrm{c}},32)$ \\
attention ($\times2$)
 & feed-forward, 128 hidden units, GELU
 & $(N_{\mathrm{t}},N_{\mathrm{c}},32)$ \\
 & $4\times4$ convolution, stride 4
 & $(N_{\mathrm{t}},N_{\mathrm{p}},128)$ \\
 & spatial self-attention, 4 heads
 & $(N_{\mathrm{t}},N_{\mathrm{p}},128)$ \\
 & feed-forward, 512 hidden units, GELU
 & $(N_{\mathrm{t}},N_{\mathrm{p}},128)$ \\
 & $4\times4$ transposed convolution, stride 4
 & $(N_{\mathrm{t}},N_{\mathrm{c}},32)$ \\
\hline
Grid reconstruction
 & pointwise MLP, GELU
 & $(N_{\mathrm{t}},N_{\mathrm{c}},1)$ \\
 & spatial feature projection
 & $(N_{\mathrm{c}},32)$ \\
 & feature concatenation
 & $(N_{\mathrm{t}},N_{\mathrm{c}},64)$ \\
 & $3\times3$ convolution, GELU
 & $(N_{\mathrm{t}},N_{\mathrm{c}},32)$ \\
 & $3\times3$ convolution, GELU
 & $(N_{\mathrm{t}},N_{\mathrm{c}},32)$ \\
 & $1\times1$ convolution
 & $(N_{\mathrm{t}},N_{\mathrm{c}},1)$ \\
 & output summation; softplus for tracer
 & $(N_{\mathrm{t}},N_{\mathrm{c}})$ \\
\hline
\end{tabular}
\end{table}

Before the encoder rounds, the fracture latent features are processed through four self-attention layers with pairwise geometric bias. For fractures $i$ and $j$, the geometric relation is represented by $\mbox{\boldmath$\psi$}_{ij}=[\Delta x_{ij},\Delta y_{ij},D_{ij},|\cos\theta_{ij}|,I_{ij}]$, where $(\Delta x_{ij},\Delta y_{ij})$ is the signed offset from the midpoint of fracture $i$ to that of fracture $j$, $D_{ij}$ is the minimum distance between the two fracture segments, $\theta_{ij}$ is their relative orientation angle, and $I_{ij}$ indicates whether the fractures intersect. An MLP maps these features to a bias value $B_{ij}^{(h)}$ for each attention head $h$. The attention operation is given by
\begin{equation}
\mathrm{Attention}^{(h)}
=
\mathrm{softmax}
\left(
\frac{Q^{(h)}(K^{(h)})^{\mathsf{T}}}{\sqrt{d_{\mathrm{k}}}}
+
B^{(h)}
\right)V^{(h)},
\label{eq:si_bias}
\end{equation}
where $Q^{(h)}$, $K^{(h)}$, and $V^{(h)}$ are the query, key, and value matrices for head $h$, $d_{\mathrm{k}}$ is the query and key dimension, and $B^{(h)}$ contains the pairwise biases $B_{ij}^{(h)}$. This formulation incorporates relative fracture geometry into the attention weights.

In the spatial-temporal decoder, the scale vector in Eq.~(10) of the main text includes an offset of one, while the shift vector includes the temporal query embedding $\mathbf{u}_j$. The output fields are reconstructed using a pointwise MLP and a convolutional correction consisting of two $3\times3$ convolutional layers followed by a $1\times1$ layer on the $64\times64$ matrix grid. The convolutional correction accounts for local spatial variations in the predicted responses.

The input features are normalized prior to training. Coordinates are scaled by the domain length, while distances and connection areas are scaled by the matrix cell width and the matrix-matrix connection area, respectively. Permeabilities, volumes, and fracture apertures are log-transformed relative to the matrix permeability, the matrix cell volume, and the prescribed aperture, respectively. Temperature and tracer concentration are normalized using the corresponding means and standard deviations from the training dataset.

% VERIFY WITH IMPLEMENTATION: whether T_i and Tbar_i in L_d are physical or normalized, and the exact averaging axis used for Tbar_i.
%T_i and Tbar_i in L_d are physical temperatures (in °C), not normalized. Tbar_i is the mean simulated temperature in matrix cell i, averaged over all realizations and times in the minibatch.

The surrogate networks are trained using the loss function in Eq.~(11) of the main text, with $w_1=w_2=0.25$. For temperature, the data, spatial, and temporal loss terms are defined as
\begin{equation}
\begin{aligned}
L_{\mathrm{d}} &=
\frac{1}{N_{\mathrm{c}}N_{\mathrm{t}}}
\sum_{i=1}^{N_{\mathrm{c}}}
\sum_{j=1}^{N_{\mathrm{t}}}
\frac{\left|\hat{T}_i(t_j)-T_i(t_j)\right|}{\bar{T}_i},\\
L_{\mathrm{s}} &=
\frac{1}{N_{\mathrm{c}}N_{\mathrm{t}}}
\sum_{j=1}^{N_{\mathrm{t}}}
\left\|\nabla\hat{\mathbf{T}}_j-\nabla\mathbf{T}_j\right\|_1,\\
L_{\mathrm{t}} &=
\frac{1}{N_{\mathrm{c}}(N_{\mathrm{t}}-1)}
\sum_{j=1}^{N_{\mathrm{t}}-1}
\left\|
\frac{d\hat{\mathbf{T}}_j}{dt}
-
\frac{d\mathbf{T}_j}{dt}
\right\|_1,
\end{aligned}
\label{eq:si_loss_temperature}
\end{equation}
where $\hat{T}_i(t_j)$ and $T_i(t_j)$ are the predicted and simulated temperatures in matrix cell $i$ at time $t_j$, and $\bar{T}_i$ is the mean simulated temperature in cell $i$ over the minibatch. The vectors $\hat{\mathbf{T}}_j$ and $\mathbf{T}_j$ contain the normalized predicted and simulated temperature fields at time $t_j$, respectively. The spatial gradients and temporal derivatives are approximated using differences between neighboring matrix cells and consecutive prediction times.
% VERIFY WITH IMPLEMENTATION: whether differences include spatial grid spacing and time interval normalization.
%The spatial difference uses the difference between neighboring cells without dividing by the grid spacing. The temporal difference is divided by the time interval (in days). Both use the normalized fields.

For tracer concentration, $L_{\mathrm{s}}$ and $L_{\mathrm{t}}$ follow Eq.~(S3), with temperature replaced by concentration. The data loss is the mean squared error of the normalized concentration fields, with an additional term for the six monitoring locations weighted by 0.1.

Separate networks with the same architecture are trained for temperature and tracer concentration using the AdamW optimizer \cite{loshchilov2019decoupled}. Both networks use a weight decay of $10^{-3}$ and an initial learning rate of $2\times10^{-4}$. The learning rate is held constant for the first 500 epochs and subsequently reduced by a factor of 0.9 when the validation loss stops decreasing. The temperature network is trained on one NVIDIA H100 GPU with a batch size of 4, requiring approximately 1,300 epochs and 57 hours. Training is terminated when the learning rate reaches $5\times10^{-5}$. The tracer network is trained on three NVIDIA H100 GPUs with a batch size of 12, requiring approximately 1,500 epochs and 35 hours. For both networks, the parameters corresponding to the lowest validation loss are retained.

\noindent\textbf{Text S3. PT-MCMC and ESMDA Implementations}

We compare parallel tempering Markov chain Monte Carlo (PT-MCMC) and the ensemble smoother with multiple data assimilation (ESMDA) for fracture network inversion. PT-MCMC is used as a reference for posterior characterization, while ESMDA provides a less computationally expensive alternative. Both methods use the same latent parameterization, surrogate models, observations, and error covariance described in Section~4 of the main text.

For PT-MCMC, the data mismatch is defined as
\begin{equation}
\Phi(\mbox{\boldmath$\xi$})=
\frac{1}{2}
\left(\mathbf{d}_{\mathrm{obs}}-\hat{\mathbf{d}}_{\mathrm{obs}}(\mbox{\boldmath$\xi$})\right)^{\mathsf{T}}
C_{\mathrm{tot}}^{-1}
\left(\mathbf{d}_{\mathrm{obs}}-\hat{\mathbf{d}}_{\mathrm{obs}}(\mbox{\boldmath$\xi$})\right),
\label{eq:si_mismatch}
\end{equation}
such that the likelihood in Eq.~(15) of the main text is proportional to $\exp[-\Phi(\mbox{\boldmath$\xi$})]$. At each iteration, a preconditioned Crank-Nicolson (pCN) proposal is generated at each inverse-temperature level $\ell$ using Eq.~(16) of the main text, with step size $\beta_\ell$. The proposed latent vector $\mbox{\boldmath$\xi$}_\ell'$ is accepted with probability
\begin{equation}
\min\left\{1,\exp\left[
\omega_\ell\left(
\Phi(\mbox{\boldmath$\xi$}_\ell)
-\Phi(\mbox{\boldmath$\xi$}_\ell')
\right)\right]\right\},
\label{eq:si_accept}
\end{equation}
where $\mbox{\boldmath$\xi$}_\ell$ is the current latent vector and $\omega_\ell$ is the inverse temperature. Exchanges between neighboring levels $\ell$ and $\ell+1$ are accepted with probability
\begin{equation}
\min\left\{1,\exp\left[
(\omega_\ell-\omega_{\ell+1})
\left(
\Phi(\mbox{\boldmath$\xi$}_\ell)
-\Phi(\mbox{\boldmath$\xi$}_{\ell+1})
\right)\right]\right\}.
\label{eq:si_swap}
\end{equation}
Exchange attempts alternate between odd-even and even-odd neighboring pairs in successive iterations.

The pCN step sizes are adjusted during an initial adaptation period using a Robbins-Monro update \cite{robbins1951stochastic},
\begin{equation}
\beta_\ell\leftarrow\beta_\ell
\exp\left[\kappa_n(I_{\ell,n}-\bar{r})\right],
\label{eq:si_beta}
\end{equation}
where $I_{\ell,n}$ equals one if the proposal at level $\ell$ is accepted at iteration $n$ and zero otherwise, $\bar{r}$ is the target acceptance rate, and $\kappa_n$ is the adaptation gain. We use $N_\omega=8$ inverse temperatures geometrically spaced from $\omega_1=1$ to $\omega_8=10^{-3}$. The initial step sizes are $\beta_\ell=0.1/\sqrt{\omega_\ell}$, restricted to $[10^{-3},0.25]$. We set $\bar{r}=0.2$ and $\kappa_n=0.05/\sqrt{1+n/100}$. Adaptation is performed during the first 1,000 iterations, after which the step sizes are fixed. Five independent PT-MCMC runs are initialized from different prior samples and performed for 20,000 iterations each. The first 40\% of each run is discarded as burn-in, and the remaining samples at $\omega_1=1$ are pooled to characterize the posterior distribution.

The five PT-MCMC runs require a total of $8\times10^5$ surrogate evaluations, taking approximately 6.5~hours on a single NVIDIA H100 GPU. The corresponding TRGB simulations require approximately 5~minutes each on a single CPU core, equivalent to about 67,000 CPU-core hours.

For ESMDA \cite{emerick2013ensemble}, an ensemble of $N_{\mathrm{ens}}$ latent vectors $\mbox{\boldmath$\xi$}_j^0$, $j=1,\ldots,N_{\mathrm{ens}}$, is drawn from the standard Gaussian prior and updated over $N_{\mathrm{a}}$ assimilation steps according to
\begin{equation}
\mbox{\boldmath$\xi$}_j^{k+1}
=
\mbox{\boldmath$\xi$}_j^k
+
C_{\xi d}^{k}
\left(C_{dd}^{k}+\alpha_k C_{\mathrm{tot}}\right)^{-1}
\left(
\mathbf{d}_{\mathrm{obs}}
+\sqrt{\alpha_k}\,\mathbf{e}_j^k
-\hat{\mathbf{d}}_{\mathrm{obs}}(\mbox{\boldmath$\xi$}_j^k)
\right),
\label{eq:si_esmda}
\end{equation}
where $k=0,\ldots,N_{\mathrm{a}}-1$, $C_{\xi d}^{k}$ and $C_{dd}^{k}$ are the sample cross-covariance between latent variables and predicted data and the covariance of predicted data, respectively, $\mathbf{e}_j^k\sim\mathcal{N}(\mathbf{0},C_{\mathrm{tot}})$ is the observation perturbation at assimilation step $k$, and $\alpha_k$ is the inflation coefficient. We use $N_{\mathrm{ens}}=1000$ and $\alpha_k=N_{\mathrm{a}}$ for all assimilation steps, satisfying $\sum_{k=0}^{N_{\mathrm{a}}-1}\alpha_k^{-1}=1$. The same initial prior ensemble is used for all cases. ESMDA requires $N_{\mathrm{ens}}(N_{\mathrm{a}}+1)$ surrogate evaluations, compared with $8\times10^5$ for PT-MCMC.

\noindent\textbf{Text S4. Posterior Comparison and Uncertainty Characterization}

ESMDA provides a computationally efficient approach for fracture network inversion, but its ensemble-based updates may have difficulty characterizing non-Gaussian posterior distributions. We therefore compare ESMDA with PT-MCMC, which is used as a reference for assessing posterior uncertainty. We first examine the sensitivity of ESMDA to the number of data assimilations, $N_{\mathrm{a}}$, for Case~4 with the six-well monitoring configuration and both temperature and tracer concentration data (T+C).

Figure~S1 shows that the mean data mismatch decreases with increasing $N_{\mathrm{a}}$, although it remains higher than that of PT-MCMC. The fracture probability maps in Figure~S2 show limited uncertainty reduction for $N_{\mathrm{a}}=5$ and 10. For $N_{\mathrm{a}}=50$ and 100, the regions of high fracture probability become more localized and show reasonable agreement with the true fracture network. We use $N_{\mathrm{a}}=50$ for the subsequent comparisons.

\begin{figure}[!ht]
% \setfigurenum{S1}
\centering
\includegraphics[width=0.55\textwidth]{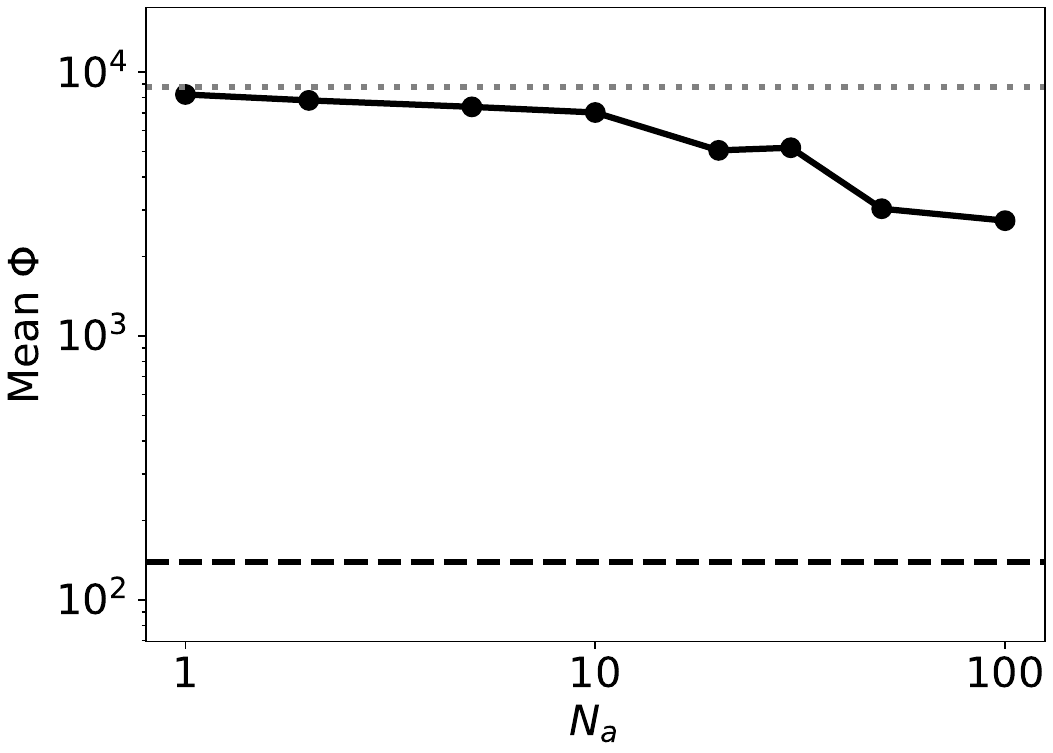}
\caption*{Figure S1: Mean data mismatch $\Phi$ of the ESMDA posterior ensemble versus the number of data assimilations $N_{\mathrm{a}}$ for Case~4, with the six-well monitoring configuration and both temperature and tracer concentration data (T+C). The solid curve shows the ESMDA results, while the dashed and dotted lines indicate the PT-MCMC and prior values, respectively.}
\label{fig:S1_esmda_na_sweep}
\end{figure}

\begin{figure}[!ht]
% \setfigurenum{S2}
% \renewcommand{\figurename}{Figure S2}
\centering
\makebox[0.98\textwidth][l]{%
\begin{minipage}[t]{0.3441\textwidth}
\centering
\includegraphics[width=\linewidth]{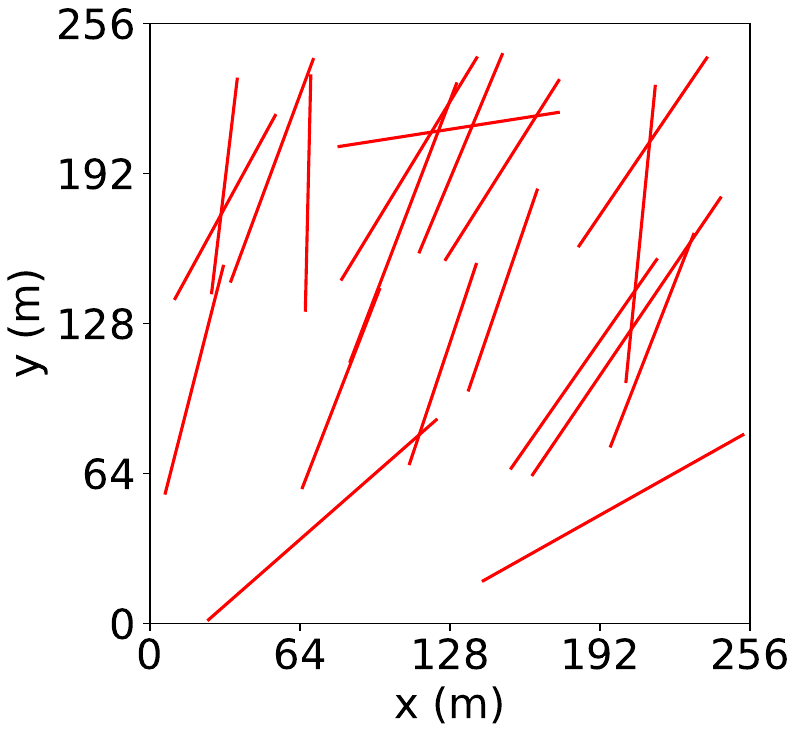}\\[2pt]
\hspace*{0.187\linewidth}\makebox[0.752\linewidth]{(a) True fracture}\hfill\mbox{}
\end{minipage}\hspace{0.0028\textwidth}%
\begin{minipage}[t]{0.2654\textwidth}
\centering
\includegraphics[width=\linewidth]{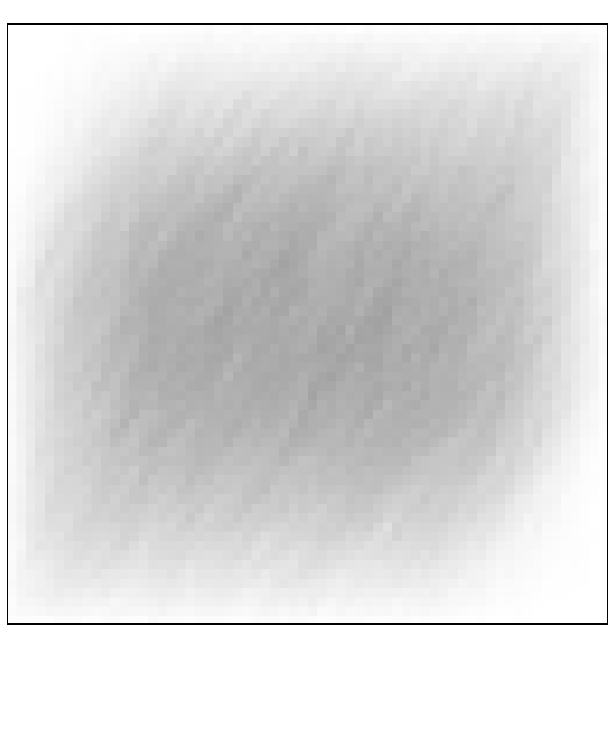}\\[2pt]
(b) Prior
\end{minipage}\hspace{0.02\textwidth}%
\begin{minipage}[t]{0.3478\textwidth}
\centering
\includegraphics[width=\linewidth]{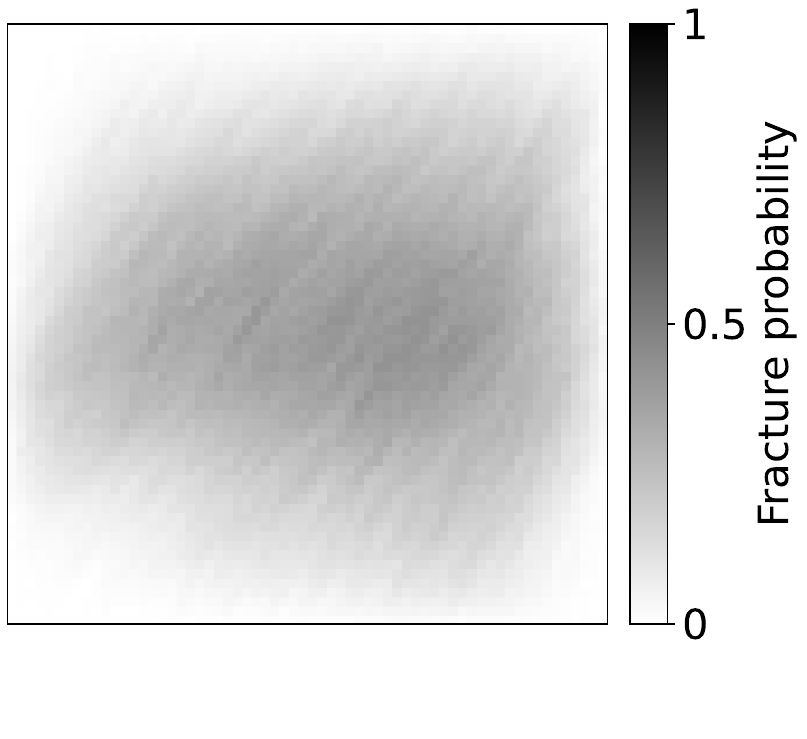}\\[2pt]
\makebox[0.763\linewidth]{(c) Posterior, $N_\mathrm{a} = 5$}\hfill\mbox{}
\end{minipage}}

\vspace{6pt}
\makebox[0.98\textwidth][l]{%
\begin{minipage}[t]{0.3441\textwidth}
\centering
\includegraphics[width=\linewidth]{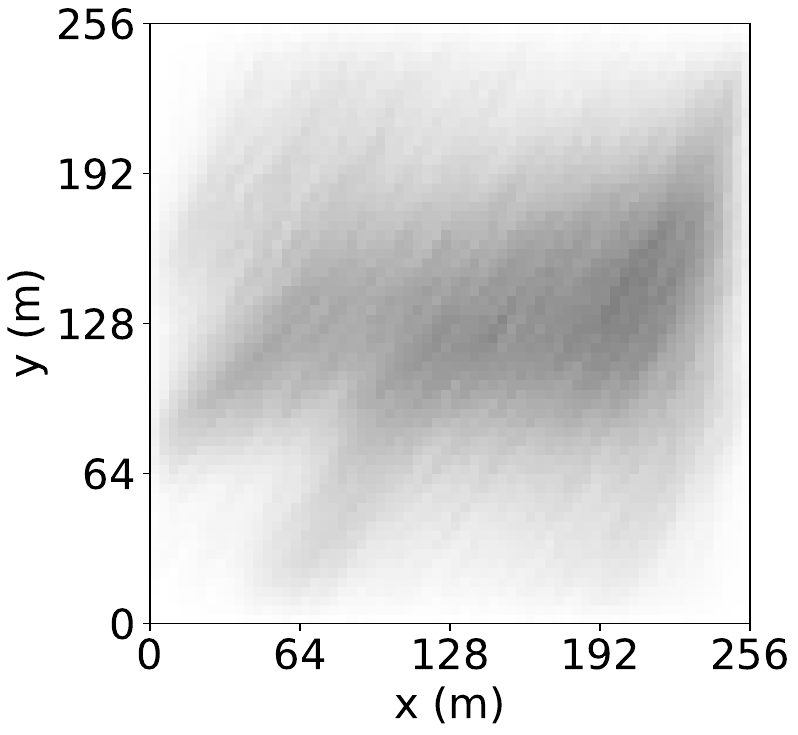}\\[2pt]
\hspace*{0.187\linewidth}\makebox[0.752\linewidth]{(d) Posterior, $N_\mathrm{a} = 10$}\hfill\mbox{}
\end{minipage}\hspace{0.0028\textwidth}%
\begin{minipage}[t]{0.2654\textwidth}
\centering
\includegraphics[width=\linewidth]{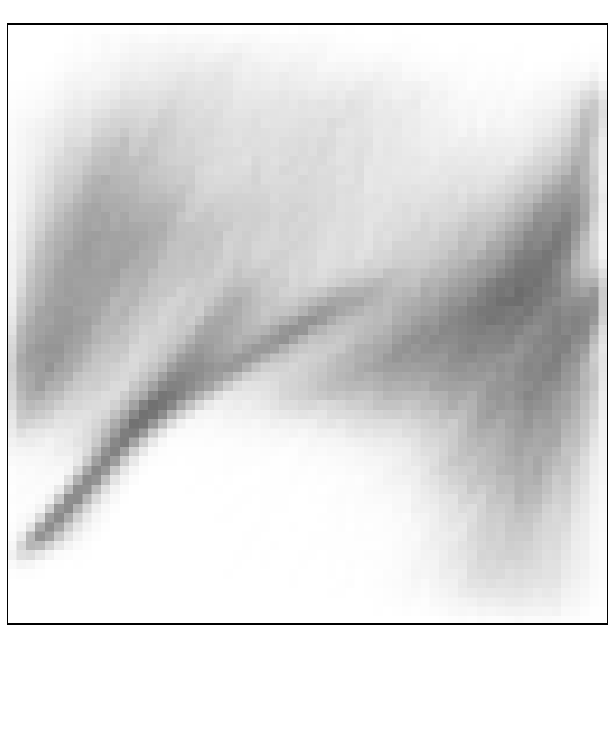}\\[2pt]
(e) Posterior, $N_\mathrm{a} = 50$
\end{minipage}\hspace{0.02\textwidth}%
\begin{minipage}[t]{0.3478\textwidth}
\centering
\includegraphics[width=\linewidth]{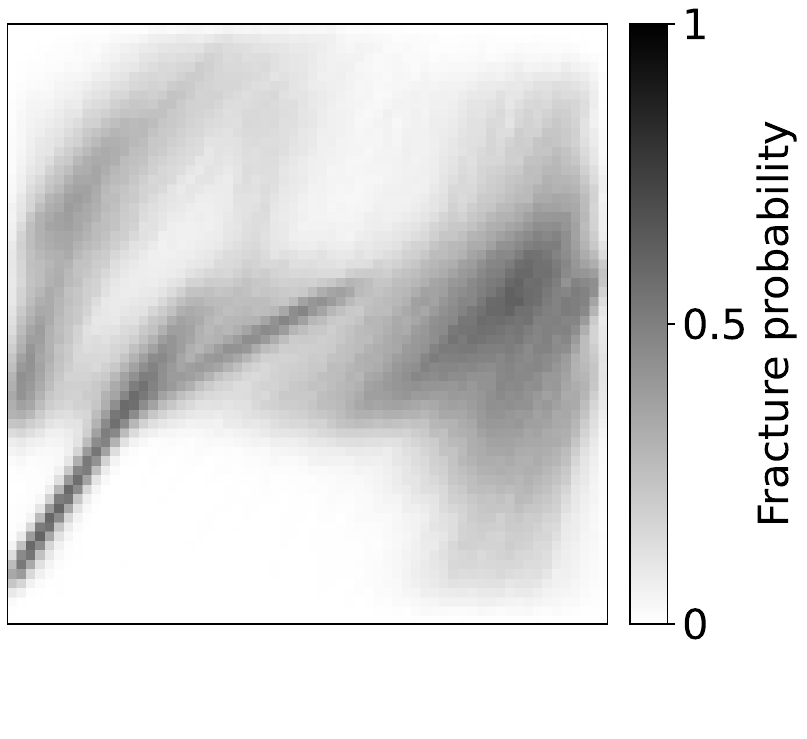}\\[2pt]
\makebox[0.763\linewidth]{(f) Posterior, $N_\mathrm{a} = 100$}\hfill\mbox{}
\end{minipage}}%
\caption*{Figure S2: Fracture probability maps for Case~4, with the six-well monitoring configuration and T+C data. (a) True fracture network; (b) prior fracture probability; (c-f) ESMDA posterior fracture probability for $N_{\mathrm{a}}=5$, 10, 50, and 100, respectively.}
\label{fig:S2_esmda_na_occupancy}
\end{figure}

Figure~S3 compares the PT-MCMC and ESMDA results for Case~4. PT-MCMC produces more localized fracture probability distributions and substantially narrower posterior uncertainty ranges for temperature predictions. ESMDA captures the main features of the true fracture network and temperature field but retains greater uncertainty in the predicted responses. Similar results are obtained for Case~5 (Figure~S4), indicating that these differences are not limited to a particular DFN realization.

\begin{figure}[!ht]
% \setfigurenum{S3}
\centering
\makebox[\textwidth][l]{%
\hspace*{0.0060\textwidth}%
\begin{minipage}[t]{0.2991\textwidth}
\centering
\includegraphics[width=\linewidth]{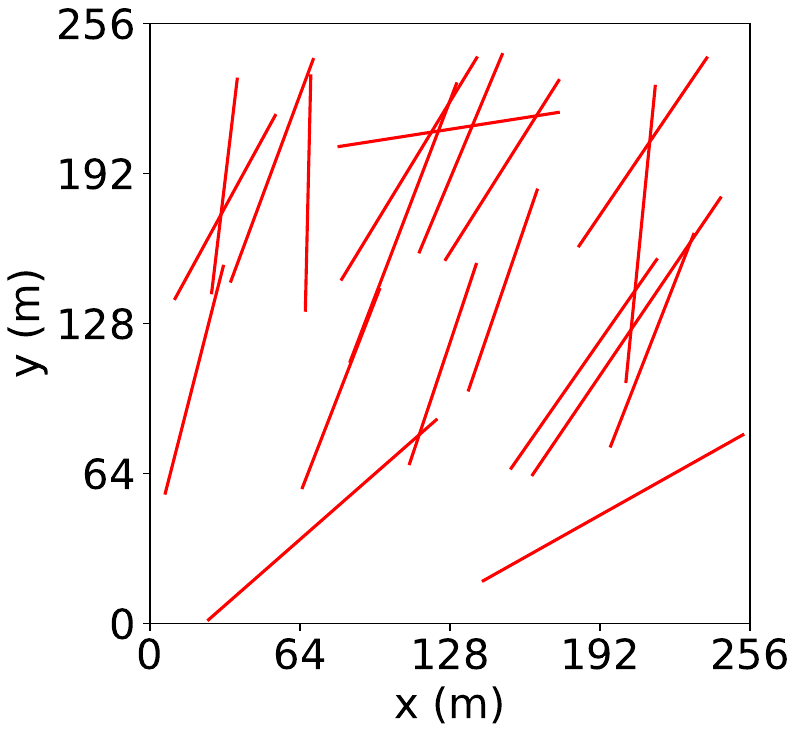}\\[2pt]
\hspace*{0.1881\linewidth}\makebox[0.7524\linewidth]{(a) True fracture}\hfill\mbox{}
\end{minipage}%
\hspace{0.0694\textwidth}%
\begin{minipage}[t]{0.2306\textwidth}
\centering
\includegraphics[width=\linewidth]{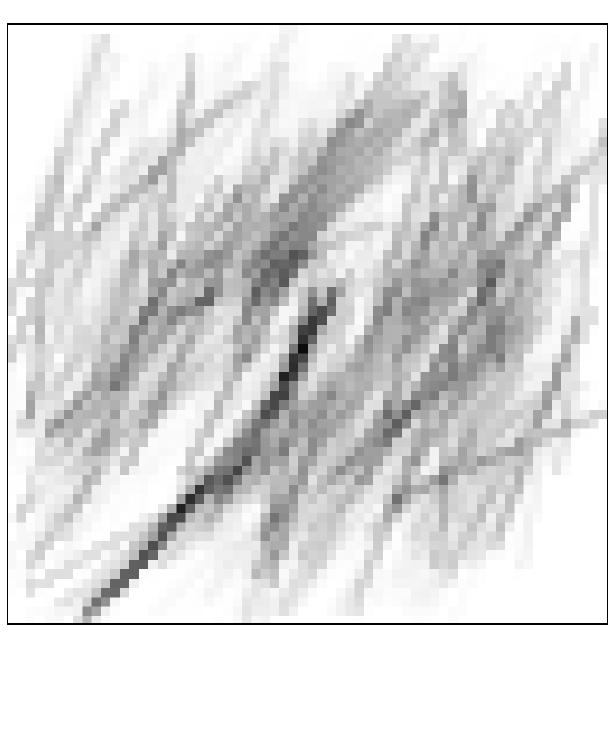}\\[2pt]
\hspace*{0.0122\linewidth}\makebox[0.9756\linewidth]{(b) PT-MCMC}\hfill\mbox{}
\end{minipage}%
\hspace{0.0844\textwidth}%
\begin{minipage}[t]{0.3023\textwidth}
\centering
\includegraphics[width=\linewidth]{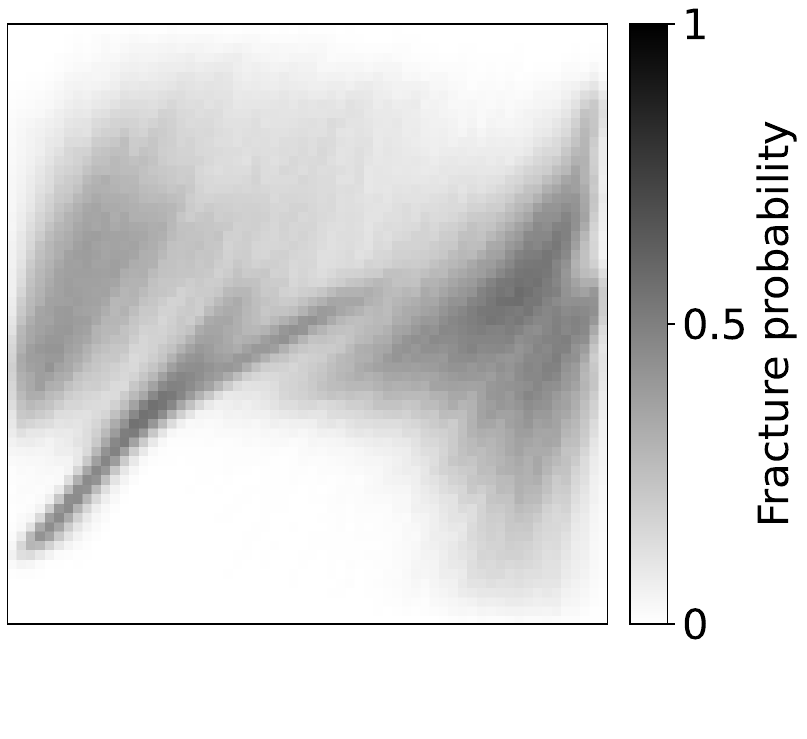}\\[2pt]
\hspace*{0.0093\linewidth}\makebox[0.7443\linewidth]{(c) ESMDA}\hfill\mbox{}
\end{minipage}}

\vspace{6pt}
\makebox[\textwidth][l]{%
\begin{minipage}[t]{0.3125\textwidth}
\centering
\includegraphics[width=\linewidth]{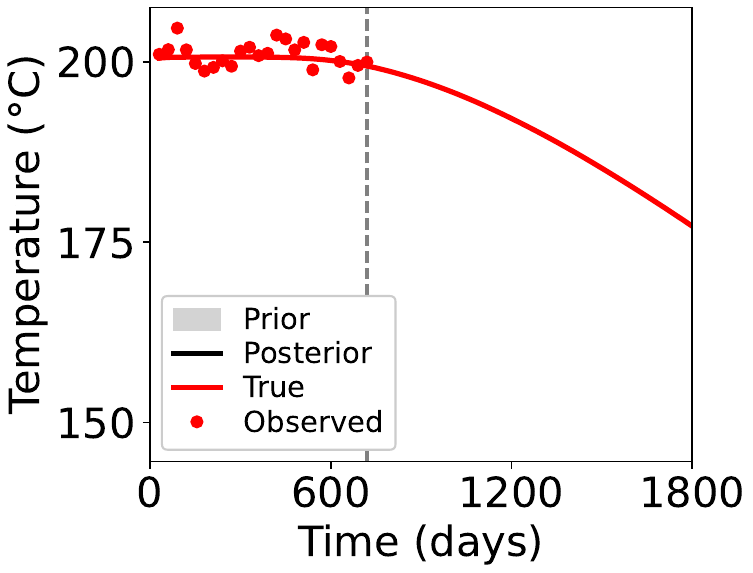}\\[2pt]
\hspace*{0.1994\linewidth}\makebox[0.7200\linewidth]{(d) True M1 temperature}\hfill\mbox{}
\end{minipage}%
\hspace{0.0025\textwidth}%
\begin{minipage}[t]{0.3125\textwidth}
\centering
\includegraphics[width=\linewidth]{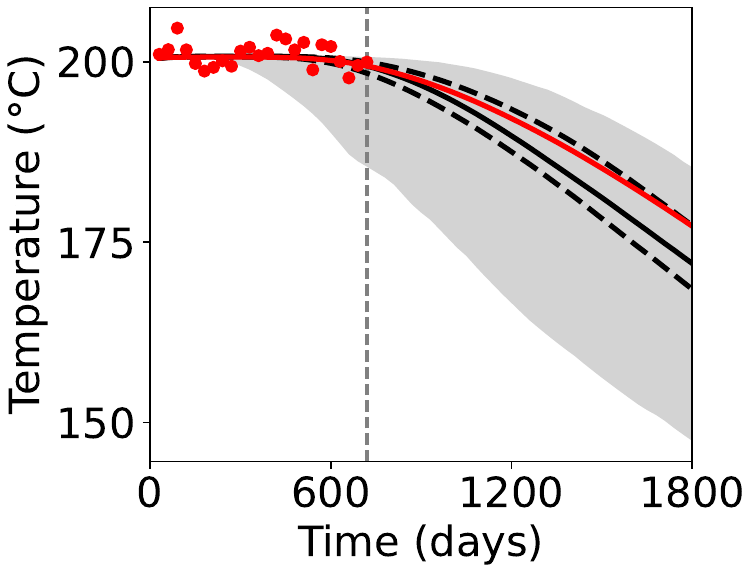}\\[2pt]
\hspace*{0.1994\linewidth}\makebox[0.7200\linewidth]{(e) PT-MCMC}\hfill\mbox{}
\end{minipage}%
\hspace{0.0025\textwidth}%
\begin{minipage}[t]{0.3125\textwidth}
\centering
\includegraphics[width=\linewidth]{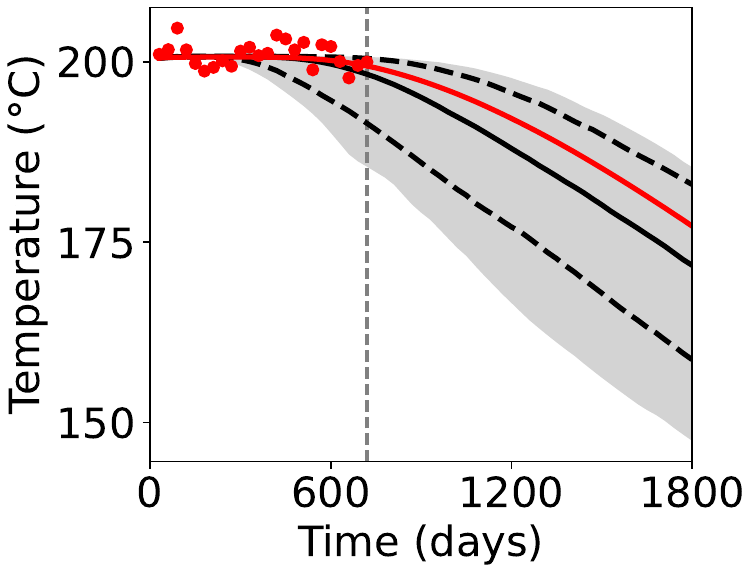}\\[2pt]
\hspace*{0.1994\linewidth}\makebox[0.7200\linewidth]{(f) ESMDA}\hfill\mbox{}
\end{minipage}}

\vspace{6pt}
\makebox[\textwidth][l]{%
\hspace*{0.0060\textwidth}%
\begin{minipage}[t]{0.2991\textwidth}
\centering
\includegraphics[width=\linewidth]{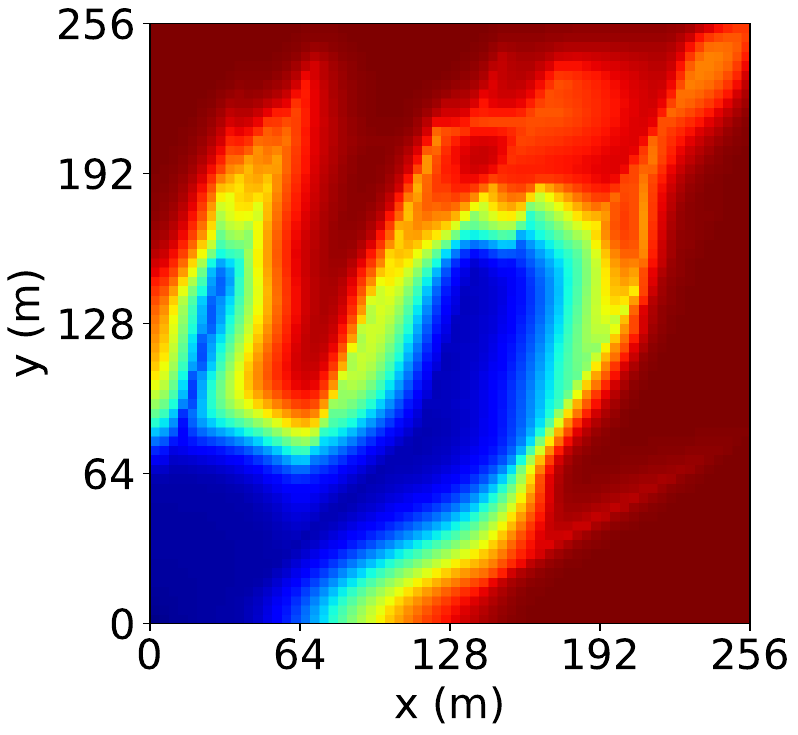}\\[2pt]
\hspace*{0.1881\linewidth}\makebox[0.7524\linewidth]{(g) True temperature}\hfill\mbox{}
\end{minipage}%
\hspace{0.0694\textwidth}%
\begin{minipage}[t]{0.2306\textwidth}
\centering
\includegraphics[width=\linewidth]{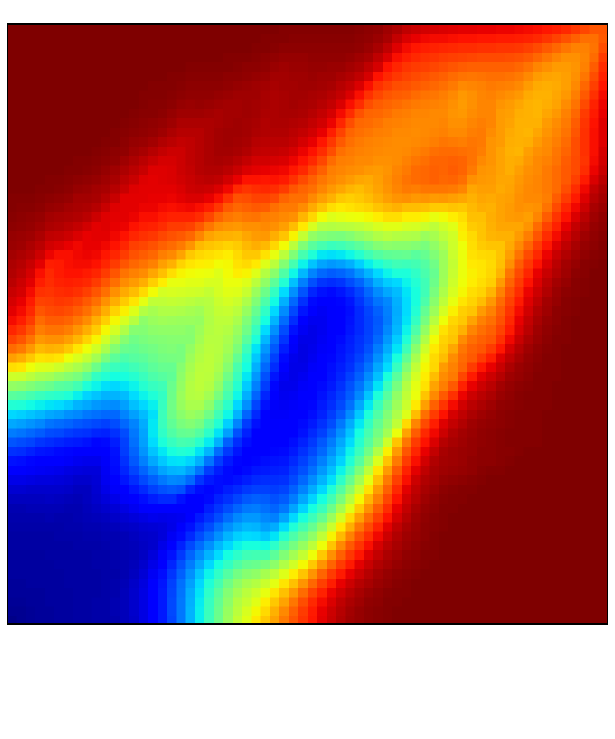}\\[2pt]
\hspace*{0.0122\linewidth}\makebox[0.9756\linewidth]{(h) PT-MCMC}\hfill\mbox{}
\end{minipage}%
\hspace{0.0844\textwidth}%
\begin{minipage}[t]{0.3072\textwidth}
\centering
\includegraphics[width=\linewidth]{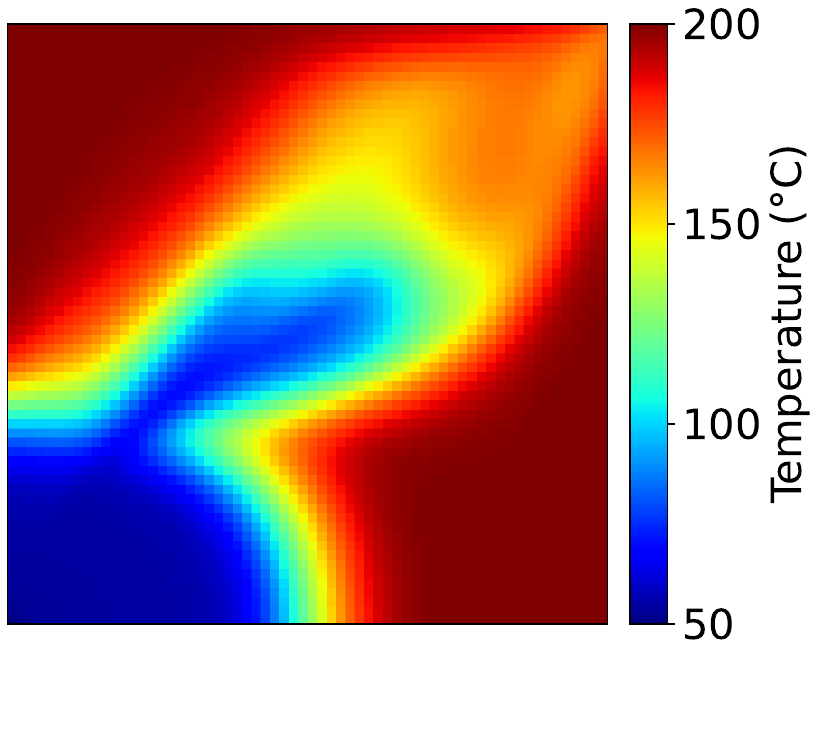}\\[2pt]
\hspace*{0.0092\linewidth}\makebox[0.7324\linewidth]{(i) ESMDA}\hfill\mbox{}
\end{minipage}}
\caption*{Figure S3: Comparison of PT-MCMC and ESMDA ($N_{\mathrm{a}}=50$) results for Case~4, with the six-well monitoring configuration and T+C data. The first row shows the true fracture network and posterior fracture probability maps, the second row shows temperature predictions at monitoring well M1, and the third row shows the true and posterior mean temperature fields at 1800~days. In the second row, gray shading indicates the prior P10-P90 range, black curves show the posterior P10, P50, and P90, and red curves and points indicate the true values and observations, respectively. The vertical gray dashed line marks the end of the history matching period.}
\label{fig:S3_esmda_case4}
\end{figure}

\begin{figure}[!ht]
% \setfigurenum{S4}
\centering
\makebox[\textwidth][l]{%
\hspace*{0.0060\textwidth}%
\begin{minipage}[t]{0.2991\textwidth}
\centering
\includegraphics[width=\linewidth]{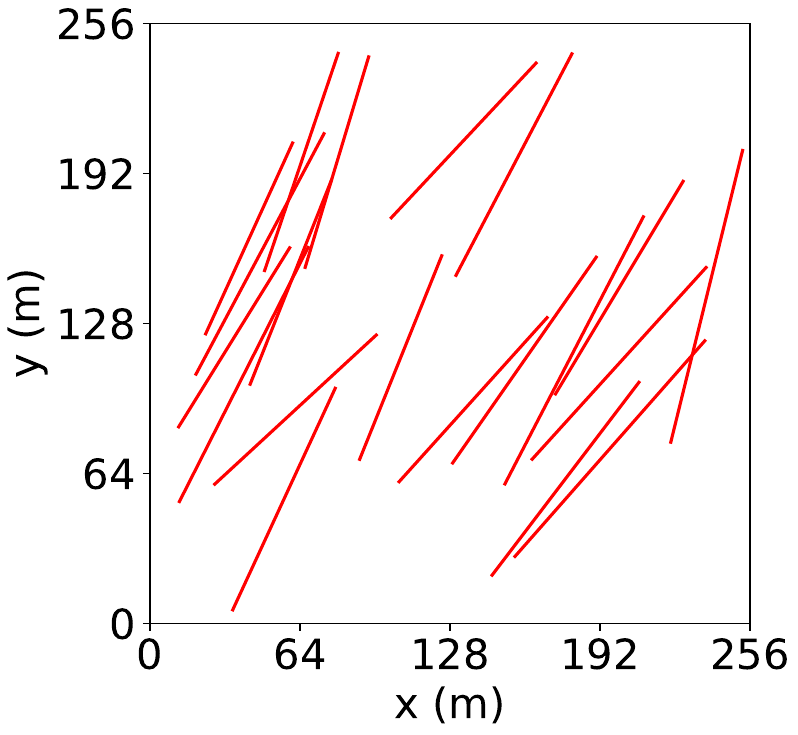}\\[2pt]
\hspace*{0.1881\linewidth}\makebox[0.7524\linewidth]{(a) True fracture}\hfill\mbox{}
\end{minipage}%
\hspace{0.0694\textwidth}%
\begin{minipage}[t]{0.2306\textwidth}
\centering
\includegraphics[width=\linewidth]{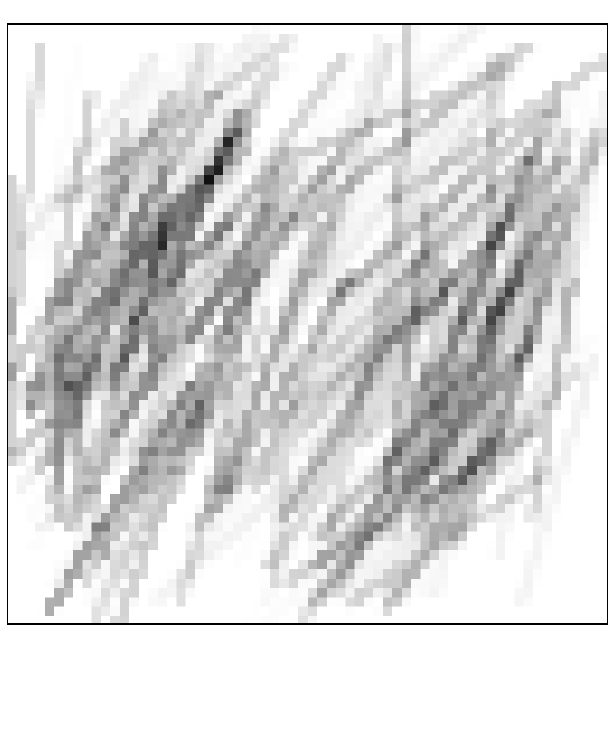}\\[2pt]
\hspace*{0.0122\linewidth}\makebox[0.9756\linewidth]{(b) PT-MCMC}\hfill\mbox{}
\end{minipage}%
\hspace{0.0844\textwidth}%
\begin{minipage}[t]{0.3023\textwidth}
\centering
\includegraphics[width=\linewidth]{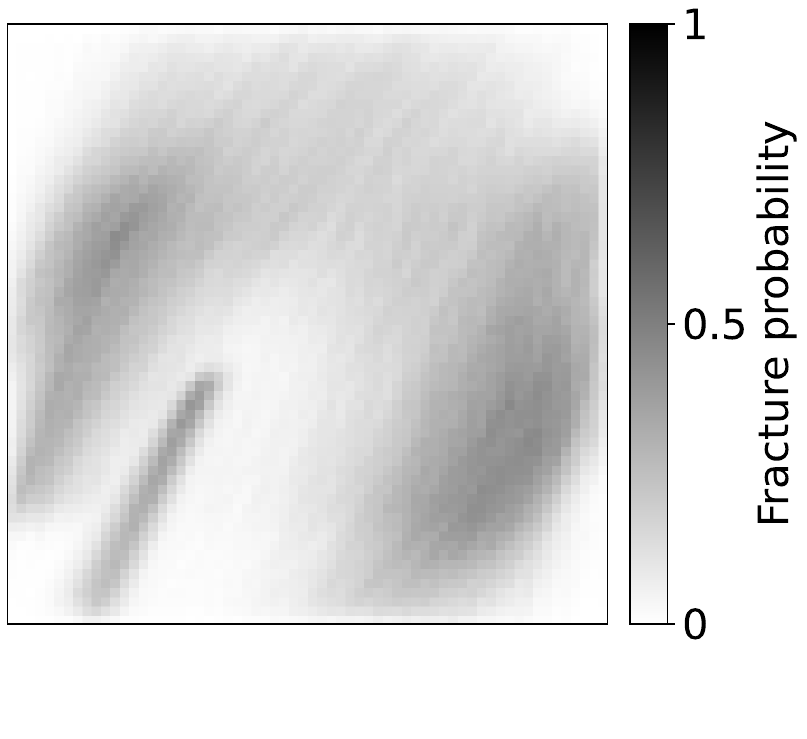}\\[2pt]
\hspace*{0.0093\linewidth}\makebox[0.7443\linewidth]{(c) ESMDA}\hfill\mbox{}
\end{minipage}}

\vspace{6pt}
\makebox[\textwidth][l]{%
\begin{minipage}[t]{0.3125\textwidth}
\centering
\includegraphics[width=\linewidth]{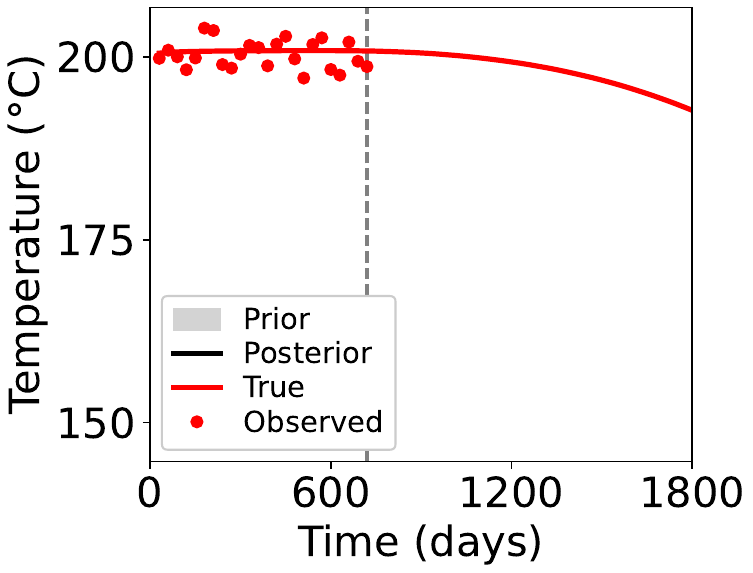}\\[2pt]
\hspace*{0.1994\linewidth}\makebox[0.7200\linewidth]{(d) True M1 temperature}\hfill\mbox{}
\end{minipage}%
\hspace{0.0025\textwidth}%
\begin{minipage}[t]{0.3125\textwidth}
\centering
\includegraphics[width=\linewidth]{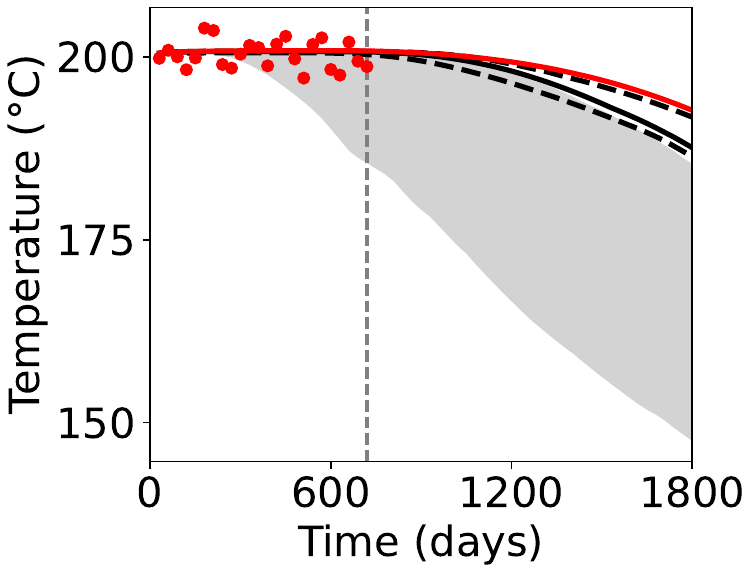}\\[2pt]
\hspace*{0.1994\linewidth}\makebox[0.7200\linewidth]{(e) PT-MCMC}\hfill\mbox{}
\end{minipage}%
\hspace{0.0025\textwidth}%
\begin{minipage}[t]{0.3125\textwidth}
\centering
\includegraphics[width=\linewidth]{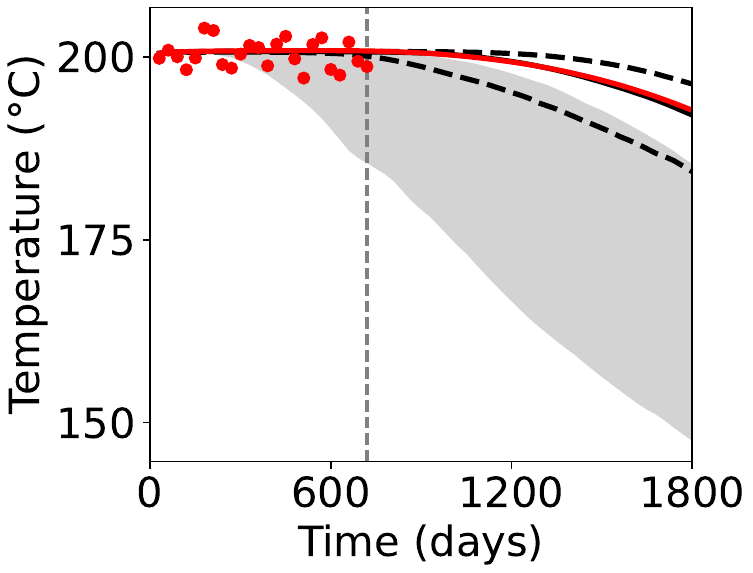}\\[2pt]
\hspace*{0.1994\linewidth}\makebox[0.7200\linewidth]{(f) ESMDA}\hfill\mbox{}
\end{minipage}}

\vspace{6pt}
\makebox[\textwidth][l]{%
\hspace*{0.0060\textwidth}%
\begin{minipage}[t]{0.2991\textwidth}
\centering
\includegraphics[width=\linewidth]{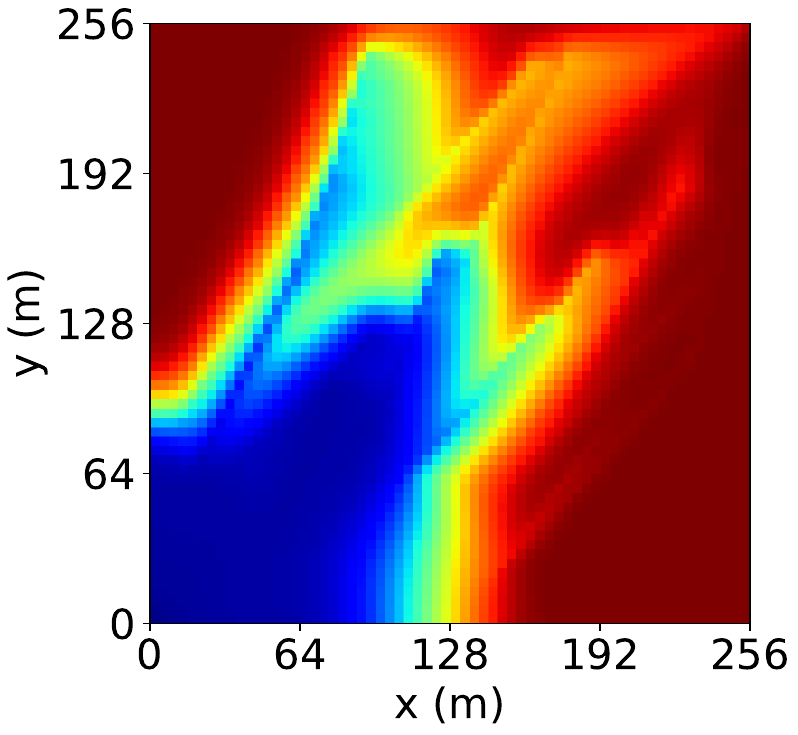}\\[2pt]
\hspace*{0.1881\linewidth}\makebox[0.7524\linewidth]{(g) True temperature}\hfill\mbox{}
\end{minipage}%
\hspace{0.0694\textwidth}%
\begin{minipage}[t]{0.2306\textwidth}
\centering
\includegraphics[width=\linewidth]{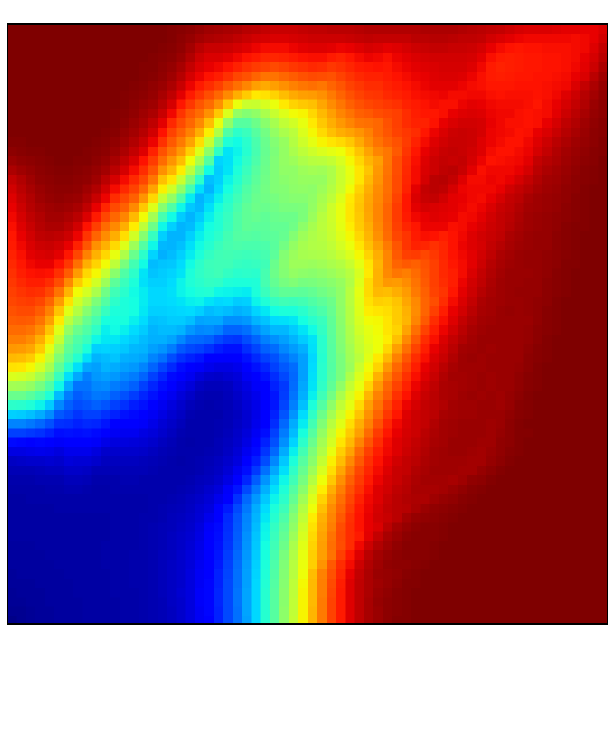}\\[2pt]
\hspace*{0.0122\linewidth}\makebox[0.9756\linewidth]{(h) PT-MCMC}\hfill\mbox{}
\end{minipage}%
\hspace{0.0844\textwidth}%
\begin{minipage}[t]{0.3072\textwidth}
\centering
\includegraphics[width=\linewidth]{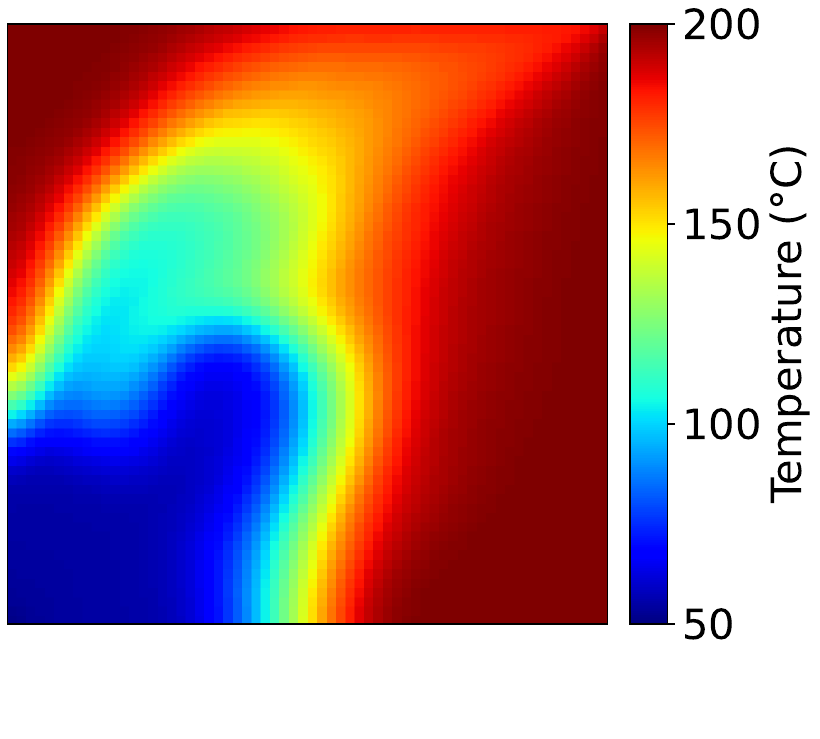}\\[2pt]
\hspace*{0.0092\linewidth}\makebox[0.7324\linewidth]{(i) ESMDA}\hfill\mbox{}
\end{minipage}}
\caption*{Figure S4: Same as Figure~S3, but for Case~5.}
\label{fig:S4_esmda_case5}
\end{figure}

To better understand these differences, we examine the prior and posterior distributions of selected latent variables for Case~4 (Figure~S5). The prior distribution of $\xi_{60}$ is standard Gaussian, whereas the PT-MCMC posterior exhibits multiple modes. ESMDA fails to reproduce this multimodal structure and yields a shifted distribution. The joint distribution of $\xi_{40}$ and $\xi_{60}$ shows similar differences. PT-MCMC identifies multiple regions of high posterior probability, whereas ESMDA concentrates the ensemble in a more restricted region. These results suggest that the ensemble-based updates in ESMDA have difficulty representing the non-Gaussian, multimodal posterior distributions encountered in fracture network inversion.

\begin{figure}[!ht]
% \setfigurenum{S5}
\centering

\begin{minipage}[t]{0.33\linewidth}
\centering
\includegraphics[trim={70pt, 10pt, 45pt, 10pt},width=\linewidth]{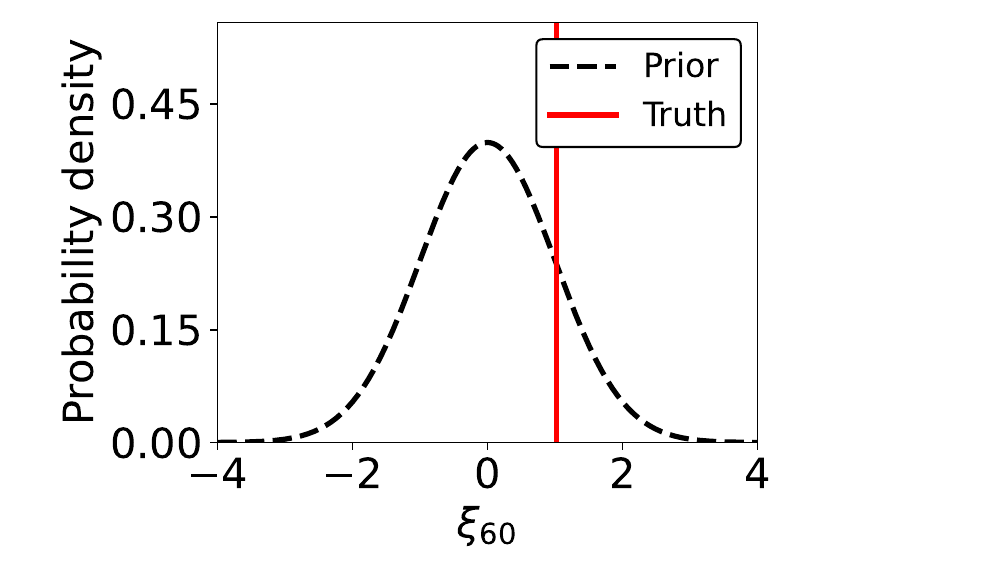}\\[2pt]
(a) Prior, $\xi_{60}$
\end{minipage}\hfill
\begin{minipage}[t]{0.33\linewidth}
\centering
\includegraphics[trim={70pt, 10pt, 45pt, 10pt},width=\linewidth]{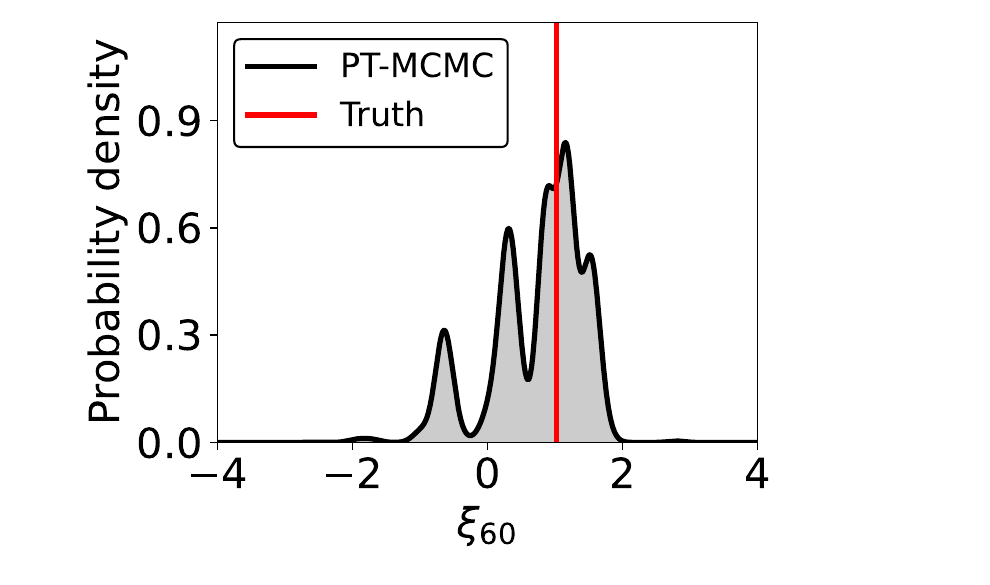}\\[2pt]
(b) PT-MCMC, $\xi_{60}$
\end{minipage}\hfill
\begin{minipage}[t]{0.33\linewidth}
\centering
\includegraphics[trim={70pt, 10pt, 45pt, 10pt}, width=\linewidth]{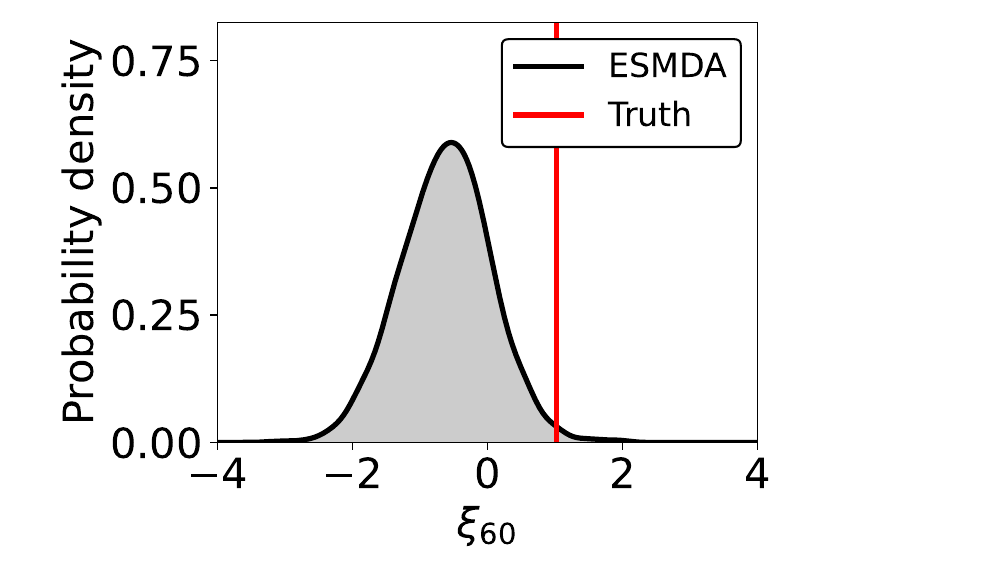}\\[2pt]
(c) ESMDA, $\xi_{60}$
\end{minipage}

\vspace{6pt}

\begin{minipage}[t]{0.33\linewidth}
\centering
\includegraphics[trim={70pt, 10pt, 45pt, 10pt},width=\linewidth]{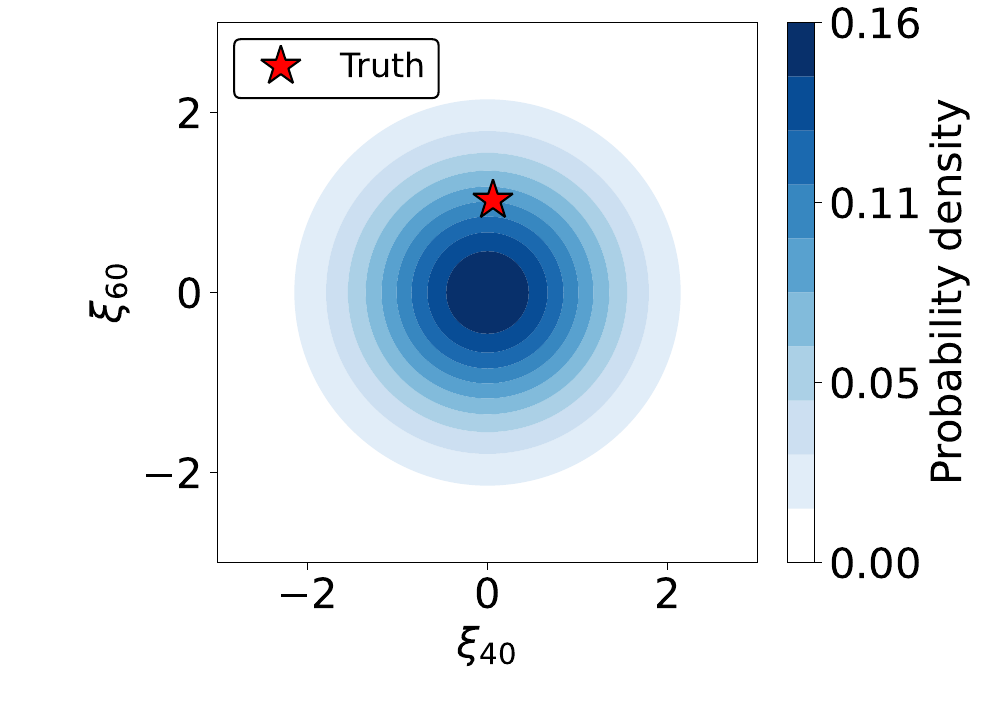}\\[2pt]
(d) Prior, $\xi_{40}$ and $\xi_{60}$
\end{minipage}\hfill
\begin{minipage}[t]{0.33\linewidth}
\centering
\includegraphics[trim={70pt, 10pt, 45pt, 10pt}, width=\linewidth]{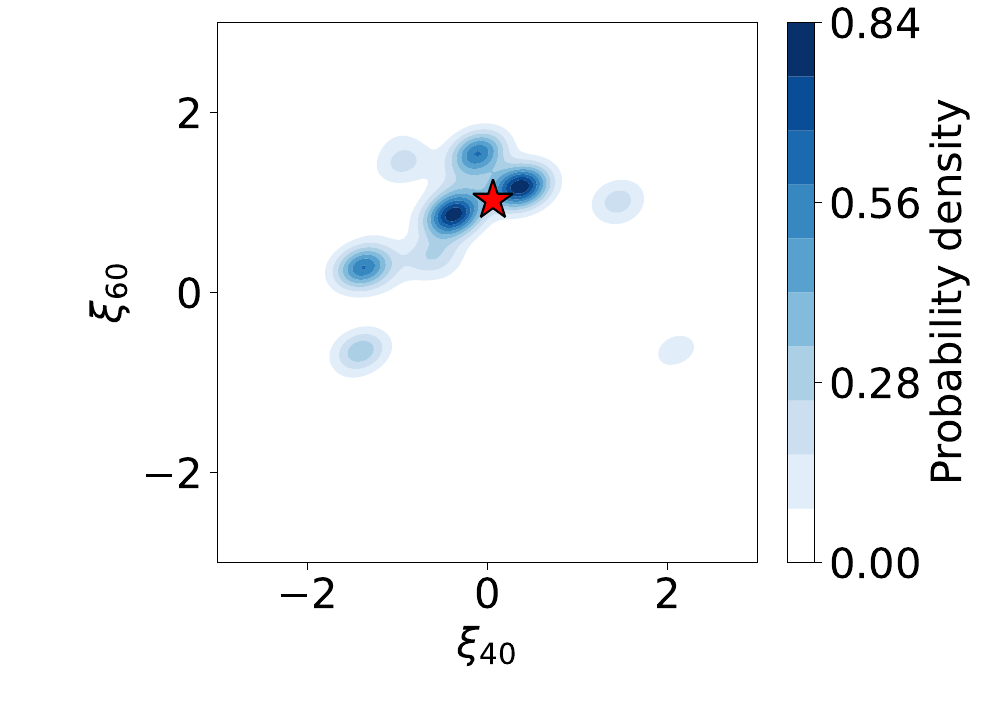}\\[2pt]
(e) PT-MCMC, $\xi_{40}$ and $\xi_{60}$
\end{minipage}\hfill
\begin{minipage}[t]{0.33\linewidth}
\centering
\includegraphics[trim={70pt, 10pt, 45pt, 10pt}, width=\linewidth]{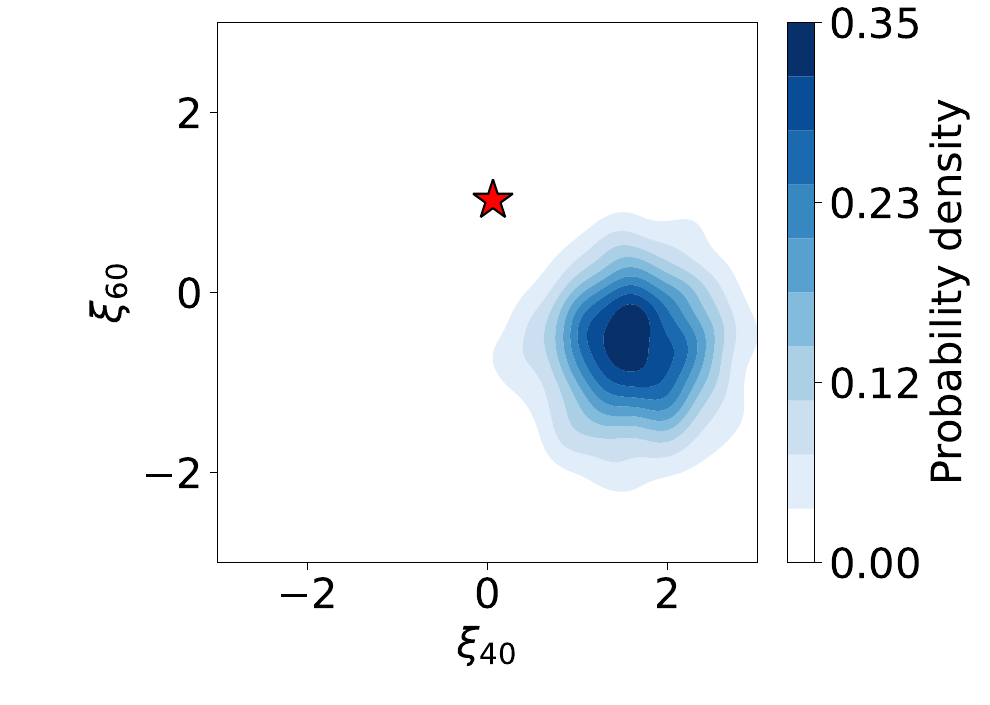}\\[2pt]
(f) ESMDA, $\xi_{40}$ and $\xi_{60}$
\end{minipage}

\caption*{Figure S5: Prior and posterior distributions of selected latent variables for Case~4, with the six-well monitoring configuration and both temperature and tracer concentration data (T+C). The top row shows the marginal distribution of $\xi_{60}$, and the bottom row shows the joint distribution of $\xi_{40}$ and $\xi_{60}$. The red line and star indicate the true values.}
\label{fig:S5_latent}
\end{figure}

We further examine representative prior and PT-MCMC posterior realizations for Case~4 (Figure~S6). We randomly select 1,000 realizations from each distribution and apply k-means clustering separately to their predicted temperature fields at 1800~days. Five clusters are identified for each distribution, and the realization closest to each cluster centroid is selected as representative. The prior realizations show considerable variability in fracture geometry and temperature distribution. In contrast, the posterior realizations consistently reproduce the main fracture features in the lower-left and right portions of the domain, although the locations of individual fractures vary. Their temperature fields show similar cooling patterns and reasonable agreement with the true temperature field. These results indicate that different fracture configurations can produce similar reservoir responses, consistent with the multimodal posterior distribution obtained using PT-MCMC.

\begin{figure}[!ht]
% \setfigurenum{S6}
\centering

\makebox[\textwidth][l]{%
\hspace*{0.0000\textwidth}%
\includegraphics[width=0.2177\textwidth]{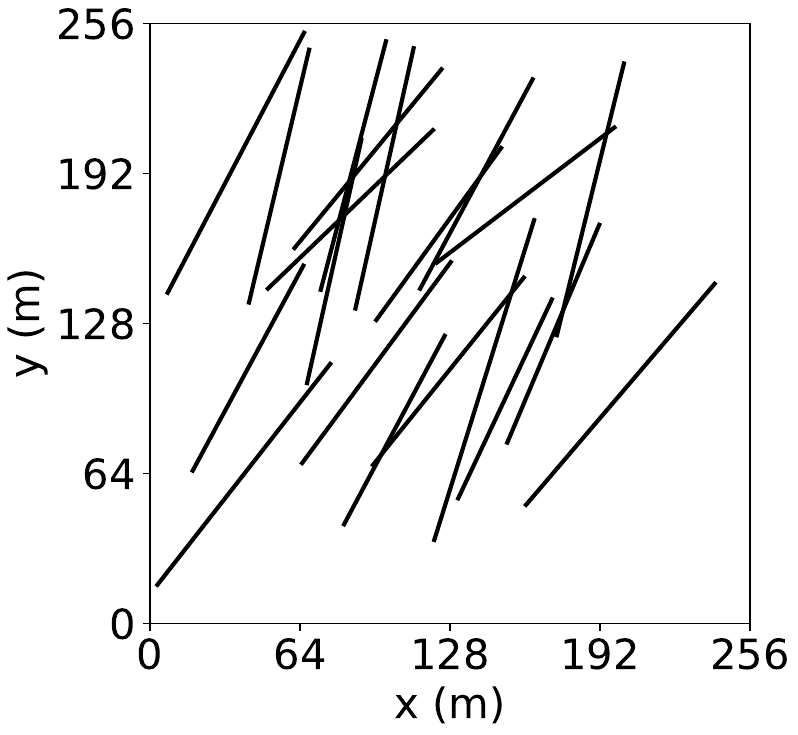}%
\hspace{0.0056\textwidth}%
\includegraphics[width=0.1679\textwidth]{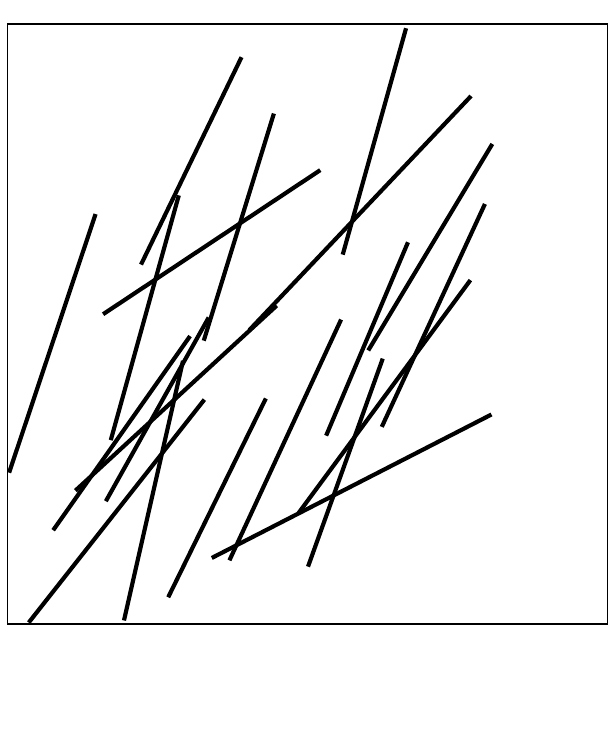}%
\hspace{0.0165\textwidth}%
\includegraphics[width=0.1679\textwidth]{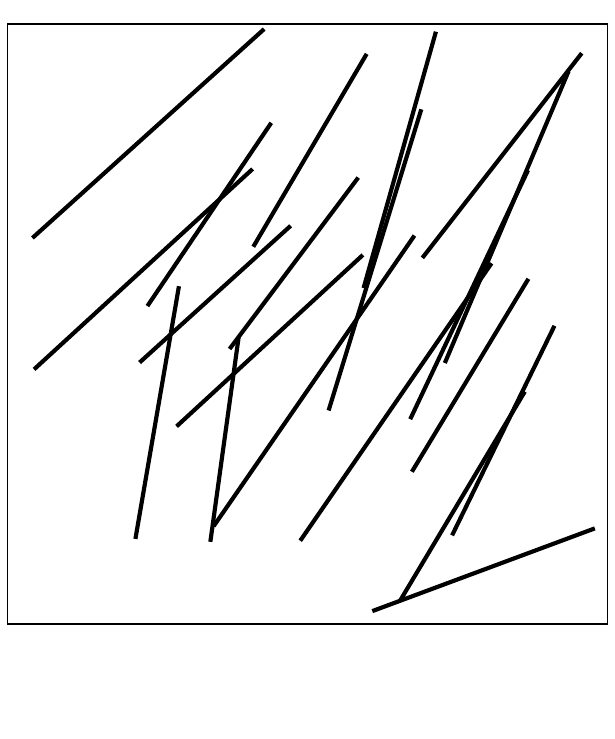}%
\hspace{0.0165\textwidth}%
\includegraphics[width=0.1679\textwidth]{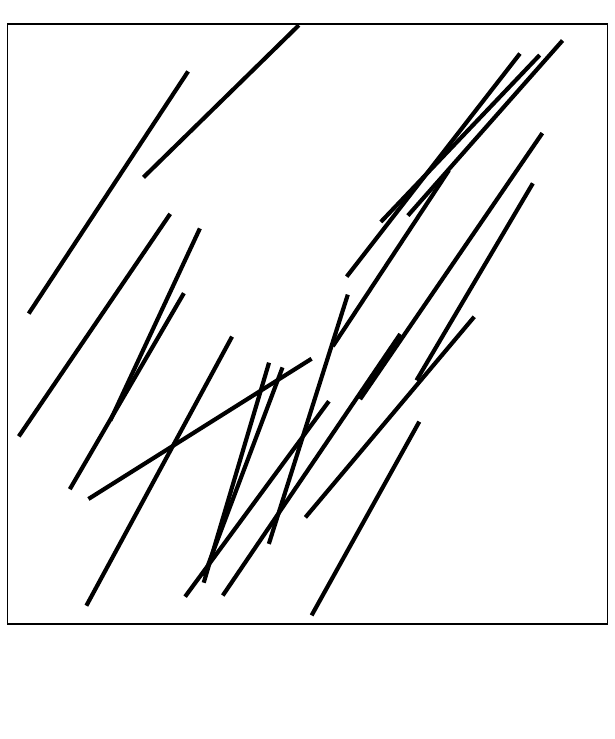}%
\hspace{0.0165\textwidth}%
\includegraphics[width=0.1679\textwidth]{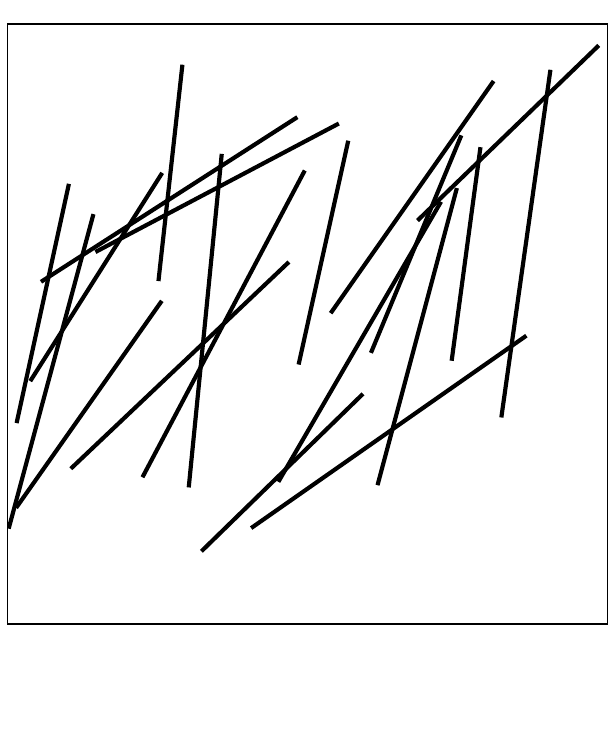}%
}

\makebox[\textwidth][l]{%
\hspace*{0.0410\textwidth}%
\makebox[0.1638\textwidth]{\footnotesize (a) Prior 1}%
\hspace{0.0206\textwidth}%
\makebox[0.1638\textwidth]{\footnotesize (b) Prior 2}%
\hspace{0.0206\textwidth}%
\makebox[0.1638\textwidth]{\footnotesize (c) Prior 3}%
\hspace{0.0206\textwidth}%
\makebox[0.1638\textwidth]{\footnotesize (d) Prior 4}%
\hspace{0.0206\textwidth}%
\makebox[0.1638\textwidth]{\footnotesize (e) Prior 5}%
}

\makebox[\textwidth][l]{%
\hspace*{0.0000\textwidth}%
\includegraphics[width=0.2177\textwidth]{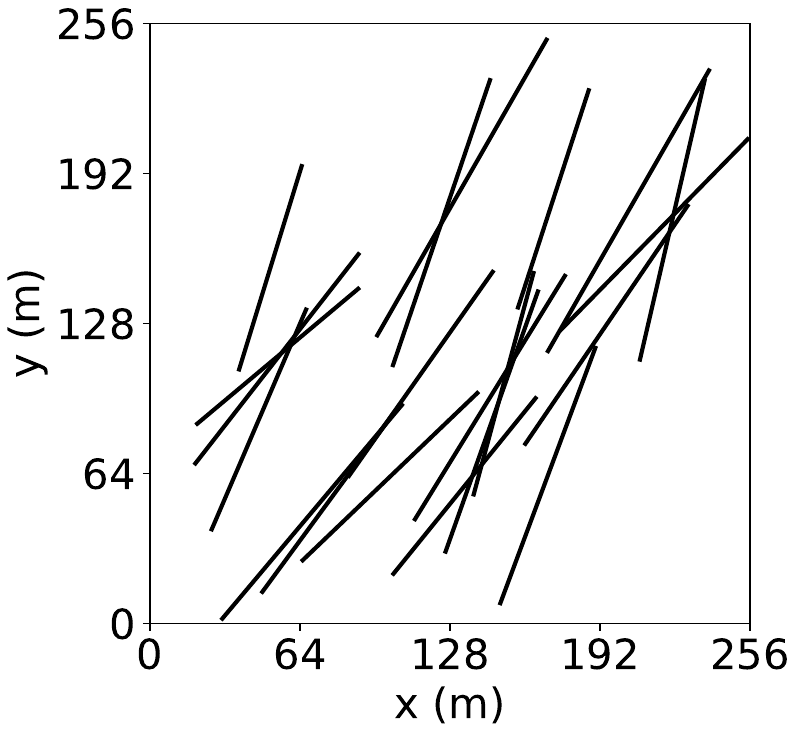}%
\hspace{0.0056\textwidth}%
\includegraphics[width=0.1679\textwidth]{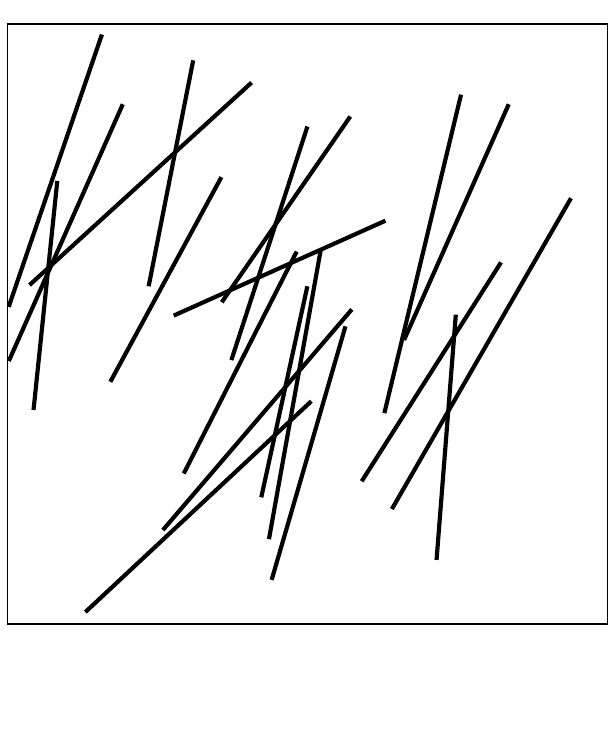}%
\hspace{0.0165\textwidth}%
\includegraphics[width=0.1679\textwidth]{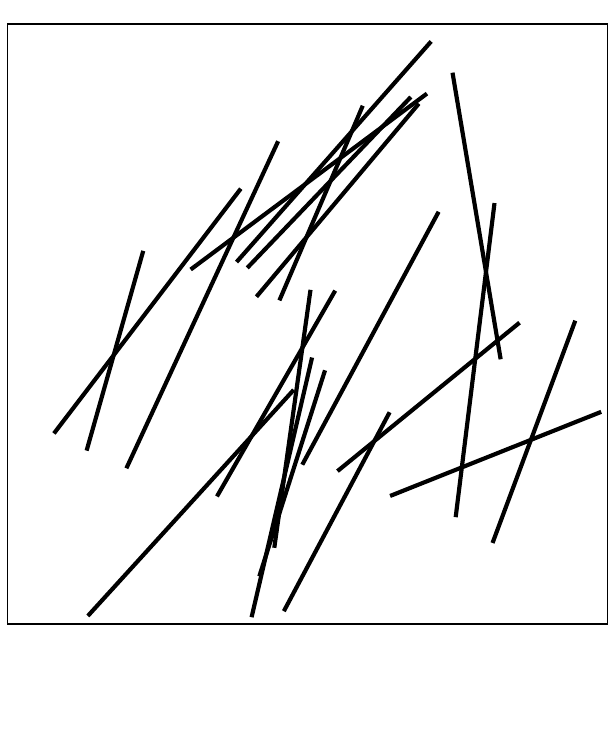}%
\hspace{0.0165\textwidth}%
\includegraphics[width=0.1679\textwidth]{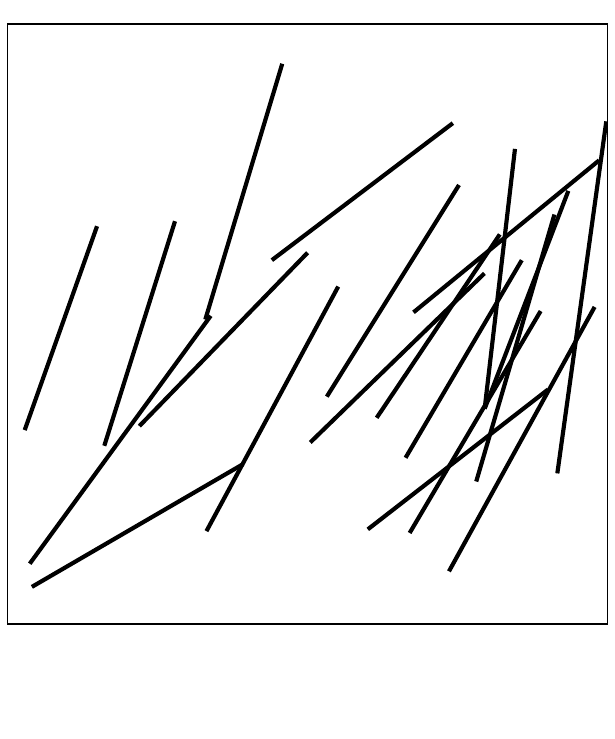}%
\hspace{0.0165\textwidth}%
\includegraphics[width=0.1679\textwidth]{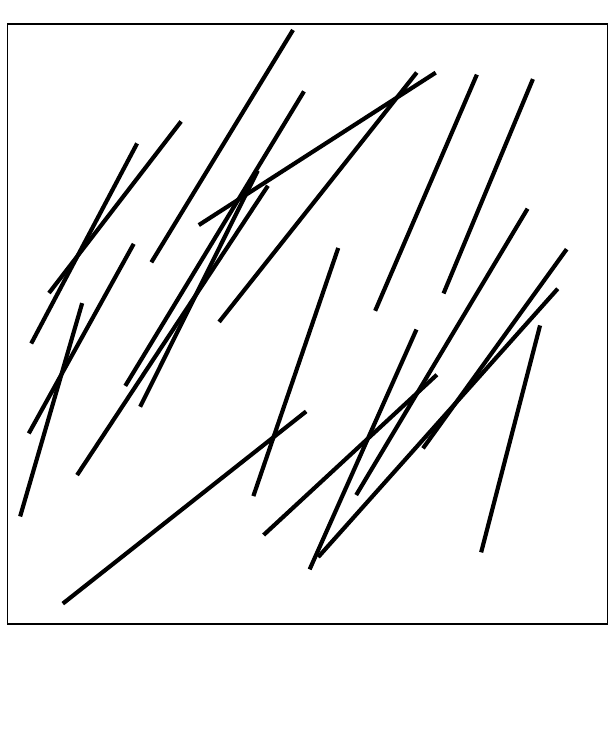}%
}

\makebox[\textwidth][l]{%
\hspace*{0.0410\textwidth}%
\makebox[0.1638\textwidth]{\footnotesize (f) Posterior 1}%
\hspace{0.0206\textwidth}%
\makebox[0.1638\textwidth]{\footnotesize (g) Posterior 2}%
\hspace{0.0206\textwidth}%
\makebox[0.1638\textwidth]{\footnotesize (h) Posterior 3}%
\hspace{0.0206\textwidth}%
\makebox[0.1638\textwidth]{\footnotesize (i) Posterior 4}%
\hspace{0.0206\textwidth}%
\makebox[0.1638\textwidth]{\footnotesize (j) Posterior 5}%
}

\makebox[\textwidth][l]{%
\hspace*{0.0000\textwidth}%
\includegraphics[width=0.2177\textwidth]{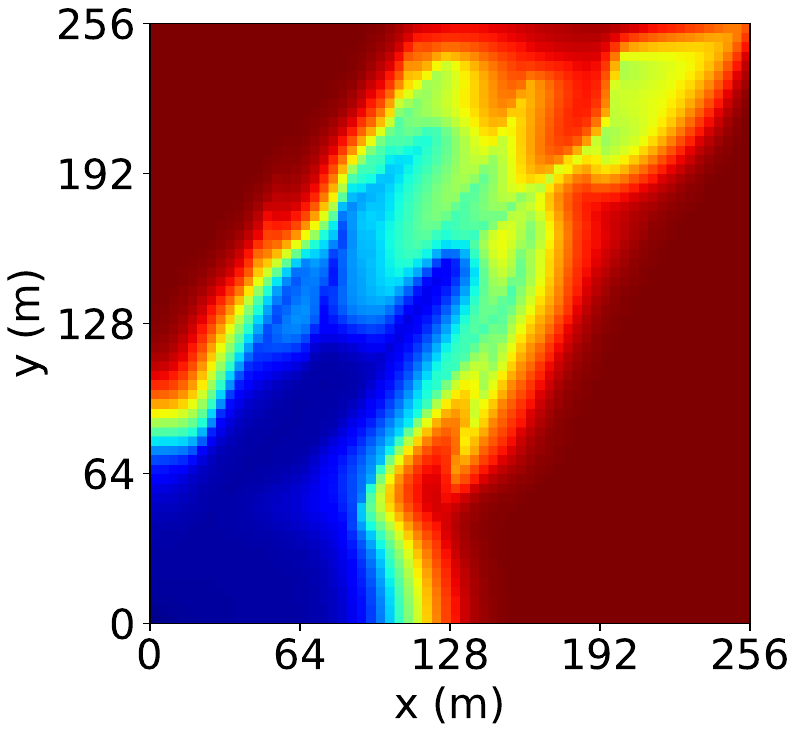}%
\hspace{0.0056\textwidth}%
\includegraphics[width=0.1679\textwidth]{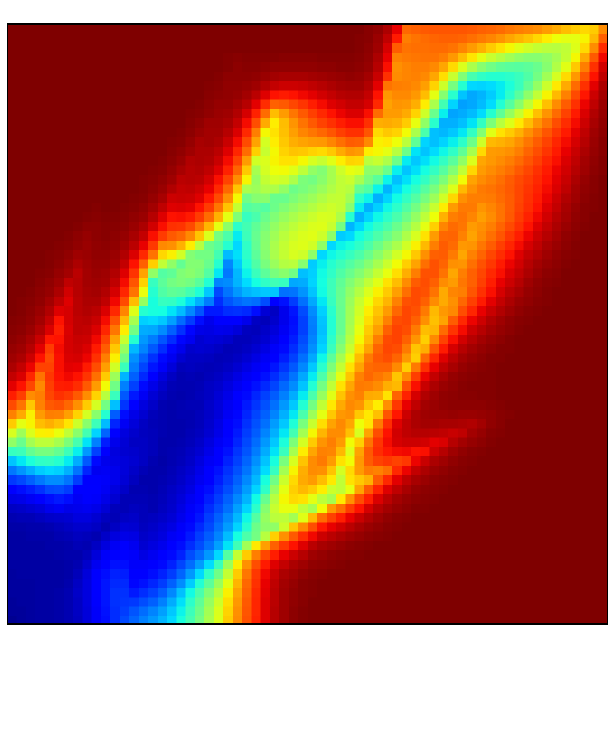}%
\hspace{0.0165\textwidth}%
\includegraphics[width=0.1679\textwidth]{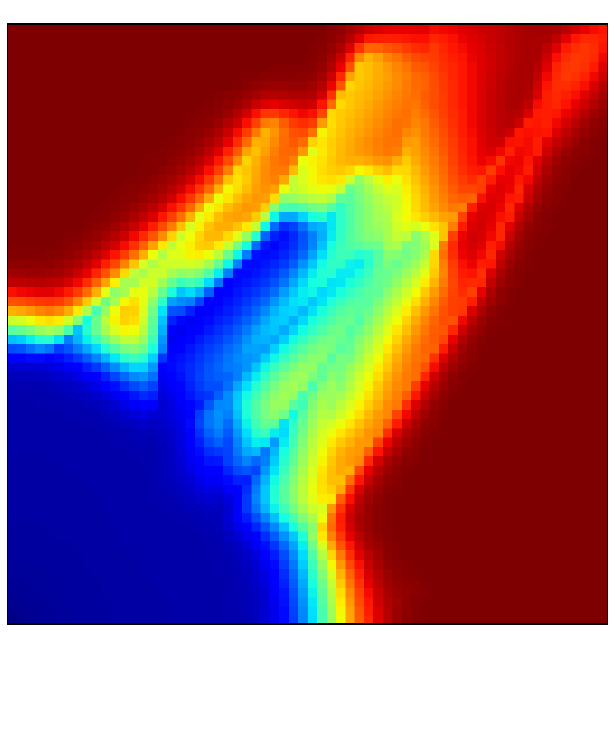}%
\hspace{0.0165\textwidth}%
\includegraphics[width=0.1679\textwidth]{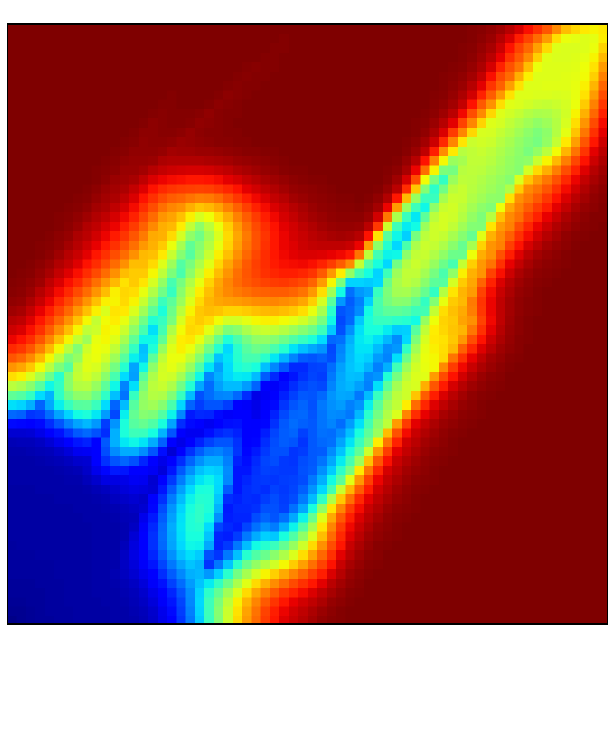}%
\hspace{0.0165\textwidth}%
\includegraphics[width=0.2236\textwidth]{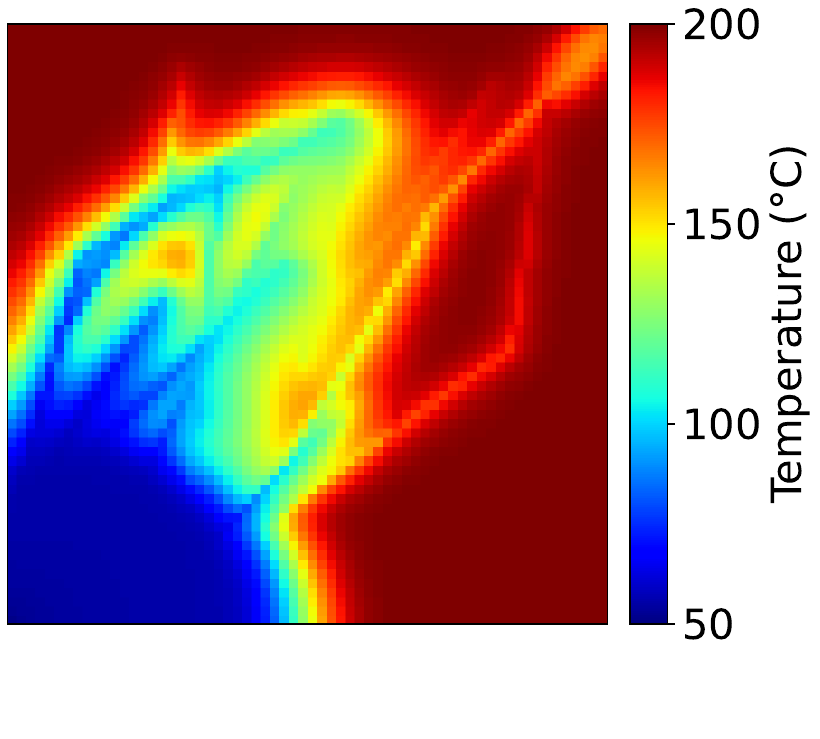}%
}

\makebox[\textwidth][l]{%
\hspace*{0.0410\textwidth}%
\makebox[0.1638\textwidth]{\footnotesize (k) Prior 1}%
\hspace{0.0206\textwidth}%
\makebox[0.1638\textwidth]{\footnotesize (l) Prior 2}%
\hspace{0.0206\textwidth}%
\makebox[0.1638\textwidth]{\footnotesize (m) Prior 3}%
\hspace{0.0206\textwidth}%
\makebox[0.1638\textwidth]{\footnotesize (n) Prior 4}%
\hspace{0.0206\textwidth}%
\makebox[0.1638\textwidth]{\footnotesize (o) Prior 5}%
}

\makebox[\textwidth][l]{%
\hspace*{0.0000\textwidth}%
\includegraphics[width=0.2177\textwidth]{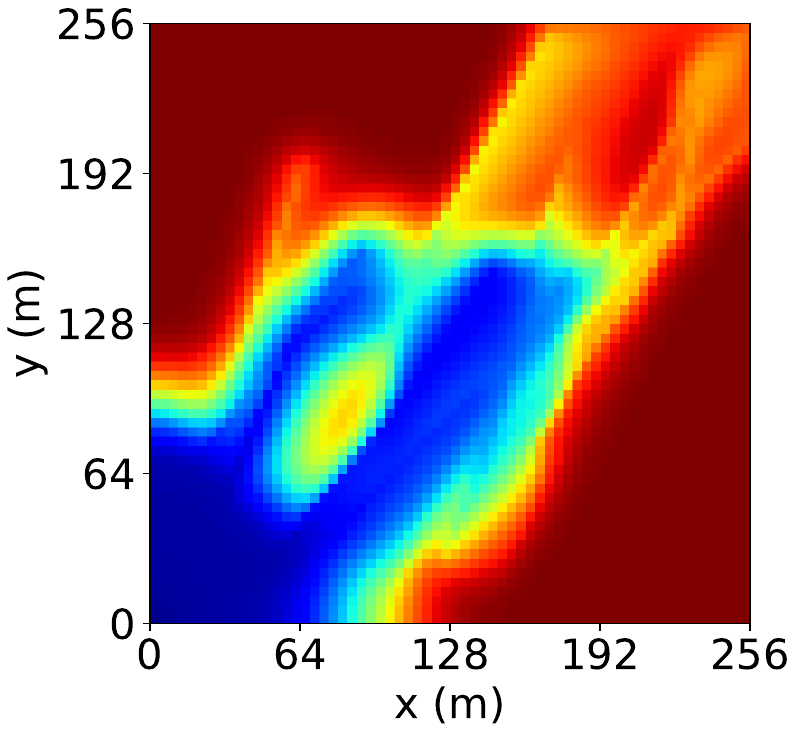}%
\hspace{0.0056\textwidth}%
\includegraphics[width=0.1679\textwidth]{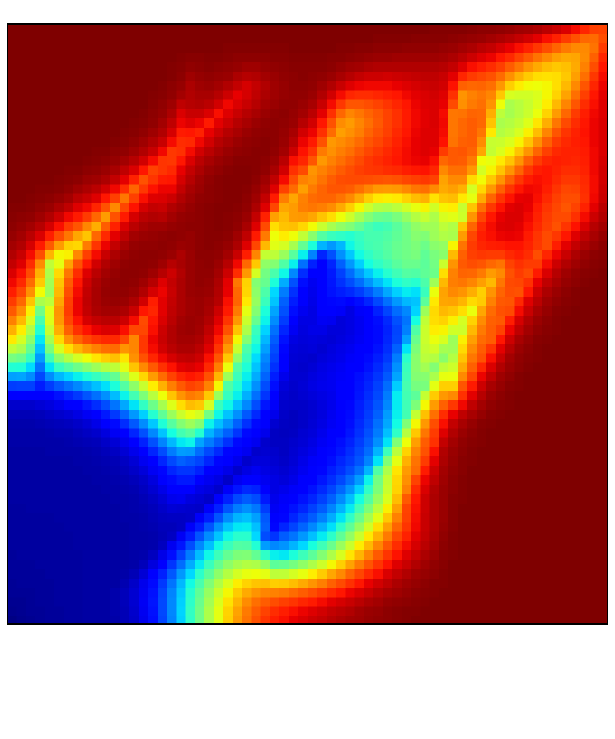}%
\hspace{0.0165\textwidth}%
\includegraphics[width=0.1679\textwidth]{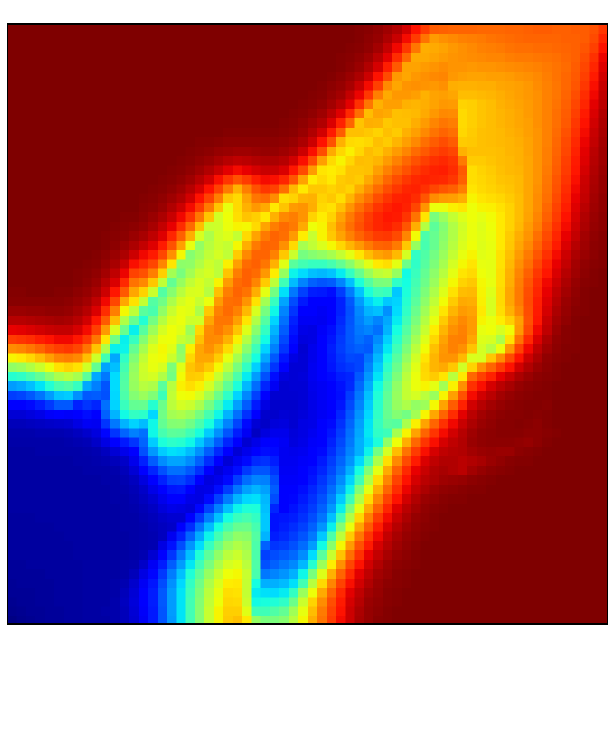}%
\hspace{0.0165\textwidth}%
\includegraphics[width=0.1679\textwidth]{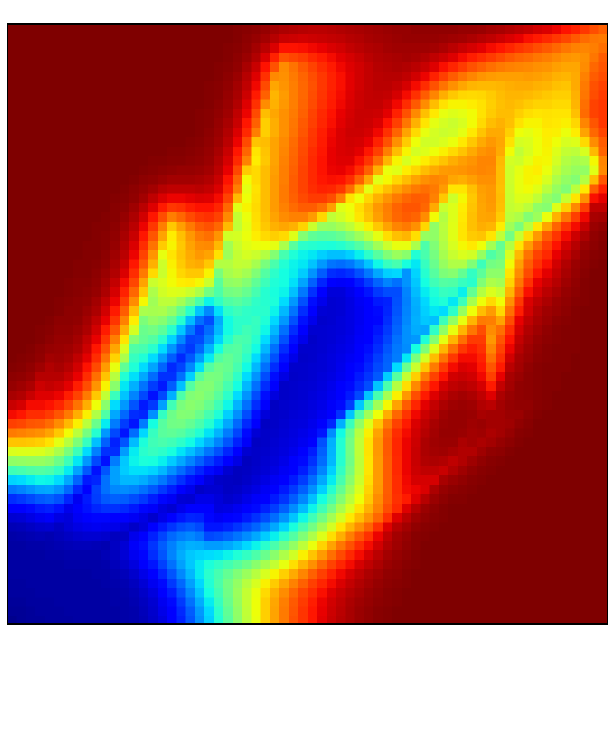}%
\hspace{0.0165\textwidth}%
\includegraphics[width=0.2236\textwidth]{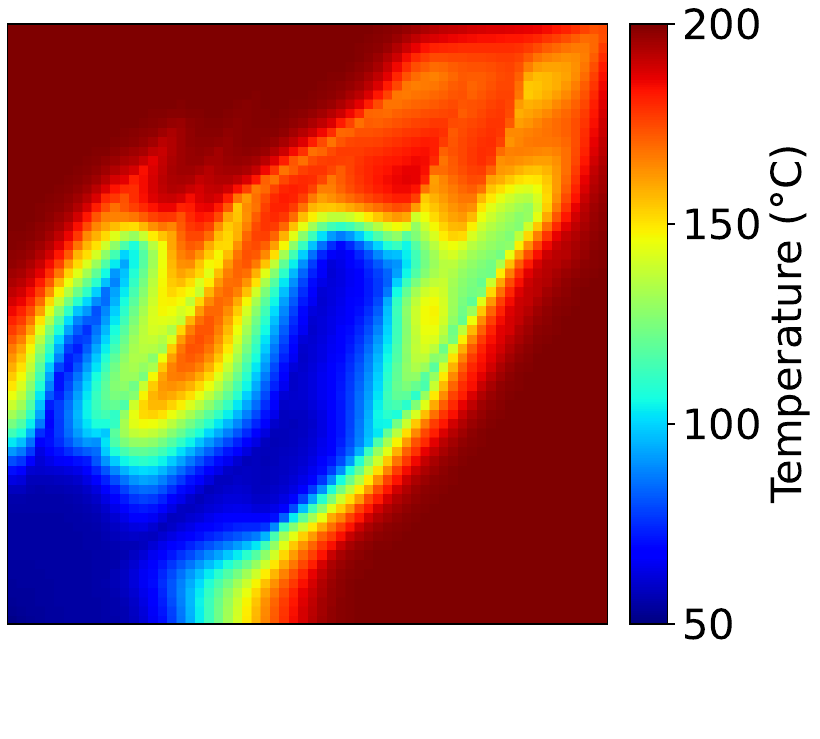}%
}

\makebox[\textwidth][l]{%
\hspace*{0.0410\textwidth}%
\makebox[0.1638\textwidth]{\footnotesize (p) Posterior 1}%
\hspace{0.0206\textwidth}%
\makebox[0.1638\textwidth]{\footnotesize (q) Posterior 2}%
\hspace{0.0206\textwidth}%
\makebox[0.1638\textwidth]{\footnotesize (r) Posterior 3}%
\hspace{0.0206\textwidth}%
\makebox[0.1638\textwidth]{\footnotesize (s) Posterior 4}%
\hspace{0.0206\textwidth}%
\makebox[0.1638\textwidth]{\footnotesize (t) Posterior 5}%
}

\caption*{Figure S6: Representative prior and PT-MCMC posterior DFN realizations and corresponding temperature fields at 1800~days for Case~4, with the six-well monitoring configuration and both temperature and tracer concentration data (T+C). (a-e) prior DFN realizations; (f-j) posterior DFN realizations; (k-o) prior temperature fields; (p-t) posterior temperature fields. For each distribution, five clusters are obtained by applying k-means to the predicted temperature fields from 1,000 realizations; the realization closest to each cluster centroid is selected as representative.}
\label{fig:S6_representatives}
\end{figure}

\end{document}